\documentclass{article}

\usepackage{enumitem}

\usepackage[preprint, nonatbib]{neurips_2025}

\usepackage{comment}
\usepackage[utf8]{inputenc} 
\usepackage[T1]{fontenc}    
\usepackage{hyperref}       
\usepackage{url}            
\usepackage{booktabs}       
\usepackage{amsfonts}       
\usepackage{nicefrac}       
\usepackage{microtype}      
\usepackage{xcolor}         
\usepackage[dvipsnames]{xcolor}

\usepackage{graphicx}
\usepackage{subcaption}
\usepackage{amsmath}
\usepackage{makecell}
\usepackage{array}
\usepackage{multirow}
\usepackage{ragged2e}
\usepackage{longtable}
\newcommand\rev[1]{\textcolor{black}{#1}}
\newcommand\revv[1]{\textcolor{black}{#1}}

\title{TI-DeepONet: Learnable Time Integration for Stable Long-Term Extrapolation}

\author{%
  Dibyajyoti Nayak \\
  Department of Civil and Systems Engineering\\
  Johns Hopkins University\\
  Baltimore, MD, 21218 \\
  \texttt{dnayak2@jh.edu} \\
  \And
  Somdatta Goswami \\
  Department of Civil and Systems Engineering \\
  Johns Hopkins University\\
  Baltimore, MD, 21218 \\
  \texttt{somdatta@jhu.edu} \\
}

\begin{document}

\maketitle

\begin{abstract}
Accurate temporal extrapolation presents a fundamental challenge for neural operators in modeling dynamical systems, where reliable predictions must extend significantly beyond the training time horizon. Conventional deep operator network (DeepONet) approaches rely on two inherently limited training paradigms: fixed-horizon rollouts, which predict complete spatiotemporal solutions while disregarding temporal causality, and autoregressive formulations, which accumulate errors through sequential predictions in time. We introduce \rev{TI-DeepONet (Time-Integrator-embedded Deep Operator Network)}, a framework that integrates neural operators with adaptive numerical time-stepping techniques to preserve the underlying Markovian structure of dynamical systems while substantially mitigating error propagation in extended temporal forecasting. Our approach reformulates the learning objective from direct state prediction to the approximation of instantaneous time-derivative fields, which are subsequently integrated using established numerical schemes. This design naturally supports continuous-time prediction and enables the use of higher-precision integrators at inference than those employed during training, striking a balance between computational efficiency and predictive accuracy. We further develop \rev{TI(L)-DeepONet (Learnable Time-Integrator-embedded Deep Operator Network)}, which incorporates learnable coefficients for intermediate \rev{stages} in a multi-stage numerical integration scheme, thereby adapting to solution-specific variations and enhancing predictive fidelity. Rigorous evaluation across \rev{six} canonical partial differential equations (PDEs) spanning diverse, high-dimensional, chaotic, dissipative, and dispersive dynamics shows that TI(L)-DeepONet marginally outperforms TI-DeepONet, with both frameworks achieving significant reductions in relative $L_2$ extrapolation error: approximately \rev{96.3\%} compared to autoregressive implementations and \rev{83.6\%} compared to fixed-horizon approaches. Notably, both methods maintain stable predictions over temporal domains extending to nearly twice the training interval. This research establishes a physics-aware operator learning paradigm that bridges neural approximation with numerical analysis principles, preserving the causal structure of dynamical systems while addressing a critical gap in long-term forecasting of complex physical phenomena.
\end{abstract}

\section{Introduction}
\label{sec:intro}
Time-dependent partial differential equations (PDEs) are typically solved using spatial discretization (e.g., finite element or volume) combined with time-stepping schemes such as Runge-Kutta or Adams-Bashforth. While accurate, such solvers can be prohibitively expensive for high-dimensional or many-query problems. Explicit integrators are constrained by stability conditions (e.g., CFL), while implicit schemes require solving large systems of equations at each step. Moreover, classical solvers generalize poorly: each new initial or boundary condition requires a fresh simulation.

Neural operators (NOs)~\cite{lu2021learning, lifourier, cao2024laplace, raonic2023convolutional, li2020neural, rahman2022u, tripura2023wavelet} offer an attractive alternative by learning mappings from input functions (e.g., initial conditions) to solutions, bypassing discretization and enabling fast inference. Models like \rev{DeepONet~\cite{lu2021learning, kobayashi2024improved, zhang2025deep, kushwaha2024advanced, xu2024multi}} and FNO~\cite{lifourier} have shown promise in parametric PDE settings. However, extending NOs to time-evolving systems introduces new challenges. The standard approach is to train a model that maps the solution at time $t$ to the solution at time $t+\Delta t$, and roll this forward autoregressively \rev{\cite{hart2023solving}} (see Figure \ref{fig:autoreg-schema}). Unfortunately, this approach suffers from severe error accumulation, as each prediction feeds into the next, compounding inaccuracies and often leading to instability. Alternatively, one can train a model to predict an entire trajectory (``full rollout''), which avoids stepwise feedback but tends to degrade in accuracy when extrapolating beyond the training time window. Crucially, most existing works~\cite{lin2023learning,lifourier,wang2023long,lu2021learning} only evaluate models within the training time span, neglecting the need for reliable extrapolation, a capability that is essential for real-time engineering applications and control tasks. Moreover, existing work~\cite{xu2023transfer, diab2024temporal, liu2022deeppropnet} on autoregressive and full rollout operator models tends to impose a discrete-time Markovian assumption, which misaligns with the underlying continuous-time dynamics of PDEs. These limitations often result in unstable or inaccurate long-term predictions. Thus, a key barrier remains: NOs struggle not only with long-horizon accuracy but also with extrapolating into unseen temporal domains, a prerequisite for trustworthy deployment.

Several recent works attempt to address these limitations. For example, Diab et al.~\cite{diab2025temporal} augment DeepONet with an additional temporal branch to capture dependencies, while Michalowska et al.~\cite{michalowska2024neural} employ recurrent networks (RNNs, LSTMs, GRUs) atop neural operator outputs to better handle sequences. Other approaches introduce \rev{memory modules~\cite{buitragobenefits, he2024sequential, koric2025sequential, hu5149007deepomamba}} to compensate for the non-Markovian nature of rollout-based methods. \rev{We note that such memory-based approaches are conceptually related to the Mori-Zwanzig formalism~\cite{chorin2000optimal}, which introduces memory terms to account for unresolved dynamics; however, our approach takes a different perspective by preserving the Markovian structure inherent in the full-order system through sequential time integration.} However, these strategies largely rely on generic time-series architectures rather than tackling the issue from a dynamical systems perspective.

In this work, we introduce a novel approach inspired by classical numerical analysis. Our proposed TI-DeepONet framework embeds time integration directly into the operator learning process for dynamical systems. Rather than predicting future solution states directly, we train a DeepONet to approximate the instantaneous time derivative $\partial u/\partial t$. During training, this derivative operator is coupled with a Runge-Kutta numerical integrator to evolve the state forward in time. The predicted future state is then compared against the ground truth in the loss function, enabling the network to learn dynamics that respect temporal causality and numerical stability. At inference time, we leverage high-order multistep integrators (such as Adams-Bashforth/Adams-Moulton predictor-corrector schemes) with refined timesteps to reliably propagate solutions based on the learned time derivative operator. This operator learning framework which maps between function spaces rather than from function spaces to vector spaces as in standard neural networks and neural ODEs~\cite{chen2018neural} - provides the flexibility to modify time-stepping parameters between training and testing phases, thereby ensuring stable predictions. We further introduce TI(L)-DeepONet, an enhanced variant that incorporates learnable weighting coefficients \rev{for the intermediate stages} conditioned on the system state at each timestep. This adaptation dynamically adjusts the integration process based on local solution characteristics.

While TI-DeepONet shares conceptual similarities with neural ODEs in learning time derivatives and using numerical integration, our approach differs significantly in both scope and implementation. Neural ODEs typically operate on low-dimensional latent states, whereas TI-DeepONet is specifically designed for high-dimensional spatiotemporal fields and learns an explicit operator mapping between function spaces. Furthermore, unlike neural ODEs, which treat the integrator as fixed, we explicitly embed the integrator within the solution operator, allowing us to employ different integration schemes at inference time according to accuracy and efficiency requirements.

We demonstrate the effectiveness of our approach on \rev{six} benchmark PDEs: (1) 1D Burgers', (2) 1D Korteweg-de Vries (KdV), (3) 1D Kuramuto-Sivashinsky (KS), (4) \rev{2D Burgers'}, \rev{(5) 2D Scalar Rotation Advection-Diffusion}, and \rev{(6) 3D Heat Conduction} equations. We compare our methods against four standard neural operator (NO) baselines: (1) autoregressive DeepONet (DON AR), (2) full rollout DeepONet (DON FR), (3) autoregressive FNO (FNO AR), and (4) full rollout FNO (FNO FR). 
\revv{Across the majority of cases, TI-DeepONet and its adaptive variant TI(L)-DeepONet significantly outperform standard autoregressive and full rollout operator models, achieving stable long-term predictions and accurate extrapolation for time horizons approximately twice as long as the training interval. However, for problems involving rigid body rotational dynamics (e.g., the rotation advection-diffusion problem), FNO AR demonstrates superior performance, highlighting that problem characteristics should inform method selection.}

\section{Methodology}
\label{sec:method}

\subsection{\rev{Problem Setting}}
\label{subsec:problem_setting}
We consider a general time-dependent PDE governing the spatiotemporal evolution of a solution field. Let the solution $u(\mathbf{x}, t)$ be defined over a temporal domain $t \in [0, T]$ and spatial domain $\mathbf{x} = [x_1, x_2, \dots, x_m] \in \mathcal{X} \subseteq \mathbb{R}^m$. The dynamics of $u$ are governed by a PDE that relates the temporal derivative $u_t$ to spatial derivatives $u_{\mathbf{x}}, u_{\mathbf{xx}}, \dots$ via a general nonlinear function $\mathcal{F}$:
\begin{equation}
u_t = \mathcal{F}(t, \mathbf{x}, u, u_{\mathbf{x}}, u_{\mathbf{xx}}, \dots).
\label{eq:time-dep-PDE}
\end{equation}
This study examines the DeepONet framework to investigate error accumulation in autoregressive time-stepping, which serves as the primary motivation for this work. However, it should be noted that this error accumulation phenomenon is not limited to DeepONet but also occurs in other neural operator frameworks, such as FNO. To address this issue, we propose two novel architectures: TI-DeepONet and TI(L)-DeepONet. As will be shown later in this study in Sec.~\ref{sec:results}, these models effectively tackle this exact issue of autoregressive error accumulation. They are observed to significantly stabilize error growth and enable accurate long-term extrapolation. The following subsections provide a detailed introduction to these architectures.

\subsection{Deep operator network (DeepONet)}
\label{subsec:deeponet}
Deep operator networks (DeepONet) are grounded in the universal approximation theorem for nonlinear operators~\cite{chen1995universal}, and are designed to learn mappings between infinite-dimensional input and output function spaces. A standard DeepONet architecture comprises two subnetworks: a branch network and a trunk network. The branch network encodes the input function $\mathbf{v}(\eta)$, evaluated at a set of fixed sensor locations $\{ \eta_1, \eta_2, \dots, \eta_m \}$, producing a set of coefficient weights. The trunk network takes as input the query coordinates $\zeta = (x, y, z, t)$ and outputs the corresponding basis functions. 

The goal of DeepONet is to learn an operator $\mathcal{G}(\mathbf{v})$ such that, for any input realization $\mathbf{v}_j$, the output is a scalar-valued function evaluated at arbitrary spatiotemporal locations $\zeta$. The predicted solution is computed as the weighted inner product of the outputs of the two networks defined as: $\mathcal{G}_{\theta}(\mathbf{v}_j)(\zeta) = \sum_{i=1}^p br_i(\mathbf{v}_j(\eta_1), \dots, \mathbf{v}_j(\eta_m)) \cdot tr_i(\zeta)$, where $br_i(\cdot)$ and $tr_i(\cdot)$ denote the outputs of the branch and trunk networks, respectively, and $\boldsymbol{\theta}$ represents all trainable parameters.

In general, operator learning frameworks like DeepONet aim to learn a solution operator that maps any functional input, such as initial conditions (ICs), boundary conditions (BCs), or source terms, to the corresponding spatiotemporal solution field \rev{(full rollout)}. However, this mapping is inherently non-causal with respect to time: the operator does not explicitly enforce the Markovian nature of dynamical systems, where the future state depends on the present (or past) states. As a result, direct operator prediction over time often fails to preserve temporal coherence, particularly for long-term simulations.

\begin{figure}[htb!]
    \centering
    \includegraphics[width=\linewidth]{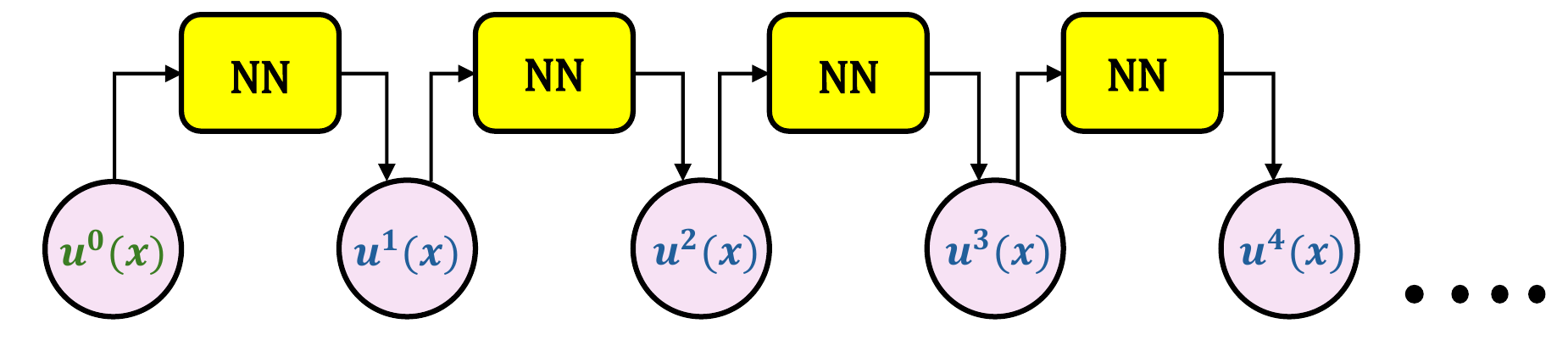}
    \caption{A schematic showing how autoregressive predictions in time are performed.}
    \vspace{-10pt}
    \label{fig:autoreg-schema}
\end{figure}

To account for temporal dynamics, a common strategy is to use autoregressive prediction, where the model recursively predicts the next state using the current state as input. This approach aligns more closely with the natural evolution of dynamical systems and is illustrated schematically in Fig.~\ref{fig:autoreg-schema}. However, autoregressive rollouts are prone to error accumulation and often degrade over long horizons - a core motivation for our proposed time-integrated operator learning framework.

\subsection{Time Integrator embedded deep operator network (TI-DeepONet)}
\label{subsec:deeponet_timeintegrator}
As discussed previously, full spatiotemporal rollouts from a single-step operator such as vanilla DeepONet are not physically meaningful, as they disregard the temporal dependencies between successive solution states. Additionally, autoregressive predictions suffer from error accumulation due to compounding approximation errors over time.

To mitigate these limitations, we propose the TI-DeepONet architecture. Rather than directly predicting the next state $u^{i+1}$, the network learns the right-hand side (RHS) of the time-dependent PDE (Eq.~\ref{eq:time-dep-PDE}), i.e., the time derivative $u_t = \mathcal{F}(t, \mathbf{x}, u, u_{\mathbf{x}}, u_{\mathbf{xx}}, \dots)$. This learned derivative is passed to a numerical time integrator to compute the next solution state from the current one. In particular, we use the RK4 scheme to advance the solution in time. For a timestep $\Delta t$, the RK4 update rule is given by, $k_1 = \mathcal{F}(t^i, u^i)$, $k_2 = \mathcal{F}\left(t^i + \tfrac{\Delta t}{2}, u^i + \tfrac{\Delta t}{2} k_1\right)$, $k_3 = \mathcal{F}\left(t^i + \tfrac{\Delta t}{2}, u^i + \tfrac{\Delta t}{2} k_2\right)$, $k_4 = \mathcal{F}\left(t^i + \Delta t, u^i + \Delta t \cdot k_3\right)$, and $u^{i+1} = u^i + \Delta t  \left( \tfrac{1}{6}k_1 + \tfrac{2}{6}k_2 + \tfrac{2}{6}k_3 + \tfrac{1}{6}k_4 \right)$. This formulation leads to an end-to-end differentiable pipeline, where the loss is computed between the predicted solution $\hat{u}^{i+1}$ and the ground truth $u^{i+1}$, and gradients are backpropagated through the time integrator to update the network parameters. \rev{The loss is computed across all solution states, i.e., $N = N_s \times N_t$, where $N_s$ denotes the number of input realizations (samples) in the dataset and $N_t$ denotes the number of timesteps.} However, the RK4 weighting coefficients, $\{\frac{1}{6}, \frac{2}{6}, \frac{2}{6}, \frac{1}{6}\}$, are fixed.

\paragraph{\rev{Boundary Conditions:}} 
\rev{The TI-DeepONet framework is purely data-driven and does not explicitly enforce boundary conditions in the loss function. Since the network learns a one-step mapping $\mathcal{G}_{\boldsymbol{\theta}}: u^i \rightarrow u^{i+1}$, where both $u^i$ and $u^{i+1}$ are ground truth solutions that inherently satisfy the prescribed boundary conditions, the predicted outputs implicitly respect these conditions. Consequently, the framework can naturally accommodate arbitrary boundary conditions, provided the training data adheres to them.
}
\begin{figure}[htb!]
    \centering
    \includegraphics[width=\linewidth]{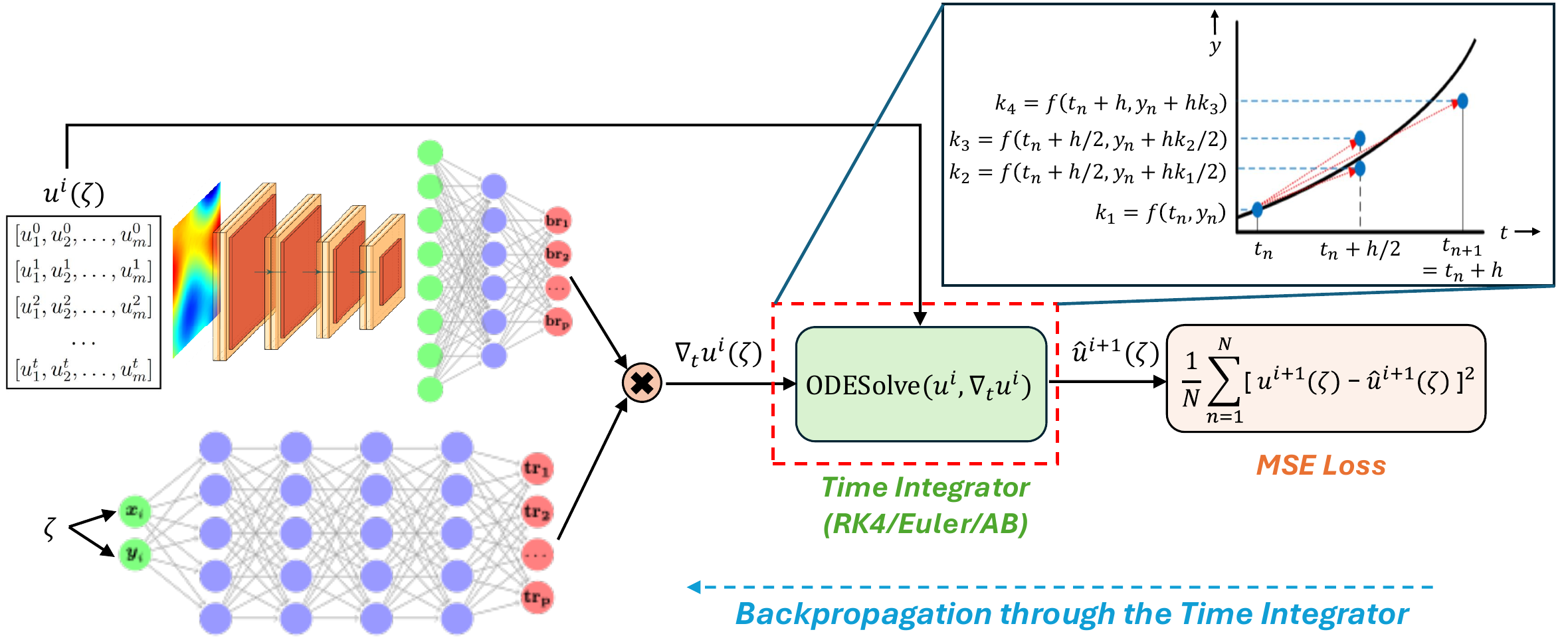}
    \caption{A schematic of the proposed TI-DeepONet architecture.}
    \label{fig:DeepONet_timeintegrator}
    \vspace{-9pt}
\end{figure}

\subsection{Learnable Time Integrator embedded deep operator network (TI(L)-DeepONet)}
\label{subsec:deeponet_learnable_timeintegrator}
To introduce flexibility into the integration process, we extend TI-DeepONet by making the Runge–Kutta weighting coefficients learnable and state-dependent. This leads to the development of TI(L)-DeepONet. Specifically, we replace the constant RK4 slope coefficients, $\{\frac{1}{6}, \frac{2}{6}, \frac{2}{6}, \frac{1}{6}\}$, with adaptive coefficients conditioned on the current solution state $u^i$. We introduce an auxiliary neural network, $\text{NN}_{\text{RK}}$, which maps the current state to a set of four weights, $\boldsymbol{\alpha} = [\alpha_1, \alpha_2, \alpha_3, \alpha_4] = \text{NN}_{\text{RK}}(u^i)$. These weights are then used to compute the next solution state via an adaptive RK4 update:
\begin{equation}
u^{i+1} = u^i + \Delta t \cdot \left( \alpha_1 k_1 + \alpha_2 k_2 + \alpha_3 k_3 + \alpha_4 k_4 \right).
\label{eq:adaptive-rk4}
\end{equation}
\rev{We note that the coefficients $\boldsymbol{\alpha} = \text{NN}_{\text{RK}}(u^i)$ are functions of the solution state $u^i$ rather than time $t$ directly. However, since the solution evolves temporally, the coefficients vary at each timestep, effectively introducing a time-dependence through the evolving state.}

For numerical stability and interpretability, we normalize the weights using \texttt{softmax} activation function, such that: $\tilde{\alpha}_j = \frac{exp(\alpha_j)}{\sum_{\ell=1}^4 exp(\alpha_\ell)}, \forall j = \{1, 2, 3, 4\}$. This adaptive scheme allows the model to adjust the influence of each intermediate slope dynamically, thereby compensating for the approximation error~\cite{lanthaler2022error} of the DeepONet prediction, enabling more robust predictions in stiff, nonlinear, or highly transient regimes. For example, in regions with rapid spatial changes, the model may place greater weight on later-stage evaluations like $k_4$, while in smoother regions, earlier evaluations may suffice.

\section{Results}
\label{sec:results}
\begin{figure}[htb!]
    \centering
    \includegraphics[width=0.85\linewidth]{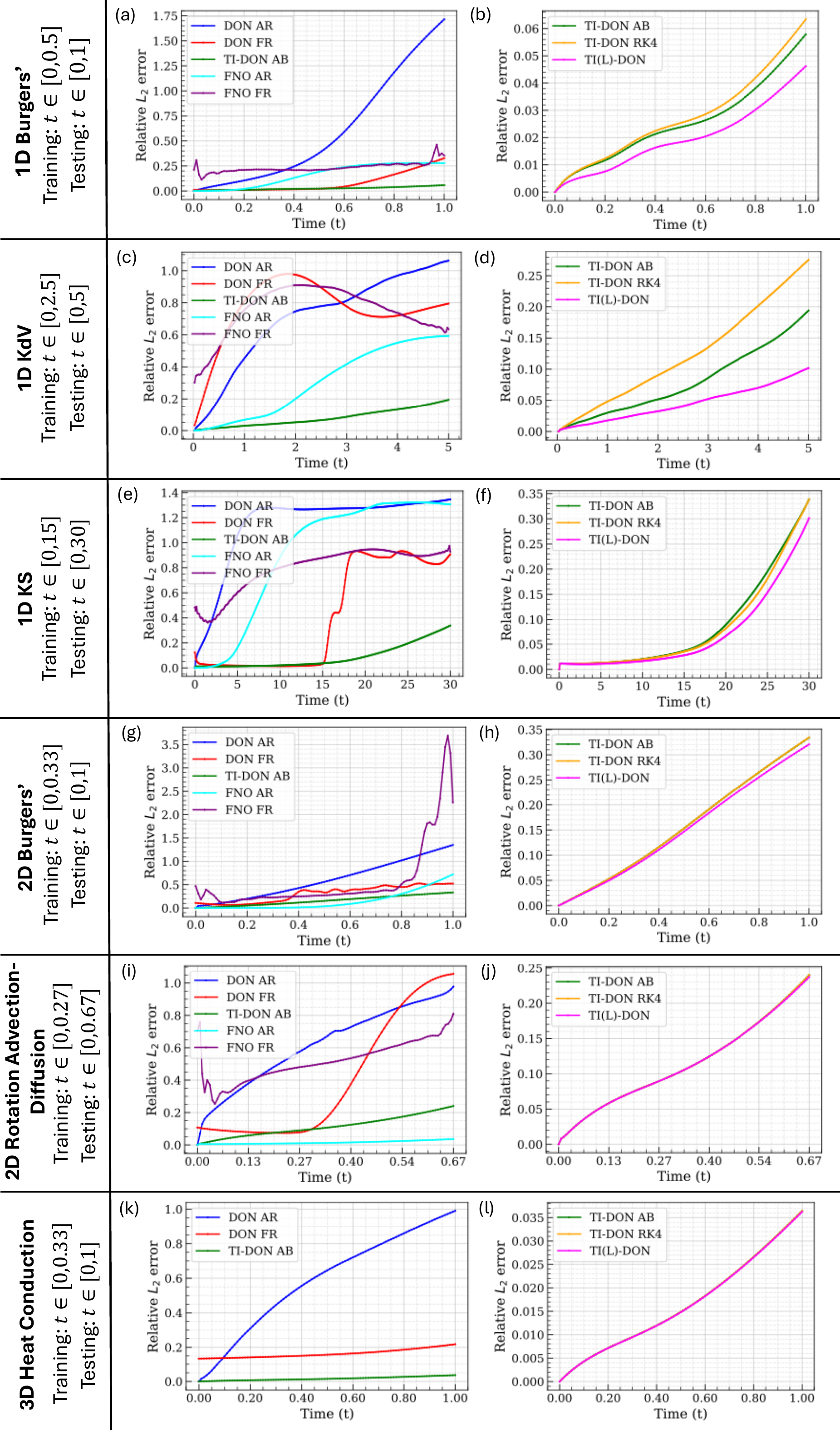}
    \caption{Temporal evolution of the relative $L_2$ error for different frameworks in all applications. Left: Comparison of baseline models with TI-DeepONet AB, which uses RK4 during training and AB2/AM3 during inference. Right: Comparison among TI-based variants. TI-DeepONet RK4 employs RK4 integration in both training and inference, while TI(L)-DeepONet adapts RK4 coefficients via a learnable scheme conditioned on the input state.}
    \vspace{-10pt}
    \label{fig:error_accumulation}
\end{figure}

To evaluate the efficacy of our framework, we consider \rev{six} canonical PDEs: (1) 1D Burgers' equation, (2) 1D KdV equation, (3) 1D KS equation, (4) \rev{2D Burgers' equation}, \rev{(5) 2D Scalar Rotation Advection-Diffusion}, and \rev{(6) 3D Heat Conduction}. Model performance is assessed using the relative $L_2$ error, defined as $\text{error} = \frac{\|u_{\text{pred}} - u_{\text{true}}\|_2}{\|u_{\text{true}}\|_2}$, where $u_{\text{pred}}$ and $u_{\text{true}}$ denote the predicted and ground truth solutions, respectively. Table~\ref{tab:problem_summary} and Figure~\ref{fig:error_accumulation} summarize the comparative training and extrapolation accuracy across all methods for all applications. To evaluate the statistical variability of the performance of the model, we report the mean and standard deviation of \rev{the error metrics} in Table~\ref{tab:problem_summary_independent_runs} based on five independent training trials. \rev{Additional details on data generation and training configurations are provided in the relevant subsections for each PDE example, with network architecture details provided in Supplementary Information (SI)~\ref{sec:add_details}}. The code to reproduce the experiments is publicly available at \url{https://github.com/Centrum-IntelliPhysics/TI-DeepONet-for-Stable-Long-Term-Extrapolation.git}.

\begin{table}[h!]
\caption{Relative $L_2$ errors across frameworks in the extrapolation regime. TI-DeepONet AB employs RK4 during training and AB2/AM3 during inference. The \textbf{bold} values indicate the best-performing method at each time step, while the \underline{underlined} values indicate the second-best.}
    \begin{center}
    \renewcommand{\arraystretch}{1.25}
    \begin{tabular}{|
    >{\RaggedRight\arraybackslash}m{1.55cm}|
    >{\centering\arraybackslash}m{0.625cm}|
    >{\centering\arraybackslash}m{0.85cm}|
    >{\RaggedRight\arraybackslash}m{3cm}|
    *{4}{>{\centering\arraybackslash}m{1.1cm}|}
}
    \hline 
    \multirow{2}{*}{Problem} 
    & \multirow{2}{*}{$t_{train}^{\ast}$} 
    & \multirow{2}{*}{$\Delta t_{e}^{\ast}$} 
    & \multirow{2}{*}{Method}
    & \multicolumn{4}{c|}{Relative $L_2$ error (in extrapolation)} \\
    \cline{5-8}
    & & & & $t$+10$\Delta t_e$ & $t$+20$\Delta t_e$ & $t$+40$\Delta t_e$ & $T^{\ast}$ \\
    \hline

    \multirow{6}{*}{\shortstack[l]{Burgers'\\ (1D)}}
    & \multirow{6}{*}{0.5}
    & \multirow{6}{*}{0.01}
    & TI(L)-DON [Ours]     & \textbf{0.0204}  & \textbf{0.0243}   & \textbf{0.0377}   & \textbf{0.0462}   \\
    &                    &                     & TI-DON AB [Ours]       & \underline{0.0264}  & \underline{0.0310}   & \underline{0.0473}   & \underline{0.0579}   \\
    &                    &                     & DON Full Rollout       & 0.0433   & 0.0965   & 0.2413   & 0.3281   \\
    &                    &                     & DON Autoregressive     & 0.5898   & 0.8742   & 1.4682   & 1.7154   \\
    &                    &                     & FNO Full Rollout     & 0.2298    & 0.2521   & 0.2721    & 0.3551
    \\
    &                    &                     & FNO Autoregressive     & 0.2377   & 0.2636   & 0.2790    & 0.2771
    \\
    \hline

    \multirow{6}{*}{\shortstack[l]{KdV \\ (1D)}}
    & \multirow{6}{*}{2.5}
    & \multirow{6}{*}{0.05}
    & TI(L)-DON [Ours]    & \bf{0.0522}  & \bf{0.0612}   & \bf{0.0701}   & \bf{0.1017}   \\
    &                    &                     & TI-DON AB [Ours]        & \underline{0.0861}  & \underline{0.1114}   & \underline{0.1330}   & \underline{0.1941}   \\
    &                    &                     & DON Full Rollout       & 0.7769   & 0.7163   & 0.7197   & 0.7951   \\
    &                    &                     & DON Autoregressive     & 0.8127   & 0.8985   & 0.9691   & 1.0626   \\
    &                    &                     & FNO Full Rollout     & 0.8666   & 0.8040  & 0.7371   & 0.6352
    \\
    &                    &                     & FNO Autoregressive     & 0.4171   & 0.4954   & 0.5501   & 0.5931
    \\
    \hline

    \multirow{6}{*}{\shortstack[l]{KS \\ (1D)}}
    & \multirow{6}{*}{15}
    & \multirow{6}{*}{0.3}
    & TI(L)-DON [Ours]    & \bf{0.0445}  & \bf{0.079}   & \bf{0.2056}   & \bf{0.3013}   \\
    &                    &                     & TI-DON AB [Ours]        & \underline{0.0589}  & \underline{0.1066}    & \underline{0.2481}  & \underline{0.3366}    \\
    &                    &                     & DON Full Rollout       & 0.8298     & 0.8917     & 0.8482   & 0.9073  \\
    &                    &                     & DON Autoregressive     & 1.2744     & 1.2804     & 1.3204     & 1.3463      \\
    &                    &                     & FNO Full Rollout     & 0.9216   & 0.9468  & 0.8943   & 0.9292
    \\
    &                    &                     & FNO Autoregressive     & 1.2246   & 1.2935   & 1.3207   & 1.3079
    \\
    \hline


    \multirow{6}{*}{\shortstack[l]{\rev{Burgers'} \\ \rev{(2D)}} }
    & \multirow{6}{*}{0.33}
    & \multirow{6}{*}{0.01}
    & TI(L)-DON [Ours]    & \underline{0.1838}  & \underline{0.2204}   & \bf{0.2889}   & \bf{0.3207}   \\
    &                    &                     & TI-DON AB [Ours]        & 0.1911  & 0.2288   & \underline{0.3003}   & \underline{0.3341}   \\
    &                    &                     & DON Full Rollout       & 0.3860  & 0.4406   & 0.5019   & 0.5232   \\
    &                    &                     & DON Autoregressive     & 0.7053   & 0.8589   & 1.1837   & 1.3495   \\
    &                    &                     & FNO Full Rollout     & 0.3095   & 0.3472  & 1.8104   &  2.2611
    \\
    &                    &                     & FNO Autoregressive     & \bf{0.0876}   & \bf{0.1729}   & 0.4856   & 0.7214
    \\
    \hline
    \multirow{6}{*}{\rev{\shortstack[l]{Rotating \\ Advection \\ Diffusion \\ (2D)}}}
    & \multirow{6}{*}{\rev{0.27}}
    & \multirow{6}{*}{\rev{0.008}}
    & \rev{TI(L)-DON [Ours]}    & \rev{\underline{0.1091}}  & \rev{\underline{0.1332}}   & \rev{\underline{0.1970}}   & \rev{\underline{0.2372}}   \\
    &                    &                     & \rev{TI-DON AB [Ours]}        & \rev{0.1091}    & \rev{0.1333}   & \rev{0.1982}     & \rev{0.2400}     \\
    &                    &                     & \rev{DON Full Rollout}       & \rev{0.1926}     & \rev{0.4830}   & \rev{0.9850}     & \rev{1.0562}     \\
    &                    &                     & \rev{DON Autoregressive}     & \rev{0.6874}   & \rev{0.7554}     & \rev{0.8925}     & \rev{0.9777}    \\
    &                    &                     & \rev{FNO Full Rollout}     & \rev{0.5093}   & \rev{0.5509}  & \rev{0.6568}   & \rev{0.8092}
    \\
    &                    &                     & \rev{FNO Autoregressive}     & \rev{\bf{0.0129}}   & \rev{\bf{0.0169}}   & \rev{\bf{0.0286}}   & \rev{\bf{0.0362}}
    \\
    \hline

    \multirow{4}{*}{\rev{\shortstack[l]{Heat \\ Conduction \\ (3D)}}}
    & \multirow{4}{*}{\rev{0.33}}
    & \multirow{4}{*}{\rev{0.015}}
    & \rev{TI(L)-DON [Ours]}    & \rev{\bf{0.0142}}  & \rev{\bf{0.0193}}   & \rev{\bf{0.0327}}   & \rev{\bf{0.0362}}   \\
    &                    &                     & \rev{TI-DON AB [Ours]}        & \rev{\underline{0.0142}}    & \rev{\underline{0.0194}}   & \rev{\underline{0.0328}}     & \rev{\underline{0.0364}}     \\
    &                    &                     & \rev{DON Full Rollout}       & \rev{0.1558}     & \rev{0.1673}   & \rev{0.2058}     & \rev{0.2179}     \\
    &                    &                     & \rev{DON Autoregressive}     & \rev{0.6288}   & \rev{0.7421}     & \rev{0.9465}     & \rev{0.9897}    \\
    \hline

    \multicolumn{8}{l}{\footnotesize{$^{\ast}$$t_{train} = t$: Beginning time of extrapolation;~~~$\Delta t_e$: Evaluation timestep;~~~ $T^{*}$: Final prediction time.}}
    \end{tabular}
    \end{center}
    \vspace{-15pt}
    \label{tab:problem_summary}
\end{table}

\subsection{One-dimensional Burgers' Equation}
\label{subsec:example1}
The viscous Burgers’ equation is a canonical PDE arising in fluid mechanics, nonlinear acoustics, and traffic flow. For the velocity field $u(x, t)$ with viscosity $ \nu = 0.01 $, it is defined as:
\begin{equation}
\frac{\partial u}{\partial t} + u \frac{\partial u}{\partial x} = \nu \frac{\partial^2 u}{\partial x^2}, \quad (x,t) \in [0,1] \times [0,1],
\end{equation}
with periodic BCs and IC, $ u(x, 0) = s(x) $. 

\paragraph{\textbf{\rev{Data generation:}}}
\rev{
The initial condition $s(x)$ is sampled from a Gaussian random field with spectral density $S(k) = \sigma^2(\tau^2 + (2\pi k)^2)^{-\gamma}$, where $\sigma = 25$, $\tau = 5$, and $\gamma = 4$. This ensures $s(x)$ is periodic on $x \in [0, 1]$. The corresponding kernel is expressed via inverse Fourier transform as $K(\mathbf{x, x'}) = \int_{-\infty}^{\infty} S(k) e^{2 \pi ik (\mathbf{x} -  \mathbf{x'})}dk$. We discretize the spatiotemporal domain using 101 grid points along each dimension.
}

\paragraph{\textbf{\rev{Data preparation:}}} 
\rev{The training and testing datasets are constructed differently depending on the learning strategy employed. In the full rollout setting, the goal is to train a DeepONet/FNO to learn a solution operator $\mathcal{G}_{\boldsymbol{\theta}}$ that maps an initial condition $s(x)$ to the full spatiotemporal solution $u(x, t)$, i.e., the mapping $\mathcal{G}_{\boldsymbol{\theta}} \colon s(x) \mapsto u(x,t)$. For this purpose, we consider a set of initial conditions (input to branch network), spatiotemporal query locations (input to trunk network), and the corresponding solution fields as ground truth output. A similar operator learning problem is defined for FNO. The Fourier blocks in FNO employ an FNO 2D-style architecture, applying the Fourier transform across both the $x$ and $t$ coordinates. A single forward pass predicts the entire solution field.}

\rev{The data is partitioned into training and testing sets using a train-test split ratio of 0.8, yielding $N_{\text{train}} = 2000$ and $N_{\text{test}} = 500$ samples. Importantly, we aim to assess the performance of various frameworks in learning the full spatiotemporal field over the entire temporal domain when given access to solutions at limited timesteps. To this end, we restrict training to only half of the temporal domain, i.e., $t_{\text{train}} \in [0, 0.5]$. Consequently, the grid for the trunk network is formed by taking a meshgrid over spatiotemporal coordinates $(x, t)$, where $x \in [0,1]$ and $t \in [0, 0.5]$ during training, while $t \in [0, 1]$ during testing. This same meshgrid is also employed for computing the 2D Fourier Transform in FNO full rollout.}

For training the autoregressive DeepONet, autoregressive FNO, TI-DeepONet, and TI(L)-DeepONet architectures, we adopt a different strategy for preparing the training and testing datasets. In autoregressive methods, future states of the system are recursively predicted using the model's own previous outputs as inputs at each timestep. This approach inherently leads to error accumulation, as prediction errors at one timestep propagate and potentially amplify at subsequent timesteps. Understanding and mitigating this issue is a key motivation behind the TI-DeepONet architecture. To this end, the training dataset is created as follows: the input to the branch network consists of stacked solution states from $t = 0$ to $t = 0.5$, i.e., $[u^0(x), u^1(x), \ldots, u^{50}(x)]$. Similarly, the output ground truth consists of stacked solution states from $t = 0.01$ to $t = 0.51$, i.e., $[u^1(x), u^2(x), \ldots, u^{51}(x)]$. This arrangement enables the solution operator $\mathcal{G}_{\boldsymbol{\theta}}$ to learn the \rev{single-step} mapping $u^i(x) \rightarrow u^{i+1}(x)$, where $u^i(x)$ and $u^{i+1}(x)$ represent the solution states at the $i^{\text{th}}$ and $(i+1)^{\text{th}}$ timesteps, respectively. The trunk network processes only the spatial coordinates $x \in [0,1]$. For autoregressive FNO, the inputs and outputs are distinct $u^i(x)$ and $u^{i+1}(x)$ solution pairs, and the Fourier transform is one-dimensional and is applied only in the $x$-dimension. Autoregressive DeepONet predicts the solution at the next time step, while TI-DeepONet and TI(L)-DeepONet predict the time derivative of the solution at the current timestep and integrate temporal dynamics using numerical solvers. TI-DeepONet employs a Heun-type AB2/AM3 predictor-corrector scheme, defined as:
\begin{align}
\hat{u}^{n+1} &= u^n + \Delta t \left[ \frac{3}{2}\mathcal{F}(t^n, u^n) - \frac{1}{2}\mathcal{F}(t^{n-1}, u^{n-1}) \right], \quad \text{(Predictor)} \\
u^{n+1} &= u^n + \Delta t \left[ \frac{5}{12}\mathcal{F}(t^{n+1}, \hat{u}^{n+1}) + \frac{8}{12}\mathcal{F}(t^n, u^n) - \frac{1}{12}\mathcal{F}(t^{n-1}, u^{n-1}) \right] \quad \text{(Corrector)}
\end{align}

\paragraph{\textbf{\rev{Training architecture:}}} 
\rev{
Tables~\ref{tab:1d_burgers_don-architecture} and \ref{tab:1d_burgers_fno-architecture} present the network architecture details, including the number of layers and neurons per layer, activation functions, total epochs to convergence, batch size, and Fourier modes (where applicable). For the TI(L)-DeepONet architecture, we additionally employ an auxiliary feedforward network designed to predict the four RK4 slope coefficients. This network consists of two hidden layers with 32 neurons each, followed by an output layer with four neurons. The hidden layers use the \texttt{tanh} activation function, while a \texttt{softmax} activation is applied at the output layer to ensure the predicted coefficients sum to one. Training was performed using the Adam optimizer with an initial learning rate of $10^{-3}$, which was exponentially decayed by a factor of 0.95 every 5000 epochs. A similar learning rate scheduling strategy was employed for the auxiliary network predicting the adaptive RK4 slope coefficients. Notably for FNO AR, the learning rate schedule also employed exponential decay but by a factor of 0.96 every 2000 steps. Both training and test losses were monitored at each epoch. The final model corresponds to the set of parameters that yielded the lowest test loss over the entire training process. Among all tested architectures, TI(L)-DeepONet exhibited the fastest convergence, attributed to its capacity to adapt to the local dynamics of the solution. }\rev{Following the performance results for this case, we examine the distribution of the learned RK4 slope coefficients and compare them to the reference values: $\frac{1}{6}, \frac{2}{6}, \frac{2}{6}, \frac{1}{6}$ in a classical RK4 scheme.}

\begin{figure}[htb!]
    \centering
    \includegraphics[width=\linewidth]{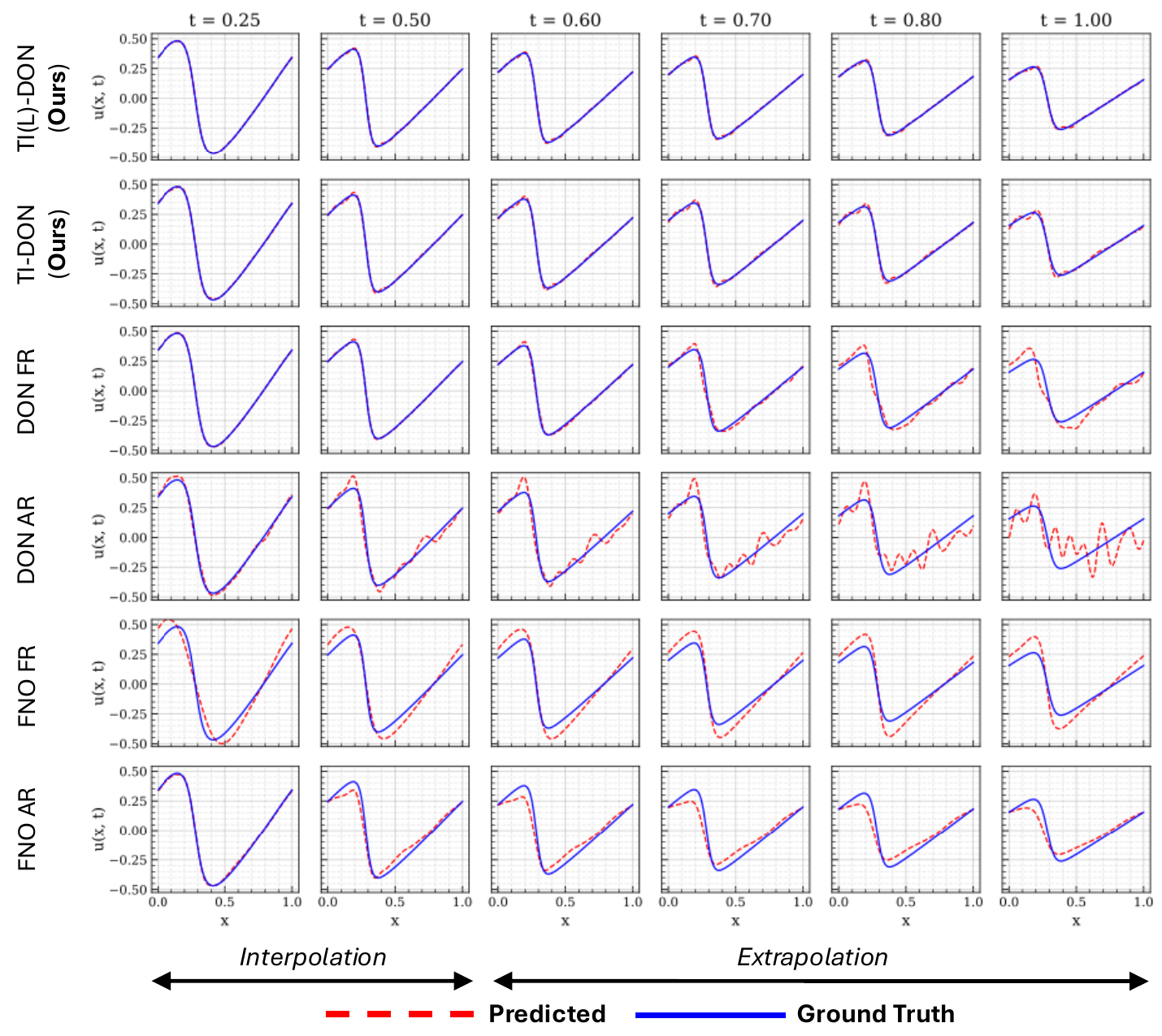}
    \caption{1D Burgers' Equation: Performance of all frameworks in the training ($t\in[0, 0.5]$) and extrapolation ($t\in[0.5, 1]$) regimes for a representative sample.}
    \label{fig:sample_lineplots_1d_burgers}
\end{figure}

\begin{figure}[htb!]
    \centering
    \includegraphics[width=\linewidth]{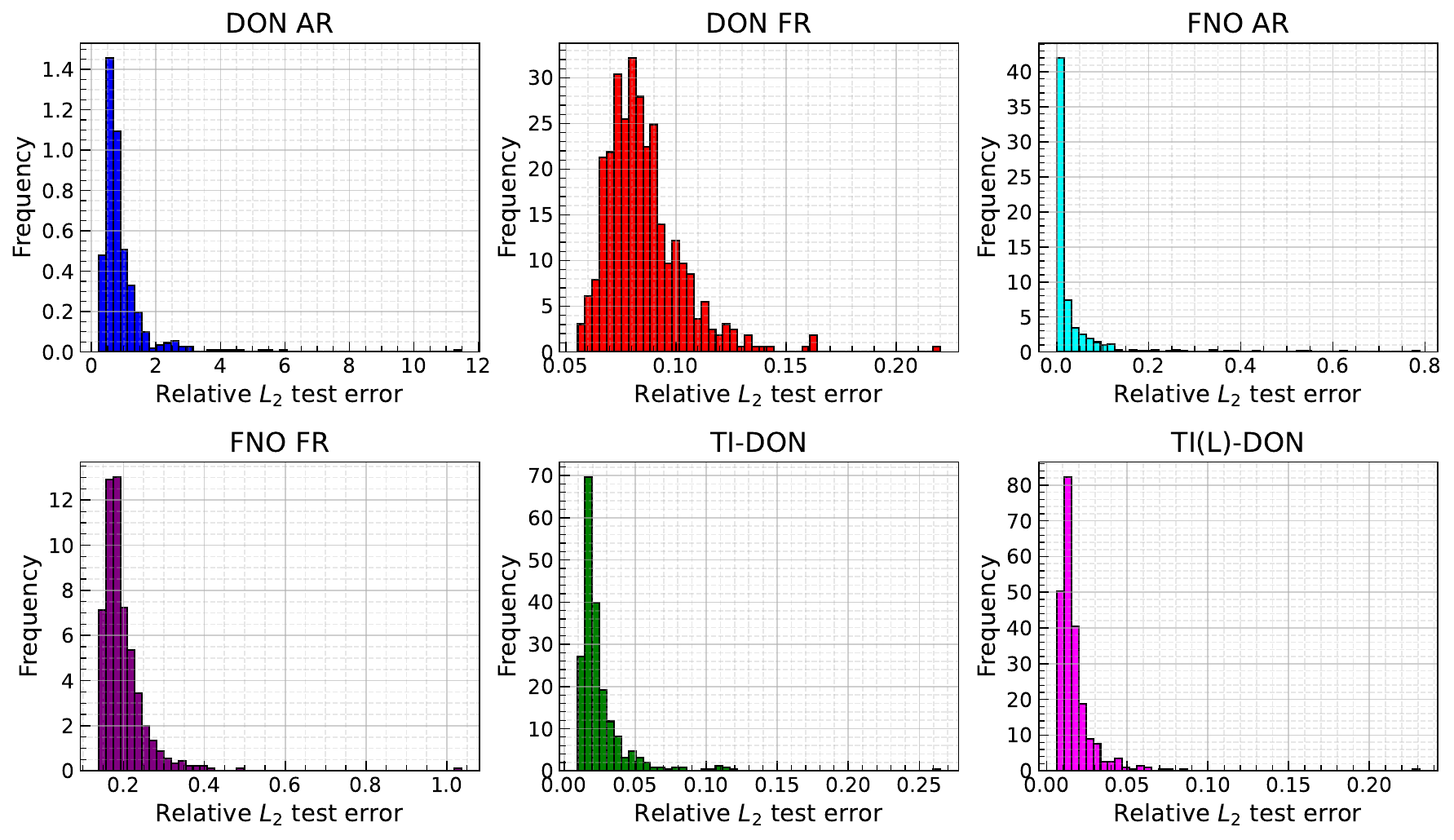}
    \caption{\revv{1D Burgers' Equation: Distribution of relative $L_2$ errors across all test samples for all frameworks, averaged over the entire spatiotemporal domain.}}
    \label{fig:test_err_dist_1d_burgers}
\end{figure}

Figure~\ref{fig:sample_lineplots_1d_burgers} illustrates the performance of all models for a representative test sample. All models perform well within the training interval; however, the autoregressive DeepONet \rev{(DON AR)} accumulates error early and diverges during extrapolation. The quantitative trends presented in Table~\ref{tab:problem_summary} underscore the qualitative observations from Figure~\ref{fig:sample_lineplots_1d_burgers} as well as Figures~\ref{fig:error_accumulation}(a) and (b). The autoregressive DeepONet \rev{(DON AR)} deteriorates rapidly during extrapolation due to cumulative errors in the absence of stabilization mechanisms. While the full rollout DeepONet \rev{(DON FR)} performs better owing to its non-recursive architecture, it fails to maintain accuracy beyond the training interval since it does not leverage the Markovian structure of time-dependent PDEs. Regarding the FNO variants, both full rollout \rev{(FNO FR)} and autoregressive FNO \rev{(FNO AR)} perform reasonably well within the training domain. However, in the extrapolation domain, both methods incur relatively higher errors, albeit lower than those of \rev{DON AR}. Notably, \rev{FNO AR} does not exhibit the steep error growth observed in \rev{DON AR} during extrapolation. The full rollout FNO \rev{(FNO FR)} maintains higher errors throughout the entire temporal domain, which can be attributed to its limited capacity to capture the full spectrum of frequencies in the system, as it is trained on solution information from only half of the temporal domain. 

\begin{figure}[htb!]
    \centering
    \includegraphics[width=\linewidth]{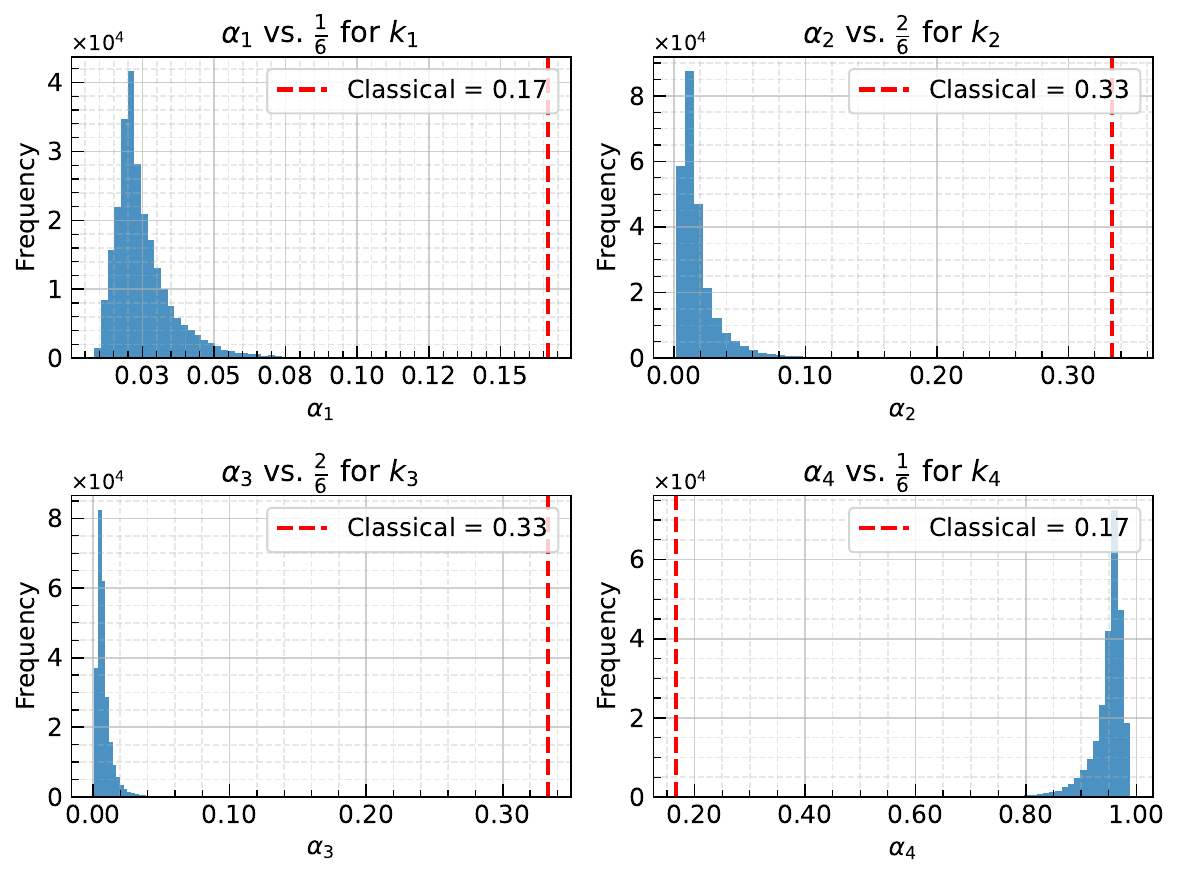}
    \caption{\rev{1D Burgers' Equation: Distribution of the optimally learned $\alpha_i~\forall i \in \lbrace1,2,3,4\rbrace$ for the RK4 slope coefficients.}}
    \label{fig:alpha_dist_1d_burgers}
\end{figure}
In contrast, the proposed TI-DeepONet variants, which embed numerical integration schemes into the learning framework, demonstrate superior extrapolation accuracy (see Figures \ref{fig:error_accumulation}(a) and (b)). TI-DeepONet with AB2/AM3 integration \rev{(TI-DON/TI-DON AB)} effectively stabilizes long-horizon predictions. TI(L)-DeepONet \rev{(TI(L)-DON)} further improves on this by learning adaptive RK4 slope coefficients conditioned on the current state, enabling it to modulate time-stepping dynamically in response to the local solution behavior. This is particularly beneficial in stiff or nonlinear regimes, as reflected in the consistently lowest errors observed across all time steps. 
\rev{This observation is further reinforced in Figure~\ref{fig:test_err_dist_1d_burgers}, where the error distributions across test samples are comparable for both TI-DeepONet variants, with TI(L)-DeepONet exhibiting marginally smaller errors across a larger number of test samples.} \revv{Among the baseline methods, the ranking in terms of test accuracy is: FNO AR $>$ DON FR $>$ FNO FR $>$ DON AR.}
Figure \ref{fig:learnableweights} in SI presents iteration-wise learning of the RK4 weighting coefficients, comparing them against the reference. An important point to note here is that the obtained optimized coefficients are compensating for the approximation error of the TI-DeepONet; hence, they may not necessarily adhere to the values obtained using the Taylor series expansion.

\rev{
A detailed investigation into the behavior of the learned $\boldsymbol{\alpha}$ coefficients is presented in Figure~\ref{fig:alpha_dist_1d_burgers}. One can observe that the optimally learned $\alpha_i$'s differ significantly from the classical RK4 weighting coefficients $\{\frac{1}{6}, \frac{2}{6}, \frac{2}{6}, \frac{1}{6}\}$. Notably, a higher emphasis is placed on the weight corresponding to the last computed slope $k_4$, with $\alpha_4$ exhibiting values closer to 1.0 compared to $\alpha_1$, $\alpha_2$, and $\alpha_3$. We hypothesize that this behavior arises because the slopes $k_i$ are computed using a neural network approximation of the time derivative, which inherently carries some approximation error. The classical RK4 weighting coefficients are derived under the assumption of exact slope evaluations to achieve 4th-order truncation error. However, when the slopes themselves are approximate, the optimal weighting may shift from minimizing truncation error to compensating for approximation error in the predicted time derivative. The auxiliary network learns coefficients $\boldsymbol{\alpha}$ that likely reflect a trade-off minimizing the overall prediction error for the specific dynamics being modeled. For the 1D Burgers' equation, the network appears to discover that assigning higher weight to $k_4$, which is evaluated at the most advanced intermediate state $u^n + \Delta t \cdot k_3$, closest to the target $u^{n+1}$ yields more accurate predictions. This suggests that for smooth dynamics, the slope evaluation nearest to the target state may provide the most reliable estimate of the integrated dynamics over the timestep. A rigorous theoretical analysis of the order of accuracy and convergence properties of the learned adaptive weighting coefficients remains an important direction for future work.
}

\subsection{One-dimensional Korteweg-de Vries (KdV) Equation}
\label{subsec:example2}
The one-dimensional KdV equation models nonlinear dispersive waves and arises in contexts such as shallow water dynamics, plasma physics, and lattice acoustics. It introduces third-order dispersion into the Burgers-like formulation:
\begin{equation}
    \frac{\partial u}{\partial t} - \eta u \frac{\partial u}{\partial x} + \gamma \frac{\partial^3 u}{\partial x^3} = 0
    \label{eq:kdv-pde},
\end{equation}
where, $u$ is the amplitude of the wave, $\gamma$ and $\eta$ are some real-valued scalar parameters, $x$ and $t$ are the spatial and temporal dimensions, respectively. 

\paragraph{\textbf{\rev{Data generation:}}}\rev{The initial condition, $u(x, 0)$ is defined as a sum of two solitons, i.e., $u = u_1 + u_2$. A soliton is a nonlinear, self-reinforcing, localized wave packet that is strongly stable. It is expressed as:
\begin{equation}
    u_i (x, 0) = 2k_i^2 \, \text{sech}^2 \left( k_i \left ( \left ( x + \frac{P}{2} - Pd_i \right )\%P-\frac{P}{2} \right ) \right )
    \label{eq:soliton}
\end{equation}
where $\text{sech}$ is the hyperbolic secant, $P$ is the period in space, \% is the modulo operator, and $i = \lbrace 1, 2 \rbrace$.} \revv{The coefficients $k_i$ and $d_i$, which determine the height and location of the soliton peak, respectively, are independently and uniformly sampled from the intervals $k_i \in [0.3, 0.7]$ and $d_i \in [0, 1]$ for each soliton. The spatial period is set to $P = 10$.} \rev{The dataset is curated by generating $N = 1000$ distinct initial conditions and their temporal evolution is modeled using the midpoint method. The 1D spatial domain, $\Omega = [0, 10]$, is discretized into 100 uniformly spaced grid points, resulting in a spatial resolution of $\Delta x = 0.1$. For the temporal domain, we consider a uniform grid of 201 points spanning the time domain $t \in [0, 5]$, which yields $\Delta t = 0.025$.} \rev{For further details regarding the problem definition, the reader is referred to \cite{michalowska2024neural}.}

\paragraph{\textbf{\rev{Data preparation:}}} \rev{To assess the extrapolation accuracy of the different methods, we train only on the first half of the temporal domain, i.e., $t_{\text{train}} \in [0, 2.5]$, and evaluate predictive performance over the entire domain ($t \in [0, 5]$). To reiterate, in the full rollout setting, the inputs to the trunk network are constructed from a meshgrid of $(x, t_{\text{train}})$, whereas in the autoregressive setting, only the spatial coordinates $(x)$ are used as trunk inputs. Similarly, the same meshgrid that serves as trunk inputs during DeepONet FR training becomes the axes along which the 2D Fourier transform is performed (all $x$ coordinates, seen $t$ coordinates) for FNO FR training. While the initial condition samples serve as input to the branch network in the DeepONet full rollout model, the input to the branch network in the autoregressive and TI-based DeepONet frameworks consists of stacked solution states from $t = 0$ to $t = 2.5$, i.e., $[u^0(x), u^1(x), \ldots, u^{100}(x)]$. For autoregressive FNO, the inputs and outputs are the solution pairs $u^i(x)$ and $u^{i+1}(x)$, with the 1D Fourier transform applied along the spatial dimension $x$ only. The train-test split is maintained at 0.8.}

\paragraph{\textbf{\rev{Training architecture:}}}
\rev{
Tables~\ref{tab:1d_kdv_don-architecture} and \ref{tab:1d_kdv_fno-architecture} present the network architecture details, including the number of layers, neurons per layer, activation functions, total epochs to convergence, batch size, and Fourier modes (where applicable). For the TI(L)-DeepONet framework, we employ a secondary feedforward network to predict the four RK4 slope coefficients. This network consists of two hidden layers with 64 and 48 neurons, respectively, followed by an output layer with four neurons. The hidden layers use the \texttt{tanh} activation function, while a \texttt{softmax} activation is applied at the output layer to ensure that the predicted coefficients sum to one. Training was performed using the Adam optimizer with an initial learning rate of $10^{-3}$, which was exponentially decayed by a factor of 0.95 every 5000 epochs. A similar learning rate schedule was used for training the auxiliary network tasked with learning the RK4 slope coefficients. The same learning rate schedule with identical hyperparameters was employed for FNO FR training. For FNO AR, an exponential decay learning rate schedule was also used, but with an initial learning rate of $10^{-3}$ that decayed by a factor of 0.96 every 2000 steps.
}
\begin{figure}[htb!]
    \centering
    \includegraphics[width=\linewidth]{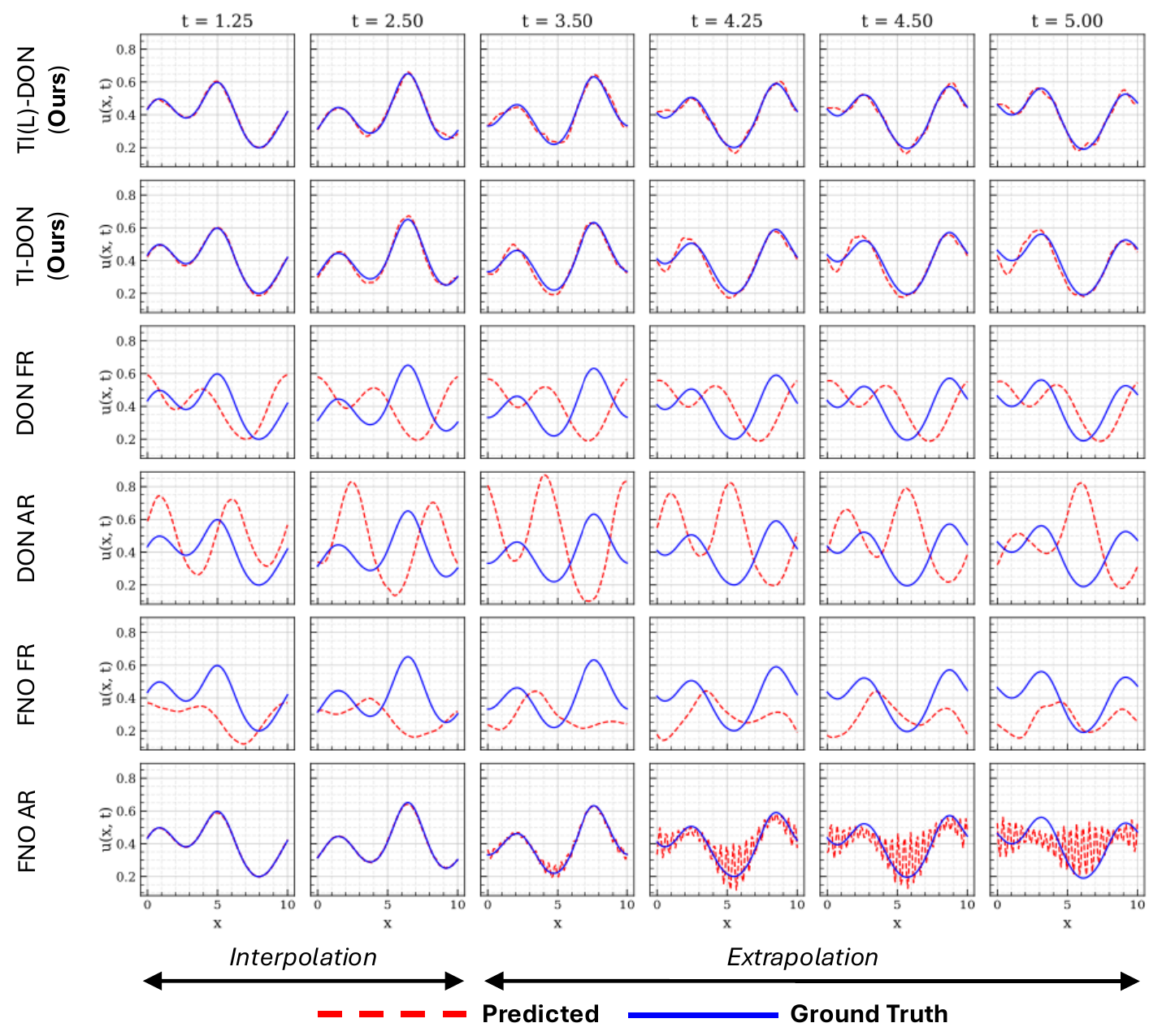}
    \caption{1D KdV Equation: Performance of all the frameworks in the training ($t\in[0, 2.5]$) and the extrapolation regime ($t\in[2.5, 5]$) for a representative sample.}
    \label{fig:sample_lineplots_1d_kdv}
\end{figure}

\begin{figure}[htb!]
    \centering
    \includegraphics[width=\linewidth]{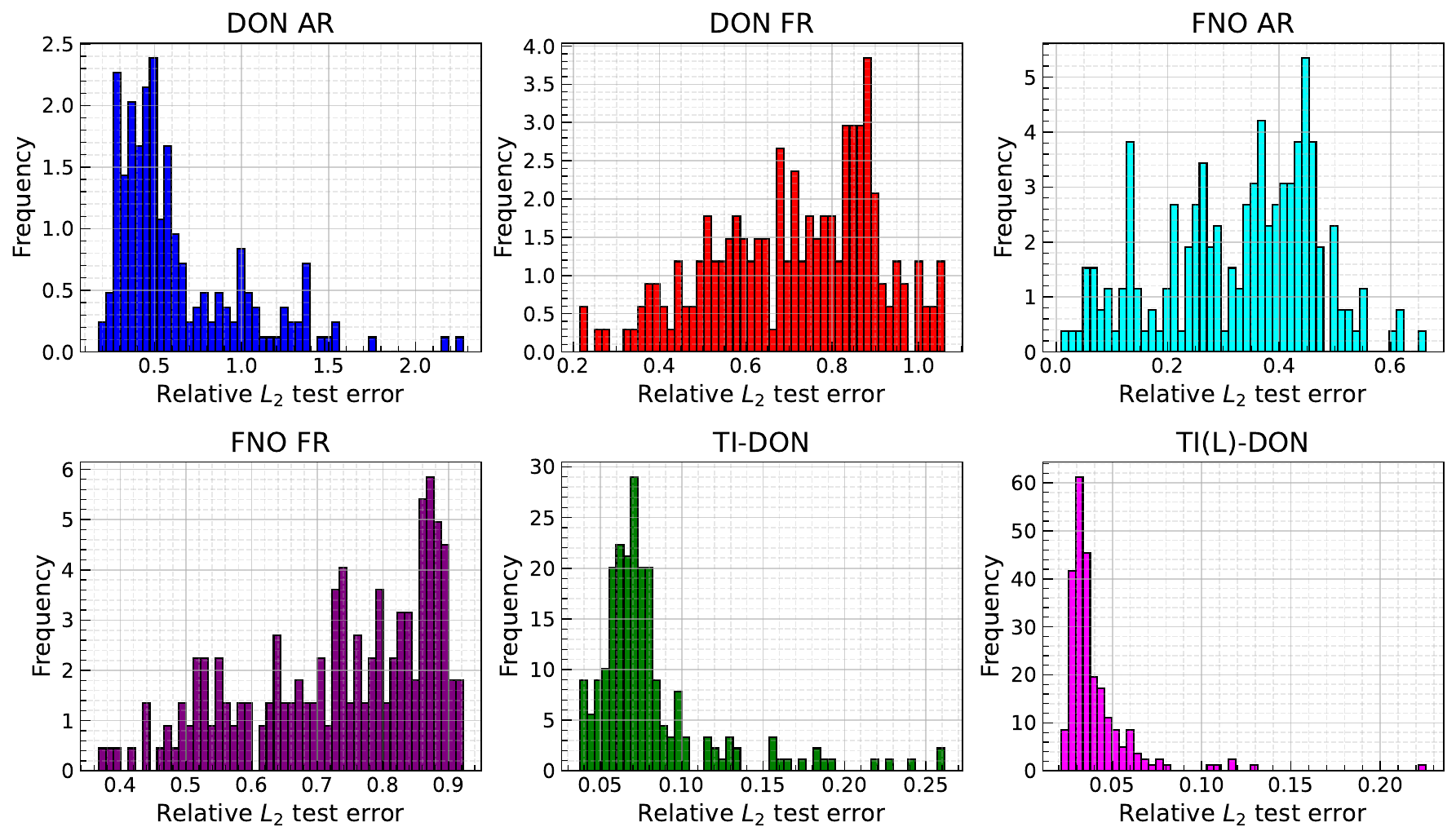}
    \caption{\revv{1D KdV Equation: Distribution of relative $L_2$ errors across all test samples for all frameworks, averaged over the entire spatiotemporal domain.}}
    \label{fig:test_err_dist_1d_kdv}
\end{figure}
Figure~\ref{fig:sample_lineplots_1d_kdv} presents predictions from different frameworks on a representative sample. Both \rev{DON FR} and \rev{DON AR} models fail to capture the dispersive, periodic dynamics of the solution. The full rollout approach, which treats time steps independently, lacks temporal continuity, while the autoregressive model suffers from compounding errors during recursive inference. Regarding the FNO variants, \rev{FNO FR} exhibits a noticeable phase shift due to its inability to capture the full spectrum of system frequencies; a consequence of being trained on a reduced temporal domain, thereby incurring significant errors throughout the entire temporal domain. \rev{FNO AR} demonstrates particularly interesting behavior: while it achieves near-perfect predictions within the training domain, its performance deteriorates markedly upon entering the extrapolation regime. Here, autoregressive errors accumulate and manifest as spectral instability, characterized by frequency leakage. Specifically, the system's frequencies become amplified at longer temporal horizons, resulting in erratic oscillations in the solution. Consequently, this approach also fails to reliably capture the inherent dispersive and periodic dynamics of the KdV PDE in the extrapolation domain. In contrast, \rev{TI-DON} and \rev{TI(L)-DON} closely match the ground truth, even in the extrapolation regime. \rev{The adaptive RK4 weighting scheme in TI(L)-DeepONet yields marginally improved accuracy compared to TI-DeepONet.}

Figures~\ref{fig:error_accumulation}(c) and (d) show the relative $L_2$ error over time. \rev{DON AR} exhibits rapid error growth beyond $t=2.5$, while the full rollout model \rev{(DON FR)} shows non-monotonic behavior due to its limited ability to model coherent temporal evolution. \rev{FNO FR} also exhibits a non-monotonic trend throughout the temporal domain, with the largest errors concentrated in the interval $t \in [1,3]$. \rev{FNO AR} achieves lower prediction errors within the training temporal domain but exhibits immediate steep error growth beyond $t = 2$. These quantitative $L_2$ error observations are in good agreement with the qualitative trends observed for the baselines in Figure~\ref{fig:sample_lineplots_1d_kdv}. In comparison, the TI-based models maintain low and stable errors across the entire prediction window. At $t=5$, TI(L)-DeepONet and TI-DeepONet achieve final errors of approximately 20\% and 19.4\%, respectively; 4-5$\times$ lower than baseline methods. The marginal performance difference between TI-DeepONet with AB2/AM3 and TI(L)-DeepONet indicates that both are reliable choices for long-term modeling of PDEs similar to the KdV equation. 
\rev{Finally, the error distributions across all test samples are provided for TI-DON and TI(L)-DON in Figure~\ref{fig:test_err_dist_1d_kdv}, where noticeably better performance of TI(L)-DON is observed compared to TI-DON, with a significantly larger number of samples exhibiting overall relative $L_2$ errors under 5\%.} 
\revv{Among the other NO baselines, the ranking in terms of test accuracy is: FNO AR $>$ FNO FR $\approx$ DON FR $>$ DON AR.}
\rev{For brevity, we do not present the distribution of the optimally learned $\boldsymbol{\alpha}$ coefficients as shown in the previous 1D Burgers' case, since in this case, across all solution states, we consistently observed $\alpha_4$ to be extremely close to 1, with $\alpha_1$, $\alpha_2$, and $\alpha_3$ tending toward zero.}

\subsection{One-dimensional Kuramuto-Sivashinsky (KS) Equation}
\label{subsec:example3}
The one-dimensional Kuramoto-Sivashinsky (KS) equation is a highly nonlinear PDE that is widely known for its chaotic behavior. It arises in various physical contexts such as thin film dynamics, flame fronts, reaction-diffusion systems, and plasma instabilities. The equation introduces a combination of nonlinear advection, second-order dissipation, and fourth-order dispersion terms, expressed as:
\begin{equation}
    \dfrac{\partial u}{\partial t} = -u \dfrac{\partial u}{\partial x} - \dfrac{\partial^2 u}{\partial x^2} - \dfrac{\partial^4u}{\partial x^4},~~~(x, t) \in [0, L] \times (0, \infty)
    \label{eq:1d_ks_pde}
\end{equation}
\begin{equation}
    u(x, t =0) = u_0~~~x\in[0,L]
    \label{eq:1d_ks_pde_ICs}
\end{equation}
Here, $u$ represents the evolving scalar field with an initial profile $u_0$. The spatial domain $L$ is equipped with periodic boundary conditions. The nonlinear term $-u\dfrac{\partial u}{\partial x}$ accounts for advection, while the dissipative term $\dfrac{\partial ^2u}{\partial x^2}$ contributes to numerical stability, and the high-order dispersive term $\dfrac{\partial^4 u}{\partial x^4}$ models the long-wavelength instability mechanisms that ultimately lead to chaotic dynamics. 

\paragraph{\textbf{\rev{Data generation:}}}
\rev{
We consider the 1D Kuramoto-Sivashinsky (KS) equation on a spatial domain $x \in [0, L]$ with $L = 6\pi$, prescribed with periodic boundary conditions. The initial conditions $u(x, 0) = s(x)$ are sampled from mean-zero Gaussian random fields (GRFs) with periodic boundary conditions. Specifically, we generate GRFs with covariance operator $\mathcal{C} = \sigma^2(-\Delta + \tau^2 I)^{-\gamma}$, where $\gamma = 1$, $\tau = 2$, and $\sigma = 1$. The spatial domain is discretized using $N_x = 128$ grid points, and the temporal domain $t \in [0, 30]$ is discretized with $N_t = 300$ timesteps. The PDE is solved using the exponential time-differencing fourth-order Runge-Kutta (ETDRK4) method with a timestep of $\Delta t = 0.1$. We generate a total of 3000 samples, each consisting of the initial condition and the corresponding spatiotemporal solution field $u(x,t)$. The dataset is partitioned using a train-test split ratio of 0.8, yielding $N_{\text{train}} = 2400$ training samples and $N_{\text{test}} = 600$ testing samples. The periodic boundary conditions ensure that the solution remains in the space of mean-zero periodic functions throughout the evolution, preserving the chaotic dynamics characteristic of the KS equation. For further details regarding the data generation procedure, the reader is referred to \cite{li2021learning}, from which the data generation code is adapted.
}

\paragraph{\textbf{\rev{Data preparation:}}} 
\rev{
For evaluating the extrapolation capabilities of various methods, we restrict training to the initial half of the temporal domain, specifically $t_{\text{train}} \in [0, 15]$, while testing spans the complete temporal interval $t \in [0, 30]$. In the full rollout configuration, the trunk network receives spatiotemporal coordinate pairs $(x, t_{\text{train}})$ arranged in a meshgrid format. Conversely, the autoregressive approach utilizes only spatial coordinates $x$ as trunk inputs. For FNO FR, the 2D Fourier transform operates along both spatial and temporal dimensions using the same coordinate grid employed in DeepONet FR training. The branch network architecture differs between approaches: full rollout DeepONet processes initial conditions $s(x)$ directly, while autoregressive and TI-based variants receive a sequence of solution snapshots spanning $t = 0$ to $t = 15$, comprising states $[u^0(x), u^1(x), \ldots, u^{150}(x)]$, with the corresponding output solution states $[u^1(x), u^2(x), \ldots, u^{151}(x)]$. The autoregressive FNO framework also operates on consecutive solution pairs $(u^i(x), u^{i+1}(x))$ as input-output mappings, employing a 1D Fourier transform exclusively in the spatial dimension.
}

\paragraph{\textbf{\rev{Training architecture:}}}
\rev{
Tables~\ref{tab:1d_ks_don-architecture} and \ref{tab:1d_ks_fno-architecture} present the network architecture details, including the number of layers, neurons per layer, activation functions, total epochs to convergence, batch size, and Fourier modes (where applicable). All DeepONet variants employ Fourier feature encoding in the trunk network. For the full rollout DeepONet, both spatial and temporal coordinates $(x, t)$ are augmented with sinusoidal features computed as $[\sin(\pi k x), \cos(\pi k x), \sin(\pi k t), \cos(\pi k t)]$ for $k \in \{1, 2, \ldots, 10\}$, expanding the input dimension from 2 to 42. For the autoregressive and TI-based variants, only the spatial coordinate $x$ is augmented with Fourier features $[\sin(\pi k x), \cos(\pi k x)]$, expanding the input dimension from 1 to 21. For the TI(L)-DeepONet framework, we employ an auxiliary feedforward network to predict the four RK4 slope coefficients. This network consists of two hidden layers with 64 neurons each, followed by an output layer with four neurons. The hidden layers utilize the \texttt{tanh} activation function, while a \texttt{softmax} activation is applied at the output layer to ensure the predicted coefficients sum to one. All DeepONet variants (DON FR, DON AR, TI-DON, and TI(L)-DON) are trained using the AdamW optimizer with an initial learning rate of $10^{-3}$, exponentially decayed by a factor of 0.95 every 5000 epochs, and a weight decay of $10^{-4}$. For TI(L)-DeepONet specifically, the auxiliary RK4 network is trained separately using the Adam optimizer with an initial learning rate of $2 \times 10^{-3}$ and the same exponential decay schedule. The FNO variants employ the Adam optimizer with an initial learning rate of $10^{-3}$, exponentially decayed by a factor of 0.96 every 2000 steps for both FNO FR and FNO AR configurations.
}
\begin{figure}[htb!]
    \centering
    \includegraphics[width=\linewidth]{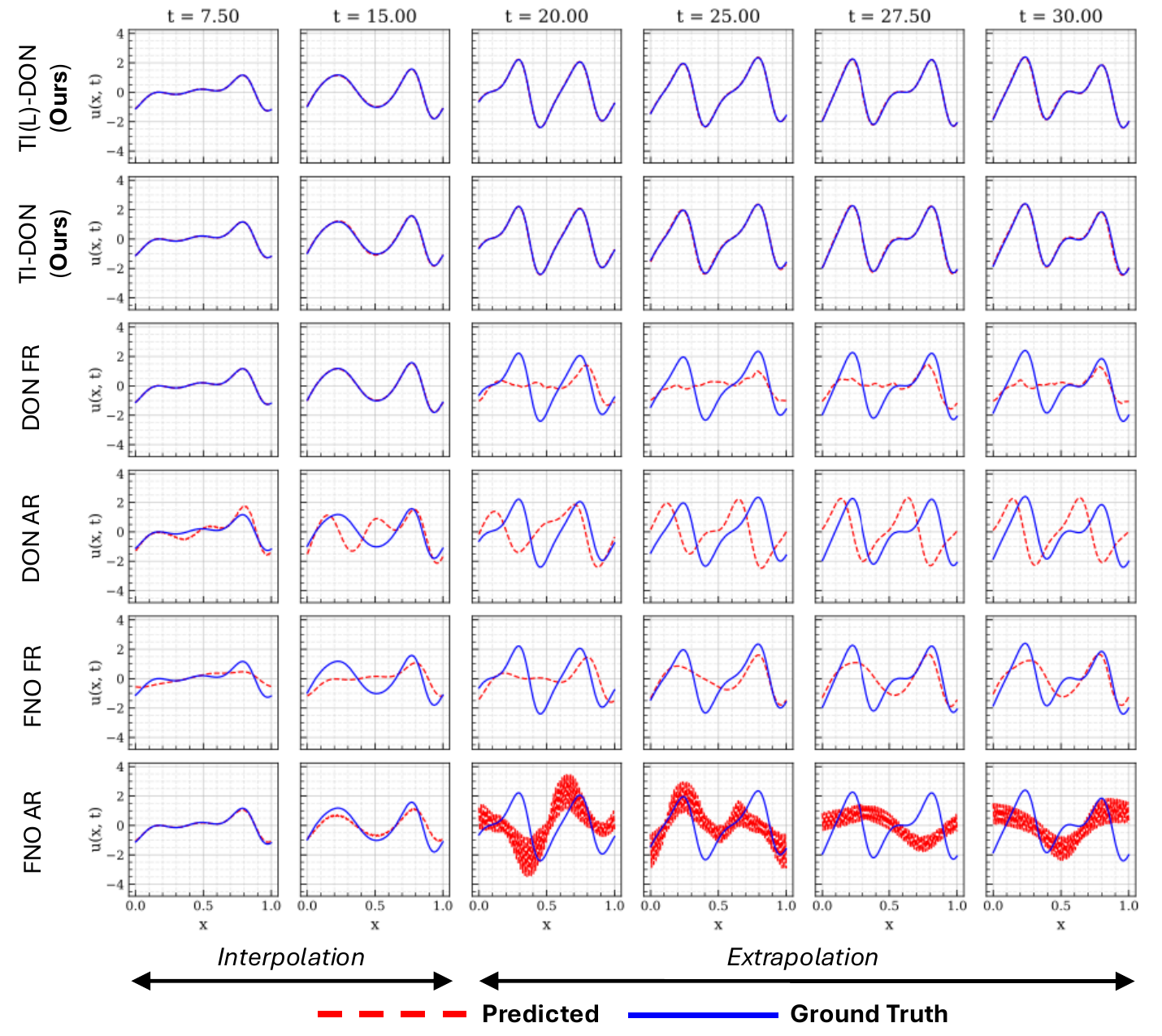}
    \caption{1D KS Equation: Performance of all the frameworks in the training ($t\in[0, 15]$) and the extrapolation regime ($t\in[15, 30]$) for a representative sample.}
    \label{fig:sample_lineplots_1d_ks}
\end{figure}

\begin{figure}[htb!]
    \centering
    \includegraphics[width=\linewidth]{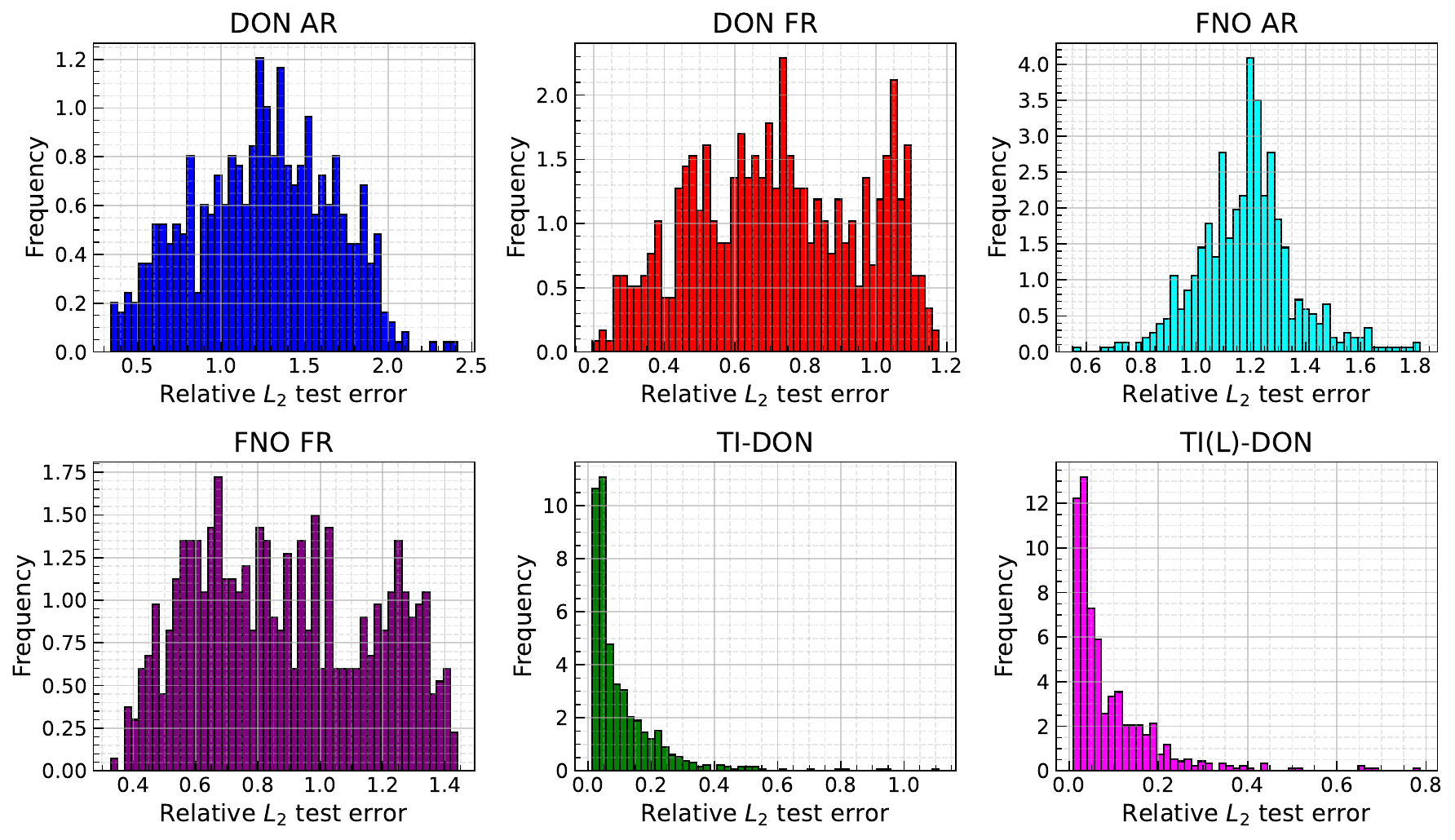}
    \caption{\revv{1D KS Equation: Distribution of relative $L_2$ errors across all test samples for all frameworks, averaged over the entire spatiotemporal domain.}}
    \label{fig:test_err_dist_1d_ks}
\end{figure}
Figure~\ref{fig:sample_lineplots_1d_ks} presents the prediction errors incurred by the \rev{TI-based neural operators} compared to standard baselines, shown for both the training and extrapolation temporal domains for a representative sample. In the training regime, DON FR and FNO AR accurately recover the solution profile, whereas DON AR and FNO FR exhibit relatively higher errors. However, in the extrapolation domain, several interesting behaviors emerge for the standard NO baselines. DON FR, DON AR, and FNO FR incur substantial errors once predictions extend beyond the training horizon, with these errors growing progressively as extrapolation time increases. Notably, FNO AR produces highly noisy and fluctuating solution profiles, indicative of spectral instability or spectral bias artifacts. This behavior manifests as high-frequency leakage, where certain unresolved or weakly controlled high-frequency modes become excessively amplified, resulting in pronounced oscillations in the predicted solution. This phenomenon closely resembles the instability observed in FNO AR predictions for the 1D KdV system. Interestingly, while such behavior was observed only for the KdV (dispersive) and KS (chaotic) systems but not for Burgers (dissipative), we also found that the training of FNO AR was inherently unstable for certain random weight initializations. This suggests that systems exhibiting dynamics beyond standard diffusion may not be well-suited for the FNO autoregressive learning paradigm. In contrast, the TI-based variants demonstrate remarkably higher predictive accuracy in both the training and extrapolation temporal domains. Once again, \rev{TI(L)-DON} performs slightly better in capturing the unseen chaotic dynamics of the KS equation, which can be attributed to its solution-specific local adaptivity during the learning process.
\begin{figure}[htb!]
    \centering
    \includegraphics[width=\linewidth]{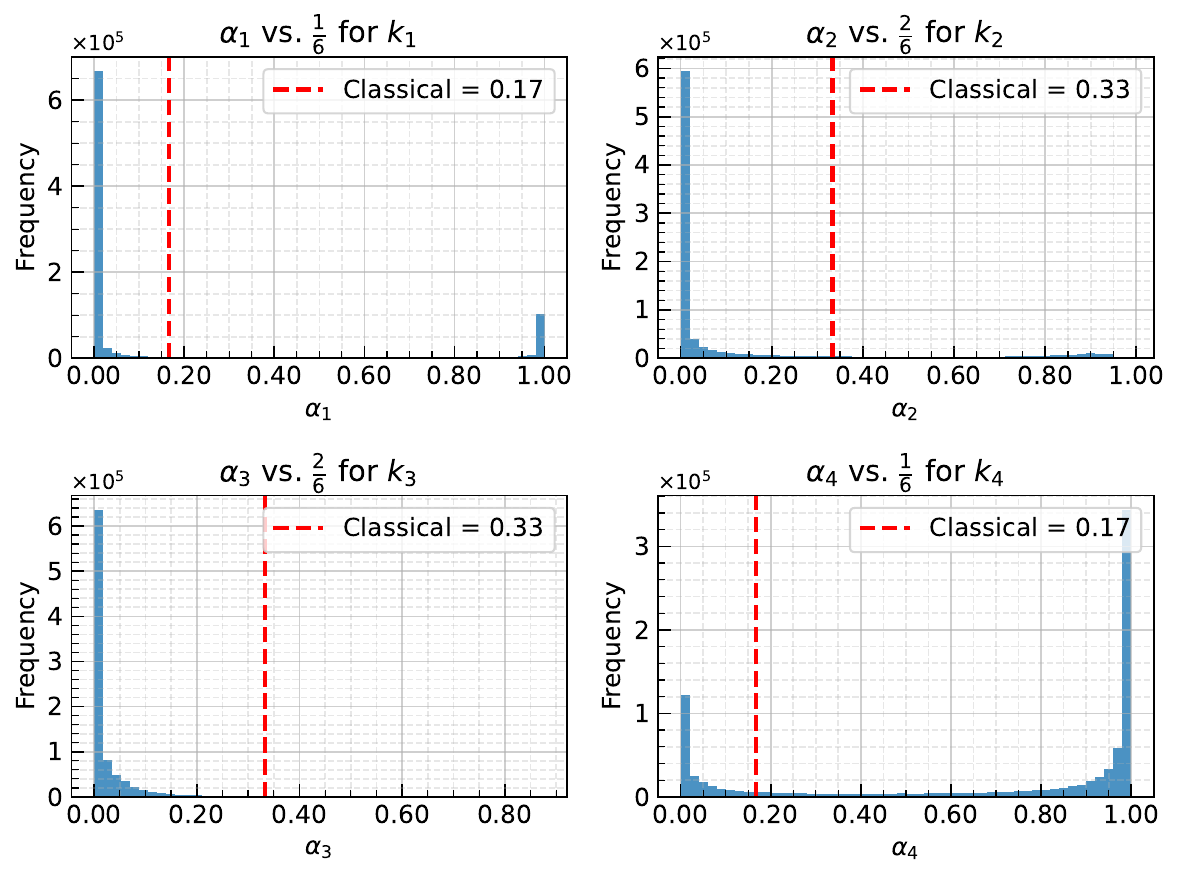}
    \caption{\rev{1D KS Equation: Distribution of the optimally learned $\alpha_i~\forall i \in \lbrace1,2,3,4\rbrace$ for the RK4 slope coefficients.}}
    \label{fig:alpha_dist_1d_ks}
\end{figure}

Figures~\ref{fig:error_accumulation}(e) and (f) show the relative $L_2$ error growth over time for all frameworks. The autoregressive models, DON AR and FNO AR, are highly susceptible to error compounding under recursive rollout. Consequently, they begin accumulating errors within the initial timesteps, with these errors escalating rapidly in the extrapolation domain to reach values as high as 140\% at the final timestep. FNO FR performs marginally better: while it still exhibits monotonically increasing error, the overall magnitude remains slightly lower, though still substantial in absolute terms. DON FR performs well in the training region due to DeepONet's strong interpolation capabilities; however, its accuracy deteriorates rapidly upon entering the extrapolation domain, where errors grow steeply to reach levels comparable to FNO FR (approximately 80\% at the final timestep). In stark contrast, the proposed TI-based variants leverage learned system dynamics by incorporating numerical time integrators within the learning framework. This coupling effectively suppresses error growth in the extrapolation region, even in the presence of chaotic dynamics. Consequently, these models not only achieve high accuracy in the \rev{training interval} but also maintain relatively stable and controlled error growth throughout the extrapolation interval $t \in [15, 30]$.
\rev{Between the TI-based variants, TI(L)-DON marginally outperforms TI-DON, as evidenced in Table~\ref{tab:problem_summary} and Figure~\ref{fig:error_accumulation}(e). The error distributions across test samples, shown in Figure~\ref{fig:test_err_dist_1d_ks}, remain largely similar for both variants, with TI(L)-DON exhibiting slightly better performance characterized by a higher density of samples with errors under 2\% compared to TI-DON. This indicates that the solution-specific local adaptivity of TI(L)-DON is effective in capturing the underlying stiff, chaotic dynamics of the KS equation.}
\revv{On the contrary, the test performance of other NO baselines consistent with the previous observations remains as: DON FR > FNO FR > FNO AR > DON AR.}
\rev{In a similar spirit to the previous cases, we present the distribution of the learned weighting coefficients $\alpha_1$, $\alpha_2$, $\alpha_3$, and $\alpha_4$ in Figure~\ref{fig:alpha_dist_1d_ks}. Consistent with earlier observations, $\alpha_4$ predominantly takes large values close to 1, indicating that greater emphasis is placed on the slope $k_4$ for a majority of solution states. However, unlike the previous cases, a subset of solution states exhibits lower values of $\alpha_4$ accompanied by higher values of $\alpha_1$, suggesting that for certain solution configurations, the network learns to place greater weight on the initial slope $k_1$. This variability in the learned coefficients reflects the complex, chaotic nature of the KS dynamics, where different solution states may benefit from different weighting strategies within the adaptive RK4 scheme.}

\subsection{\rev{Two-dimensional Burgers' Equation}}
\label{subsec:example4}
To assess performance on complex high-dimensional spatiotemporal dynamics, we consider the 2D Burgers' equation, defined as:
\begin{equation}
    \frac{\partial u}{\partial t} + u \frac{\partial u}{\partial x} + u \frac{\partial u}{\partial y} = \nu \left( \frac{\partial^2 u}{\partial x^2} + \frac{\partial^2 u}{\partial y^2} \right), \quad \forall ~ (x, y, t) \in [0, 1]^2 \times [0, 1],
    \label{eq:2d_burgers}
\end{equation}
where $u(x, y, t)$ is a scalar field and $\nu = 10^{-4}$ is the kinematic viscosity. The initial condition, $u(x, y, 0) = s(x, y)$, is sampled from 2D Gaussian random fields, with periodic boundary conditions applied in both spatial dimensions. \rev{Although the equation includes a finite viscosity term, its magnitude is sufficiently small ($\nu = 10^{-4}$) as a result of which the system approximately behaves as inviscid, with shock formation possible at longer temporal horizons.} This rich dynamical behavior, characterized by the development of sharp discontinuities and complex shock wave interactions, makes this problem an attractive benchmark for evaluating the performance of neural operator surrogates. \rev{The following subsections contain comprehensive details of our data generation approach, which builds upon the code developed by \cite{rosofsky2023applications}}.

\paragraph{\textbf{\rev{Data generation:}}}
\rev{
For the \rev{2D Burgers' equation}, we consider periodic boundary conditions on a unit square domain $(x, y) \in [0, 1]^2$. The initial conditions $u(x, y, 0) = s(x, y)$ are sampled from periodic Matérn-type Gaussian random fields with length scale $l = 0.1$ and standard deviation $\sigma = 0.2$. The spatial domain is discretized using a $64 \times 64$ grid with $N_x = N_y = 64$ points. The PDE is solved using a fourth-order Runge-Kutta scheme with timestep $\Delta t = 10^{-4}$ up to final time $t = 1.0$. Solutions are saved at intervals of $\Delta t_{\text{save}} = 0.01$, resulting in 101 temporal snapshots. The kinematic viscosity is set to $\nu = 10^{-4}$, which is sufficiently small to induce inviscid behavior while maintaining numerical stability. We generate a total of 1000 samples, each consisting of the initial condition and the corresponding spatiotemporal solution field $u(t, x, y)$ of size $101 \times 64 \times 64$.
}

\paragraph{\textbf{\rev{Data preparation:}}}
\rev{
The dataset is partitioned using a train-test split ratio of 0.8, yielding $N_{\text{train}} = 800$ training samples and $N_{\text{test}} = 200$ testing samples. To evaluate extrapolation capabilities, we train only on the first third of the temporal domain, i.e., $t_{\text{train}} \in [0, 0.33]$ (33 out of 101 time steps), and assess performance over the entire domain $t \in [0, 1]$. This represents one of the most challenging extrapolation scenarios in our study, requiring $3\times$ temporal extrapolation with no exposure to shock formation during training. For full rollout methods, our aim is to learn a solution operator $\mathcal{G}_{\boldsymbol{\theta}}$ that maps initial conditions to the full solution: $u_0(x,y) \rightarrow u(t, x, y)$. The trunk network receives $(t_{\text{train}}, x, y)$ coordinates as input, while the branch network processes the initial conditions. For FNO FR, the 3D Fourier transform is performed along all three dimensions $(t_{\text{train}}, x, y)$, capturing the full spatiotemporal frequency content within the training domain. In contrast, autoregressive and TI-based methods use only spatial coordinates $(x, y)$ for the trunk network input. Their branch networks receive stacked solution states $[u^0(x,y), u^1(x,y), \ldots, u^{33}(x,y)]$ from the training temporal window. FNO AR employs a 2D Fourier transform along the spatial dimensions $(x, y)$ only, consistent with its autoregressive nature to learn a one-step mapping between pairs of consecutive solution states $(u^i(x,y), u^{i+1}(x,y))$. Note that for the autoregressive/TI-based variants, explicit temporal information is absent and instead is assimilated in the batch dimension.
}

\paragraph{\textbf{\rev{Training architecture}}}
\rev{
Tables~\ref{tab:2d_burgers_don-architecture} and \ref{tab:2d_burgers_fno-architecture} presents the network architecture details, including the number of layers and neurons per layer, activation functions, total epochs to convergence, batch size and Fourier modes (if applicable). All DeepONet variants employ Fourier feature encoding in the trunk network. For DON FR, the spatial and temporal coordinates $(x, y, t)$ are augmented with sinusoidal features computed as $[\sin(\pi k x), \cos(\pi k x), \sin(\pi k y), \cos(\pi k y), \sin(\pi k t), \cos(\pi k t)]$ for $k \in \{1, 2, \ldots, 10\}$, expanding the input dimension from 3 to 63. For the autoregressive and TI-based variants, only the spatial coordinates $(x, y)$ are augmented with Fourier features $[\sin(\pi k x), \cos(\pi k x), \sin(\pi k y), \cos(\pi k y)]$, expanding the input dimension from 2 to 42. In this case, for the TI(L)-DON architecture, we employ a similar combination of 2D convolutional blocks and MLP layers as used in the branch network to learn the RK4 slope coefficients. The convolutional blocks utilize the same combination of Conv2D and pooling layers as the branch, with the only difference being that the Conv2D layers have a reduced number of filters (32 instead of 64). This is followed by a flattening layer, after which the intermediate representation is passed through two fully connected hidden layers and finally to an output layer with 4 neurons that predicts the RK4 slope coefficients. The Conv2D layers use a \texttt{ReLU} activation function, the fully connected hidden layers use \texttt{tanh}, and the output layer applies a \texttt{softmax} activation to ensure that the predicted slope coefficients sum to one. Training was performed using the Adam optimizer with an initial learning rate of $10^{-3}$, which was exponentially decayed by a factor of 0.95 every 5000 steps. The auxiliary network responsible for predicting the learnable RK4 slope coefficients was trained with a similar exponential decay schedule, but with a higher initial learning rate of $2 \times 10^{-3}$. For both FNO AR and FNO FR variants, we employ a similar exponential decay schedule as the DeepONet variants albeit with a decay factor of 0.96 every 2000 steps.
}
\begin{figure}[htb!]
    \centering
    \includegraphics[width=\linewidth]{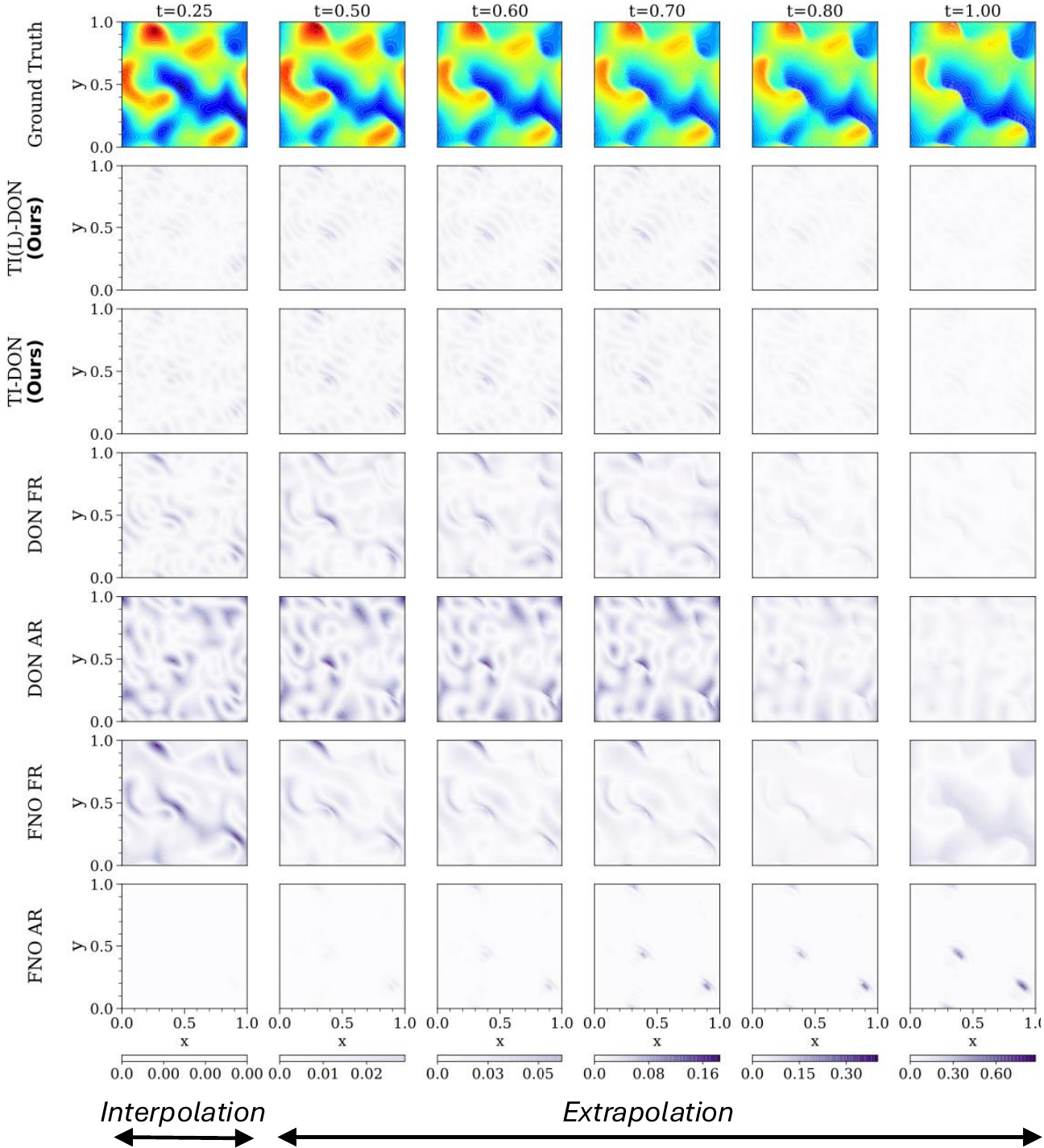}
    \caption{\rev{2D Burgers' Equation}: Spatial error distribution across training ($t\in[0, 0.33]$) and extrapolation ($t\in[0.33, 1]$) regimes for all frameworks, illustrated with a representative sample.}
    \label{fig:sample_contourplots_2d_burgers_error}
\end{figure}
\begin{figure}[htb!]
    \centering
    \includegraphics[width=\linewidth]{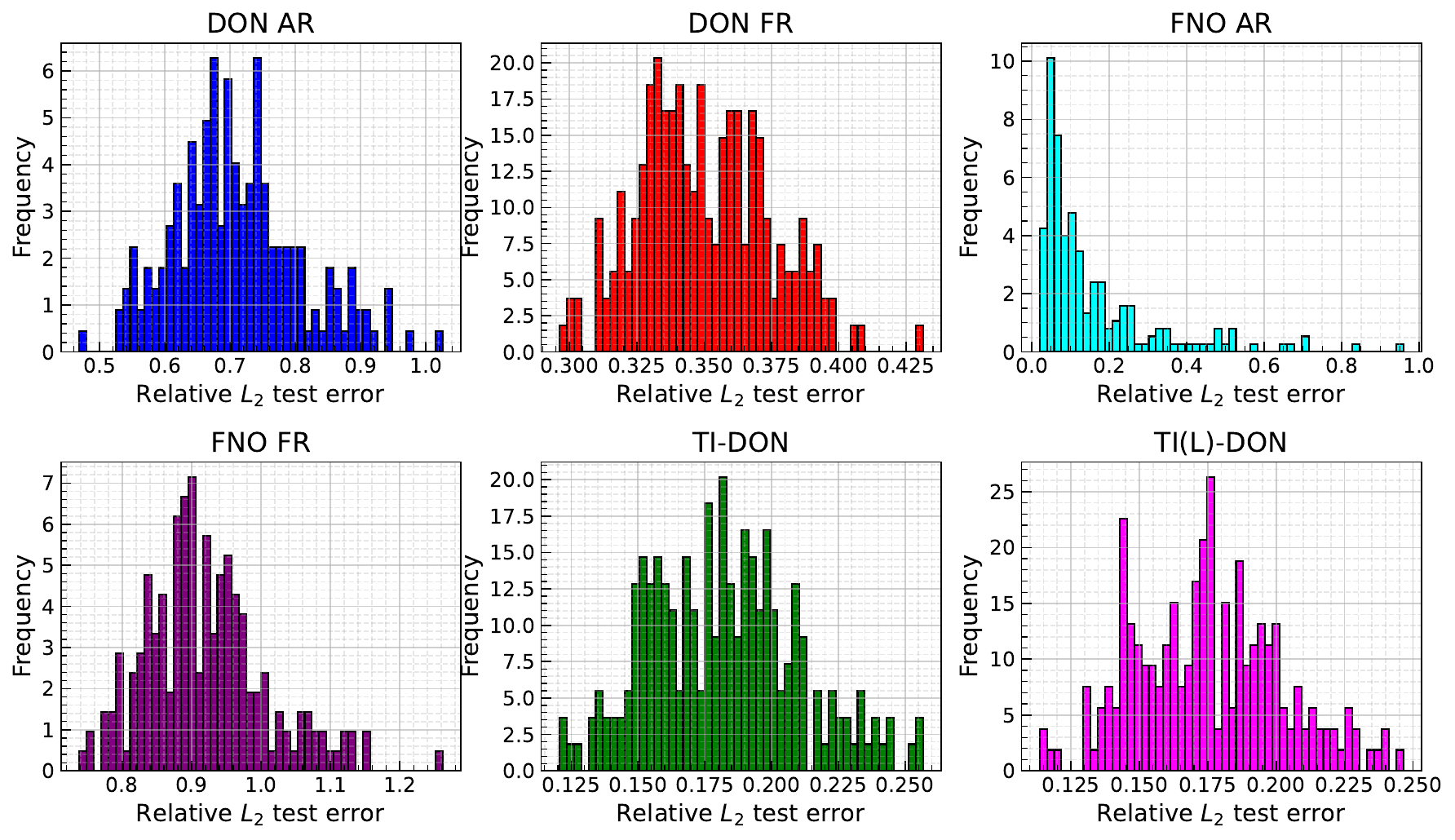}
    \caption{\revv{2D Burgers' Equation: Distribution of relative $L_2$ errors across all test samples for all frameworks, averaged over the entire spatiotemporal domain.}}
    \label{fig:test_err_dist_2d_burgers}
\end{figure}
\begin{figure}[htb!]
    \centering
    \includegraphics[width=\linewidth]{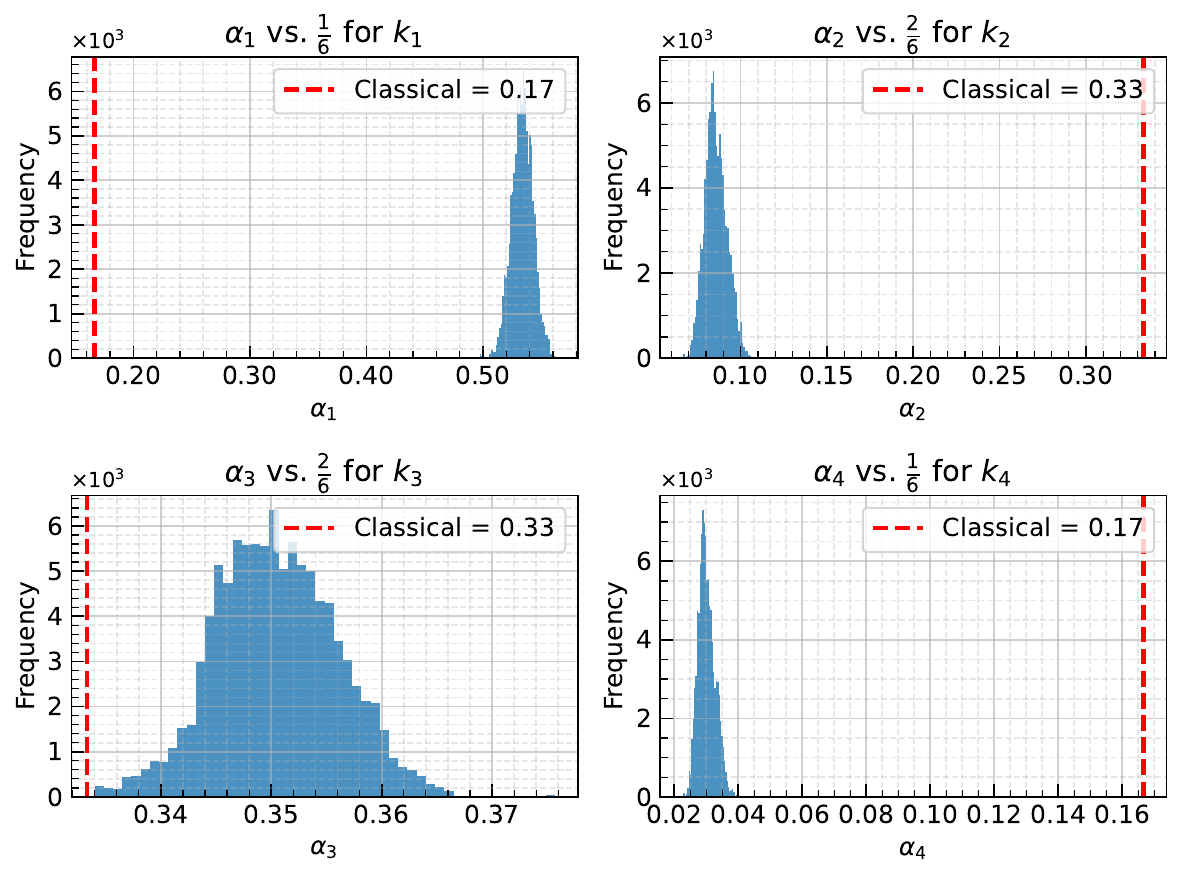}
    \caption{\rev{2D Burgers' Equation: Distribution of the optimally learned $\alpha_i~\forall i \in \lbrace1,2,3,4\rbrace$ for the RK4 slope coefficients.}}
    \label{fig:alpha_dist_2d_burgers}
\end{figure}

Figure~\ref{fig:sample_contourplots_2d_burgers_error} presents the spatial error plots in both training and extrapolation regimes, demonstrating that the time integrator models overall achieve higher accuracy than the baseline models. The solution plot for the same sample is presented in Figure~\ref{fig:sample_contourplots_2d_burgers}. Figures~\ref{fig:error_accumulation}(g) and (h) compare the relative $L_2$ error trends over the entire temporal domain. Notably, we reduced the training temporal domain to $t\in[0,0.33]$ with the extrapolation domain spanning $t\in[0.33,1.0]$, yielding \rev{one of} the most challenging case in this work; all networks must perform $3\times$ extrapolation with no prior knowledge of shock formation at longer temporal horizons (no training data contained shocks for any framework). \rev{DON AR} rapidly accumulates error, exceeding 130\% by the final timestep. \rev{DON FR} initially performs better, but its fixed-basis representation limits temporal generalization, with errors surpassing \rev{TI-DON} beyond $t=0.5$. Regarding the FNO variants, FNO FR fails catastrophically in capturing global system frequencies due to insufficient solution information across all timesteps, yielding completely unstable solution fields at longer temporal horizons with errors reaching 226\% at the final timestep. FNO AR performs significantly better than DON AR, DON FR, and FNO FR, even outperforming the TI-based variants within the training temporal domain. However, due to its inherent sensitivity to error compounding during autoregressive inference, its excellent training domain accuracy (as low as 8\% error) rapidly degrades in the extrapolation regime, reaching over 72\% error at the final timestep, surpassing DON FR, which saturates at 52\%. Consistent with previous examples, \rev{TI(L)-DON} maintains relatively slower and stable error growth, achieving final errors around 32\% and outperforming all baselines as well as marginally outperforming \rev{TI-DON} in the long-term prediction regime. 
\rev{A more detailed investigation is presented through the test error distribution plots in Figure~\ref{fig:test_err_dist_2d_burgers}, where the overall error distributions for TI-DON and TI(L)-DON show comparable performance on a per-sample basis.} 
 \revv{Among the baseline methods, FNO AR demonstrates largely comparable performance to the TI-variants, followed by DON FR, while FNO FR and DON AR exhibit similar but less favorable error distributions.}
These results demonstrate the robustness of time-integrator frameworks in modeling complex spatiotemporal dynamics involving stiff gradients and shock formation.
\rev{Finally, we present histograms of the optimally learned weighting coefficients $\alpha_i$ for $i \in \{1,2,3,4\}$ in the TI(L)-DON framework and compare them against the classical RK4 weighting coefficients in Figure~\ref{fig:alpha_dist_2d_burgers}. Interestingly, $\alpha_3$ exhibits values close to the classical RK4 coefficient of 0.33, while $\alpha_4$ and $\alpha_2$ lie within 0.04 and 0.1 respectively, indicating that they are less significant compared to the other coefficients for this problem. In contrast, $\alpha_1$ shows values greater than 0.5, substantially higher than the classical coefficient of 0.17. Overall, for the 2D Burgers' case, we observe that $\alpha_3$ and $\alpha_1$ are the dominant coefficients, with the latter bearing the most significance. These observations reinforce a key insight: the learned weighting coefficients are data-dependent and tailored to the underlying solution dynamics provided as input to the auxiliary RK4 network. This example reveals that for problems where shock formation is imminent, the use of such adaptive schemes is particularly advantageous, as they provide instantaneous solution-specific adaptivity that fixed-coefficient schemes cannot offer.}
\begin{figure}[htb!]
    \centering
    \begin{minipage}[b]{0.495\textwidth}
        \centering
        \includegraphics[width=\textwidth]{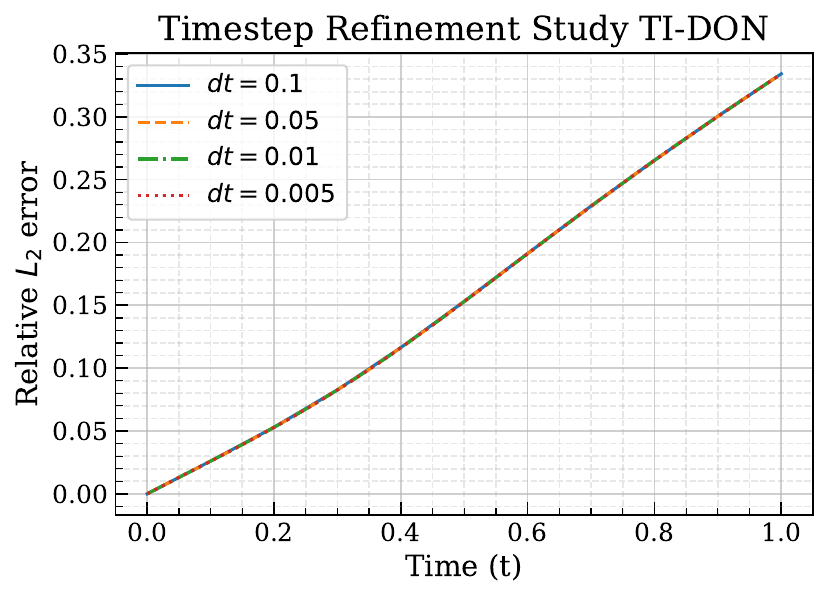}
    \end{minipage}
    \hfill
    \begin{minipage}[b]{0.495\textwidth}
        \centering
        \includegraphics[width=\textwidth]{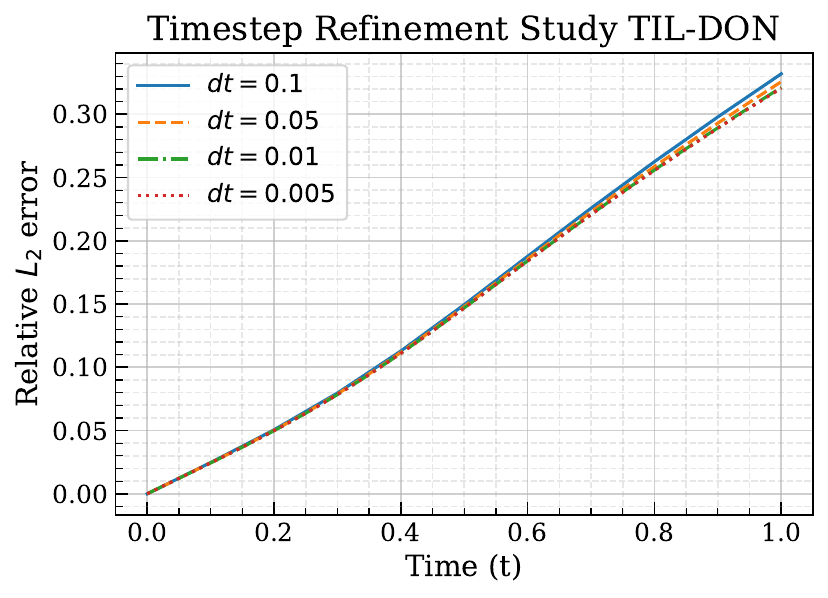}
    \end{minipage}
    \caption{\rev{2D Burgers' Equation: Timestep refinement study with different inference $\Delta t$ for the TI-based variants.}}
    \label{fig:timestep_refinement}
\end{figure}

\rev{
As described in Fig.~\ref{fig:DeepONet_timeintegrator}, the architecture of TI-DON allows for flexibility in the choice of different $\Delta t$ at training and inference. To this end, we perform a timestep refinement study for the 2D Burgers' equation, where the model was trained with $\Delta t = 0.01$ and inference was performed using $\Delta t \in \{0.1, 0.05, 0.01, 0.005\}$. The relative $L_2$ error growth is presented in Fig.~\ref{fig:timestep_refinement} for both TI-DON and TI(L)-DON with different inference $\Delta t$. For TI-DON, prediction errors remain consistent across all $\Delta t$ values, demonstrating timestep-invariant behavior and confirming that error growth stems from model approximation rather than numerical instability. For TI(L)-DON, finer temporal resolutions ($\Delta t \in \{0.01, 0.005\}$) yield marginally reduced errors at extended horizons, indicating that this method potentially benefits from smaller timesteps.
}

\subsection{\rev{Two-dimensional Scalar Rotation Advection-Diffusion}}
\label{subsec:example5}
\rev{
The fifth example that we consider is the scalar \revv{advection–diffusion} equation with a rotating advection field~\revv{as investigated in \cite{de2023explicit}}. The governing PDE reads as:
\begin{equation}
\begin{aligned}
    \dot{u}_i - \nabla \cdot F_i(u_i) &= f_i & &\text{on } \Omega_i \times [0, T], \\
    u_i &= g_i & &\text{on } d\Omega_i \times [0, T]~~i = 1, 2, \\
    u_i &= u_{i, 0} & &\text{on } \Omega_i,~t=0
\end{aligned}
\end{equation}
where $\dot{u_i}$ refers to the instantaneous time-derivative of the scalar field $u(\mathbf{x}, t)$ at the $i^{th}$ state, $F_i(u_i) = \kappa_i \nabla u_i - \mathbf{a}u_i$ is the total flux function, $f_i = f_i(\mathbf{x}, t)$ is a source term, $g_i = g_i (\mathbf{x}, t)$ is the prescribed boundary condition, $u_{i,0} = u(\mathbf{x}, t=0)$ is the defined initial condition, $\kappa_i = \kappa_i(\mathbf{x}, t) > 0$ is the diffusion coefficient in the spatial domain $\Omega_i$, and $\mathbf{a} = \mathbf{a}(\mathbf{x}, t)$ is the rotating advection field.
}

\paragraph{\textbf{\rev{Data generation and preparation:}}}
\rev{
For this study, we consider the computational domain \revv{$\Omega = (0, 1)^2$} discretized using a uniform $64 \times 64$ quadrilateral mesh. The rotating advection field is defined as $\mathbf{a} = (0.5 - y, x - 0.5)$, which induces rigid body rotation about the domain center. A constant diffusion coefficient $\kappa = 0.001$ is employed, with homogeneous Dirichlet boundary conditions prescribed on all boundaries. The initial conditions consist of randomly parameterized Gaussian blobs with amplitude $A$\revv{, center coordinates $(x_0, y_0)$, and standard deviations $\sigma_1, \sigma_2$, all uniformly sampled from the intervals} $A \in [0.3, 1.0]$, $(x_0, y_0) \in [0.2, 0.8]^2$, and $\sigma_1, \sigma_2 \in [0.03, 0.10]$\revv{, respectively}. The governing equations are solved using a streamline upwind Petrov-Galerkin stabilized finite element method (SUPG) with Crank-Nicolson time integration. The dataset comprises 800 samples, each simulated over a quarter rotation period ($T = \pi/2$), with solution snapshots saved at regular intervals of $\Delta t \approx 0.0067$, resulting in a total of 101 timesteps. For further details regarding the problem formulation, the reader is referred to~\cite{de2023explicit}. The dataset is partitioned using a train-test split ratio of 0.8, yielding $N_{\text{train}} = 640$ training samples and $N_{\text{test}} = 160$ testing samples. The dataset arrangement for full rollout, autoregressive, and TI-based methods follows the same protocol as described in Section~\ref{subsec:example4}, with the training temporal domain $t_{\text{train}} \in [0, 0.27]$ (first 40 out of 101 timesteps) and extrapolation domain $t \in [0, 0.67]$.
}

\paragraph{\textbf{\rev{Training architecture:}}}
\rev{
Tables~\ref{tab:2d_rotating_advection_diffusion_don-architecture} and \ref{tab:2d_rotating_advection_diffusion_fno-architecture} presents the network architecture details, including the number of layers and neurons per layer, activation functions, total epochs to convergence, batch size and Fourier modes (if applicable). All DeepONet variants employ Fourier feature encoding in the trunk network. For the DON FR, the spatial and temporal coordinates $(x, y, t)$ are augmented with sinusoidal features computed as $[\sin(\pi k x), \cos(\pi k x), \sin(\pi k y), \cos(\pi k y), \sin(\pi k t), \cos(\pi k t)]$ for $k \in \{1, 2, \ldots, 10\}$, expanding the input dimension from 3 to 63. For the autoregressive and TI-based variants, only the spatial coordinates $(x, y)$ are augmented with Fourier features $[\sin(\pi k x), \cos(\pi k x), \sin(\pi k y), \cos(\pi k y)]$, expanding the input dimension from 2 to 42. In this case, for the TI(L)-DeepONet architecture, we employ a similar combination of 2D convolutional blocks and MLP layers as used in the branch network to learn the RK4 slope coefficients. The convolutional blocks utilize the same combination of Conv2D and pooling layers as the branch, with the only difference being that the Conv2D layers have a reduced number of filters (32 instead of 64). This is followed by a flattening layer, after which the intermediate representation is passed through two fully connected hidden layers and finally to an output layer with 4 neurons that predicts the RK4 slope coefficients. The Conv2D layers use a \texttt{GELU} activation function, the fully connected hidden layers use \texttt{tanh}, and the output layer applies a \texttt{softmax} activation to ensure that the predicted slope coefficients sum to one. Training was performed using the Adam optimizer with an initial learning rate of $10^{-3}$, which was exponentially decayed by a factor of 0.95 every 5000 steps. The auxiliary network responsible for predicting the learnable RK4 slope coefficients was trained with a similar exponential decay schedule, but with a higher initial learning rate of $4 \times 10^{-3}$. For both FNO AR and FNO FR variants, we employ a similar exponential decay schedule as the DeepONet variants albeit with a decay factor of 0.96 every 2000 steps.
}
\begin{figure}[htb!]
    \centering
    \includegraphics[width=\linewidth]{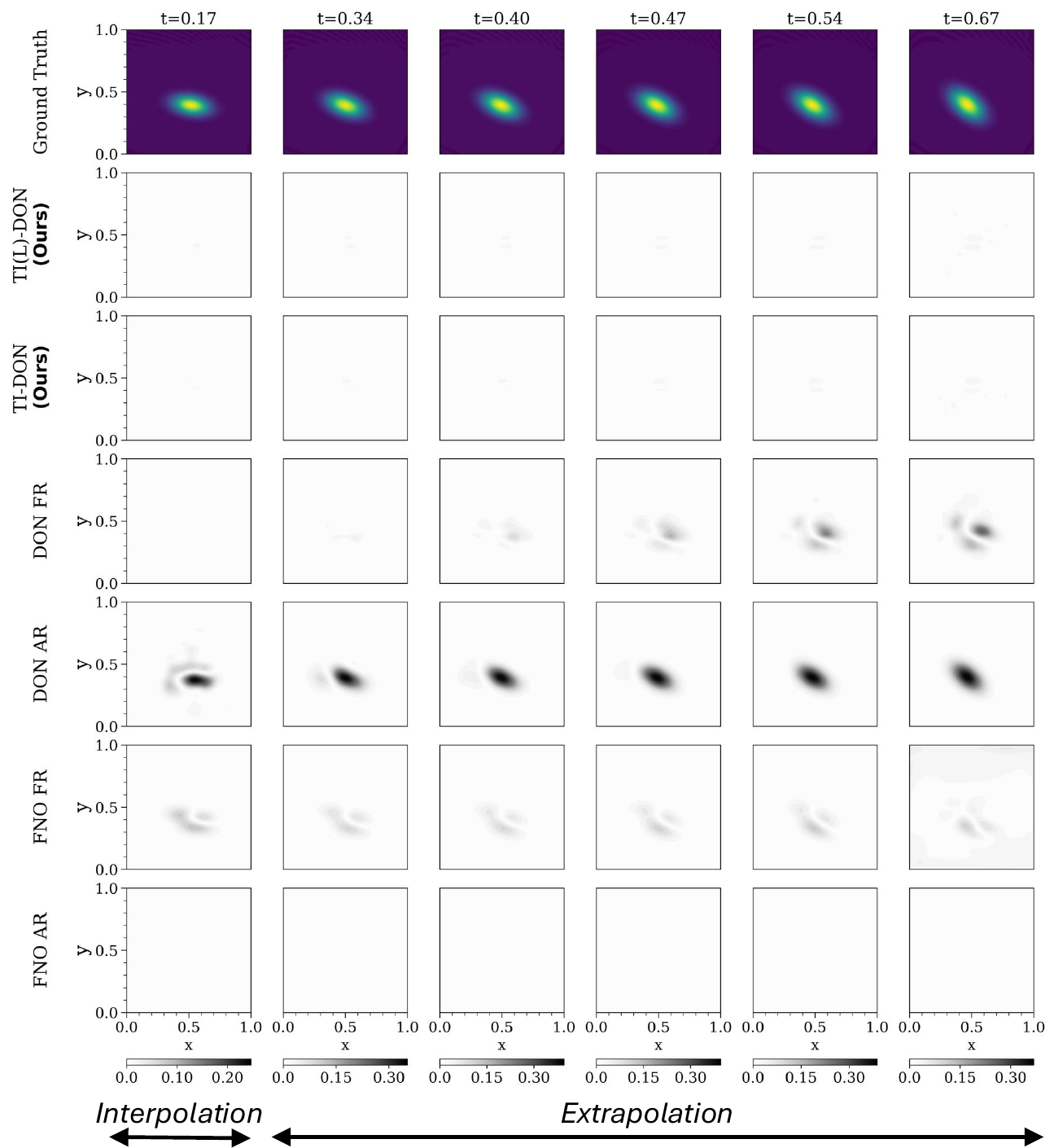}
    \caption{\rev{2D Scalar Rotation Advection-Diffusion}: \rev{Spatial error distribution across training ($t\in[0, 0.27]$) and extrapolation ($t\in[0.27, 0.67]$) regimes for all frameworks, illustrated with a representative sample.}}
\label{fig:sample_contourplots_2d_rotating_advection_diffusion_error}
\end{figure}
\begin{figure}[htb!]
    \centering
    \includegraphics[width=\linewidth]{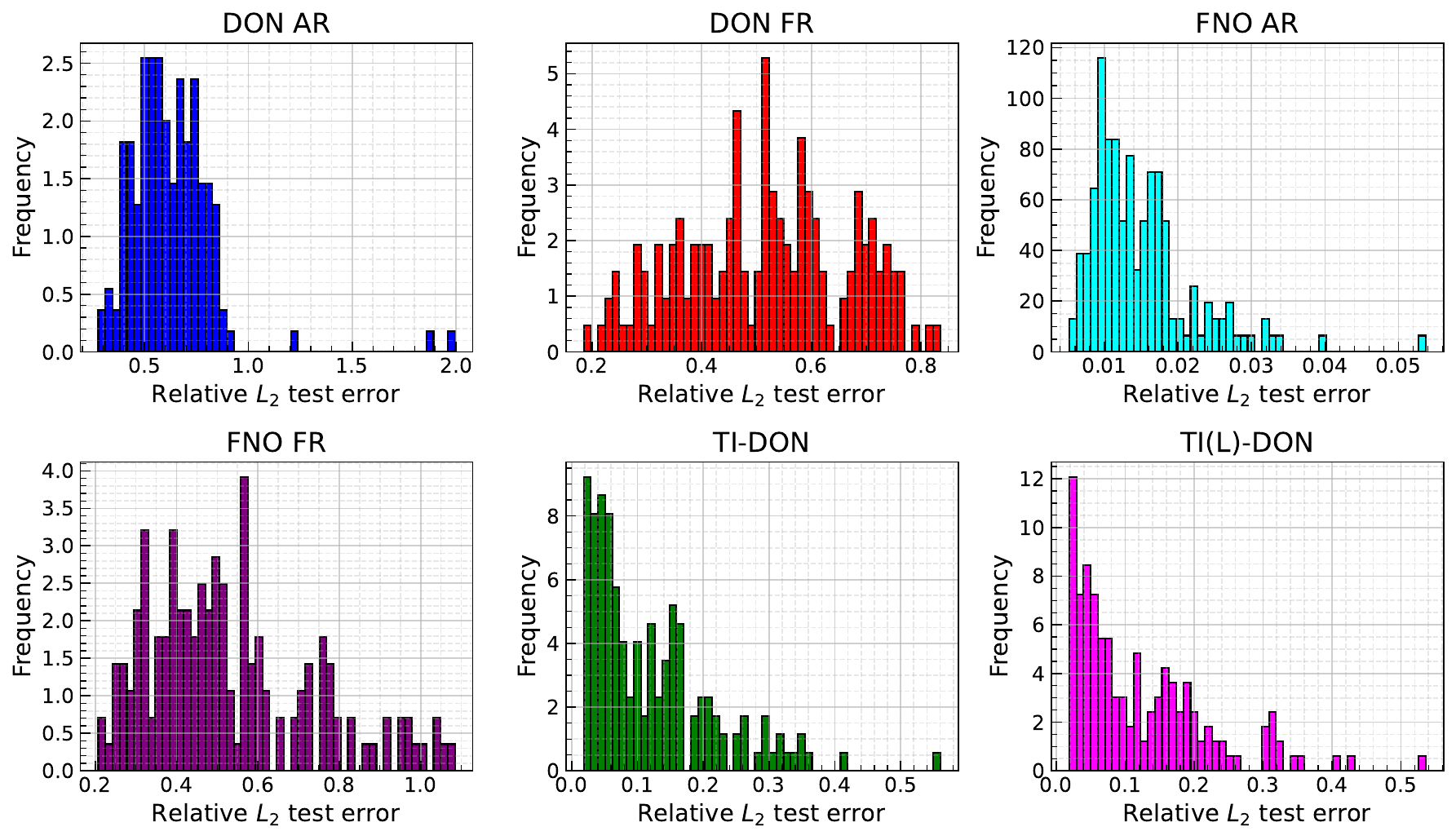}
    \caption{\revv{2D Scalar Rotation Advection-Diffusion: Distribution of relative $L_2$ errors across all test samples for all frameworks, averaged over the entire spatiotemporal domain.}}
\label{fig:test_err_dist_2d_rotation_advection_diffusion}
\end{figure}

\rev{
Following a similar analysis as the previous cases, a qualitative comparison of the performance of the various frameworks is presented through the spatial error distribution plots in the interpolation ($t \in [0, 0.27]$) and extrapolation ($t \in [0.27, 0.67]$) regimes in Figure~\ref{fig:sample_contourplots_2d_rotating_advection_diffusion_error}. DON AR incurs the highest errors due to its greater sensitivity to error accumulation in both the interpolation and extrapolation domains, thereby starting to accumulate errors early in the temporal evolution. This is followed by FNO FR, which, consistent with previous cases, suffers from incomplete spectral representation due to computing Fourier features over a limited temporal domain, and therefore incurs errors in both the interpolation and, more prominently, the extrapolation regime. Next is DON FR, which exhibits impressive accuracy during interpolation since its learned basis is tailored to the training temporal domain, but immediately breaks down upon entering the extrapolation regime. The TI-based variants follow, exhibiting much more stable error growth and demonstrating excellent performance not only in interpolation but also in extrapolation, owing to the physics prior induced by numerical time integration. However, among all baselines, FNO AR emerges as the most dominant method, exhibiting errors an order of magnitude lower than other baselines ($<4\%$), while the second-best error is approximately $23.72\%$ achieved by TI(L)-DON at the final timestep. The results presented in Table~\ref{tab:problem_summary} and Figure~\ref{fig:error_accumulation} are in strong agreement with these qualitative trends. The error growth of DON AR is, as expected, monotonic due to compounding of model approximation errors during recursive inference and surpasses FNO FR at $t=0.13$. FNO FR incurs a non-monotonic yet increasing error growth throughout the temporal domain, following DON AR, with both methods achieving $L_2$ errors above 80-97\% at the final timestep. DON FR performs well in interpolation as highlighted above and achieves comparable performance to TI-DON for $t \in [0.13, 0.27]$, but rapidly degrades upon entering the extrapolation domain, ultimately reaching errors over 105\% and surpassing all other baselines in performance degradation. Compared to DON FR, DON AR, and FNO FR, TI-DON's inherent stabilization of error growth leads to roughly 23.7\% error at the final timestep, which is eventually dominated by the impressive performance of FNO AR, which exhibits a plateau in error growth during the later timesteps in the extrapolation domain. As evidenced in Figure~\ref{fig:error_accumulation}(j) and Figure~\ref{fig:test_err_dist_2d_rotation_advection_diffusion}, there is an insignificant difference in the error growth as well as overall error distributions between TI-DON and TI(L)-DON, with both exhibiting similar performance. Therefore, for brevity, we do not analyze the distribution of the learned $\boldsymbol{\alpha}$ coefficients for the intermediate RK4 slopes in the TI(L)-DON framework for this case.
}
\revv{Moreover, the evidently superior performance of FNO AR over all the other NO baselines is solidified, with errors one order of magnitude less than the TI-based variants. The rest of the ranking is as follows: FNO FR $\approx$ DON FR $>$ DON AR.}
\rev{
This example elucidates two key conclusions. First, the rotation advection-diffusion problem poses a more challenging test case compared to the previous examples. Given the negligible diffusion coefficient ($\kappa = 0.001$), the problem is strongly advection-dominated, and the Gaussian blob should undergo near-rigid body rotation with its shape preserved throughout the temporal evolution. However, the TI-based variants incur noticeable errors in capturing this rotational dynamics. Specifically, the sequential predictions distort the boundaries of the Gaussian blob, violating the shape-preserving constraint inherent to rigid body rotation. This artifact accumulates over successive time integrations, leading to progressive deformation of the blob structure. We attribute this to the sensitivity of explicit time integration schemes to the choice of $\Delta t$ when resolving rotational motion; a finer temporal resolution may be necessary to minimize boundary distortions and achieve performance comparable to FNO AR. The superior performance of FNO AR in this case can be attributed to the global Fourier basis, which naturally captures the periodic structure inherent in rotational dynamics without requiring explicit temporal marching, thereby avoiding the accumulation of shape-distorting artifacts. Second, this example highlights that for certain dynamical systems, the choice of neural operator architecture may be as important as the time integration strategy. This motivates a more general TI-NO formulation (time-integrator embedded neural operator), wherein the neural operator and time integration scheme can be selected independently based on the problem characteristics. We hypothesize that a TI-FNO variant, which combines the time-integrator framework with FNO's spectral representation, could leverage the strengths of both approaches for problems involving rotational or periodic dynamics. Such an extension remains beyond the scope of the present work and is deferred to future investigation.
}

\subsection{\rev{Three-dimensional Heat Conduction}}
\label{subsec:example6}
\rev{
We consider three-dimensional heat conduction as the final example in this study. The heat equation is a linear parabolic PDE with a wide range of applications: it naturally arises in modeling heat transfer through a medium, appears in probability theory as the governing equation for Brownian motion, plays a fundamental role in mathematical finance through option pricing models, and is widely used in image processing for smoothing and denoising. In a three-dimensional domain, the governing equation is given by:
\begin{equation}
    \frac{\partial T}{\partial t} = \alpha\left(
    \frac{\partial^2 T}{\partial x^2} +
    \frac{\partial^2 T}{\partial y^2} +
    \frac{\partial^2 T}{\partial z^2}
    \right), \quad \forall \ (x,y,z,t) \in [0,1]^3 \times [0,1],
\end{equation}
where $\alpha = 1.0$ denotes the thermal diffusivity. Here, $T$ represents the scalar temperature field with initial condition $T_0$.} 

\paragraph{\textbf{\rev{Data generation and preparation:}}}
\rev{
The spatial domain is a three-dimensional L-shaped region constructed by removing the upper-right quadrant (where $x \geq L_x/2$ and $y \geq L_y/2$) from a rectangular block, with the L-shaped cross-section oriented in the XY-plane. The 3D heat conduction equation is solved using an explicit FDM solver on a $32 \times 32 \times 16$ Cartesian grid over the spatial domain $(x, y, z) \in [0, 1]^3$. The Laplacian term on the RHS is discretized using a second-order central difference (CD2) scheme. Based on von Neumann stability analysis, a timestep of $\Delta t = 0.01$ is chosen, and the solution is advanced in time up to $t = 1.0$. The initial condition consists of a Gaussian blob with standard deviation $\sigma = 4.0$ grid points.} \revv{The amplitude of the Gaussian is uniformly sampled from the interval $[0, 1]$. The center of the Gaussian $(x_c, y_c, z_c)$ is uniformly sampled within the lower-left block of the L-shaped domain, with a margin of $\sigma$ from all boundaries to ensure the blob remains fully contained. Specifically, the center coordinates are sampled as $(x_c, y_c, z_c) \in [4, 11]^3$ in grid indices.} \rev{Homogeneous Dirichlet boundary conditions ($T = 0$) are imposed on all faces of the L-shaped block. A total of 1000 samples are generated with an 80-20 train-test split. The training temporal domain spans $t \in [0, 0.33]$, while the extrapolation domain covers $t \in [0.33, 1.0]$. The data preparation strategy follows the same protocol as the previous PDE examples for all neural operator baselines. For further details regarding the data generation procedure, the reader is referred to~\cite{roy2025best}, from which the data generation code is adapted.}

\paragraph{\textbf{\rev{Training architecture}}}
\rev{
Table~\ref{tab:3d_heat_don-architecture} presents the network architecture details, including the number of layers and neurons per layer, activation functions, total epochs to convergence, batch size, and Fourier modes (if applicable). Unlike previous cases, the trunk network is not equipped with Fourier feature encodings, as sufficiently accurate convergence was observed without them. For the TI(L)-DON architecture, the auxiliary network for predicting the RK4 weighting coefficients employs Conv3D layers with 32 filters, analogous to those in the branch network. This is followed by a flattening layer, after which the intermediate representation is passed through two fully connected hidden layers and finally to an output layer with 4 neurons that predicts the RK4 weighting coefficients. The Conv3D layers use a \texttt{GELU} activation function, the fully connected hidden layers use \texttt{tanh}, and the output layer applies a \texttt{softmax} activation to ensure that the predicted coefficients sum to one. Training was performed using the Adam optimizer with an initial learning rate of $10^{-3}$, which was exponentially decayed by a factor of 0.96 every 2000 steps. The auxiliary network was trained with a similar exponential decay schedule but with a higher initial learning rate of $5 \times 10^{-3}$.
}
\begin{figure}[htb!]
    \centering
    \includegraphics[width=\linewidth]{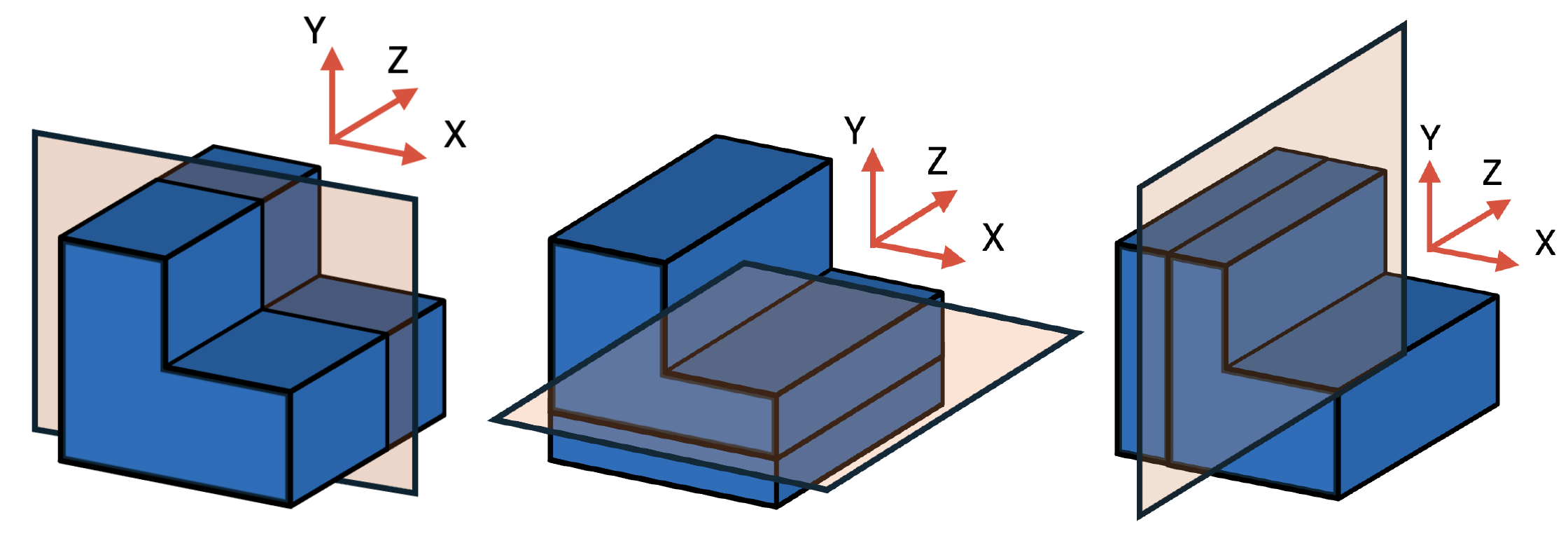}
   \caption{\rev{3D Heat Conduction: Schematic showing the XY, YZ, and XZ slicing planes (from left to right) for visualizing the error and solution profiles.}}
    \label{fig:3d_heat_slicing_planes}
\end{figure}
\begin{figure}[htb!]
    \centering
    \includegraphics[width=\linewidth]{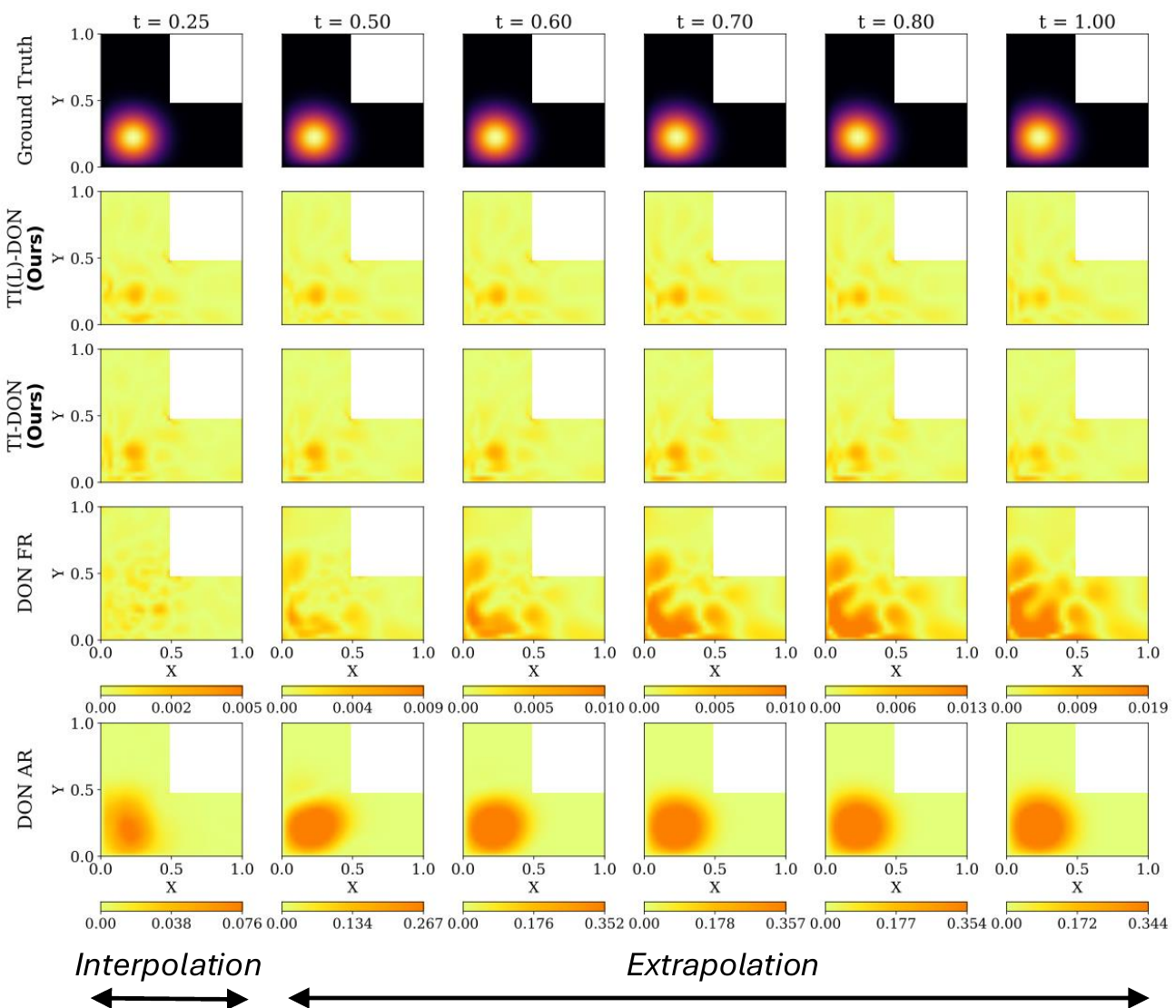}
   \caption{\rev{3D Heat Conduction: Spatial error distribution across training ($t \in [0, 0.33]$) and extrapolation ($t \in [0.33, 1]$) regimes for all frameworks, illustrated using a representative sample. The contours are presented along the XY slicing plane (see Fig.~\ref{fig:3d_heat_slicing_planes} for details regarding the location of the XY slice).}}
    \label{fig:3d_heat_error_contours_XY}
\end{figure}
\begin{figure}[htb!]
    \centering
    \includegraphics[width=\linewidth]{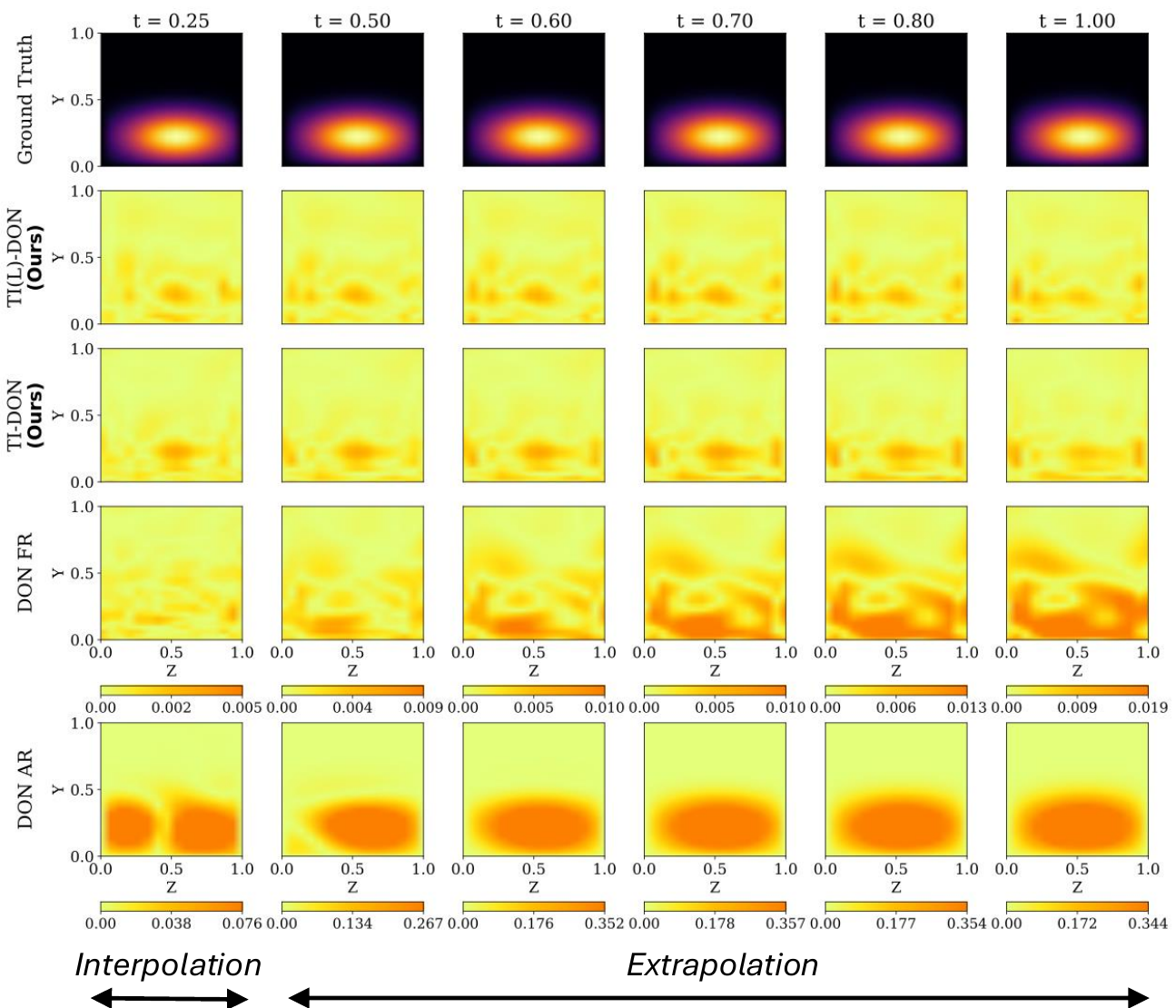}
    \caption{\rev{3D Heat Conduction: Spatial error distribution across training ($t \in [0, 0.33]$) and extrapolation ($t \in [0.33, 1]$) regimes for all frameworks, illustrated using a representative sample. The contours are presented along the YZ slicing plane (see Fig.~\ref{fig:3d_heat_slicing_planes} for details regarding the location of the YZ slice).}}
    \label{fig:3d_heat_error_contours_YZ}
\end{figure}
\begin{figure}[htb!]
    \centering
    \includegraphics[width=\linewidth]{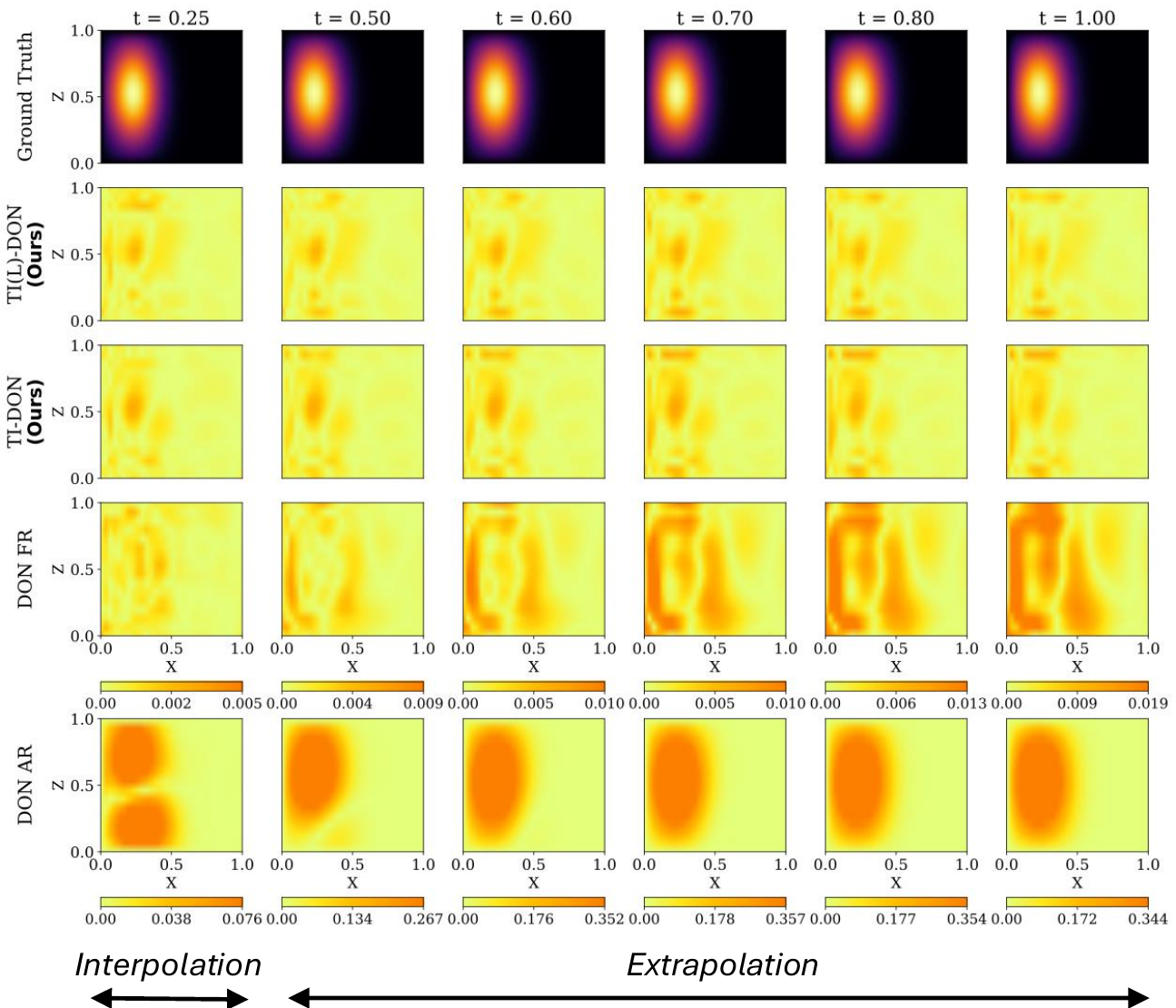}
    \caption{\rev{3D Heat Conduction: Spatial error distribution across training ($t \in [0, 0.33]$) and extrapolation ($t \in [0.33, 1]$) regimes for all frameworks, illustrated using a representative sample. The contours are presented along the XZ slicing plane (see Fig.~\ref{fig:3d_heat_slicing_planes} for details regarding the location of the XZ slice).}}
    \label{fig:3d_heat_error_contours_XZ}
\end{figure}
\begin{figure}[htb!]
    \centering
    \includegraphics[width=\linewidth]{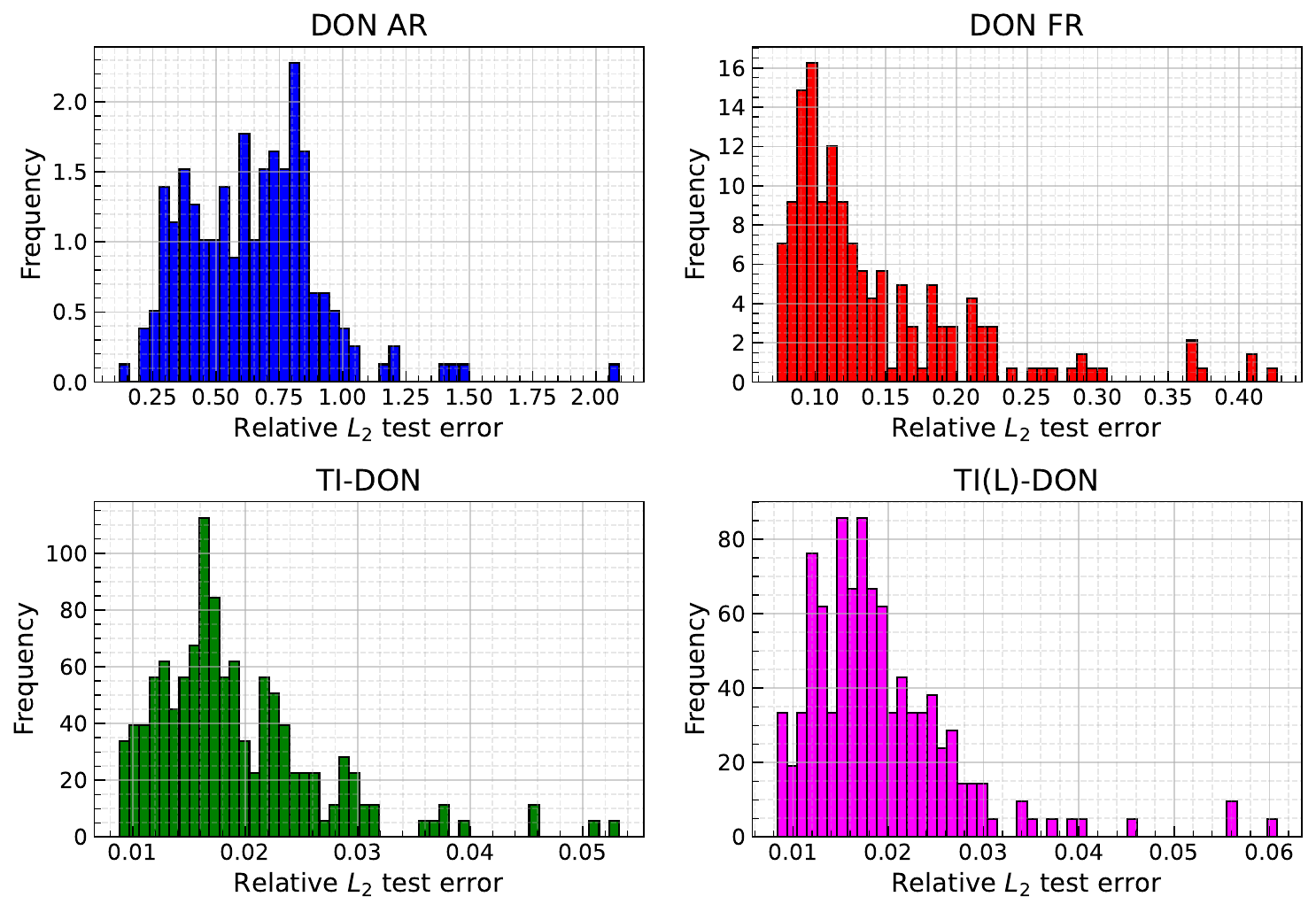}
    \caption{\revv{3D Heat Conduction: Distribution of relative $L_2$ errors across all test samples for all frameworks, averaged over the entire spatiotemporal domain.}}
\label{fig:test_err_dist_3d_heat_conduction}
\end{figure}

\rev{
Figures~\ref{fig:3d_heat_error_contours_XY}, \ref{fig:3d_heat_error_contours_YZ}, and \ref{fig:3d_heat_error_contours_XZ} present the error contours obtained for the DeepONet-based variants, i.e., TI(L)-DON, TI-DON, DON FR, and DON AR, in the interpolation ($t \in [0, 0.33]$) and extrapolation ($t \in [0.33, 1.0]$) regimes for a representative test sample, visualized along different slicing planes: XY, YZ, and XZ (in that order). FNO-based baselines are not shown here, as they cannot be directly applied to irregular domains such as the L-shaped domain considered in this study. As expected, DON AR exhibits an order of magnitude higher error due to its increased sensitivity to error accumulation. During sequential predictions, the compounding of model approximation errors causes DON AR to ``artificially'' dissipate the Gaussian blob too rapidly; consequently, the blob is visibly absent from the initial few timesteps itself (see Figures~\ref{fig:sample_contourplots_3d_heat_XY}, \ref{fig:sample_contourplots_3d_heat_YZ}, and \ref{fig:sample_contourplots_3d_heat_XZ}). This is superseded by the performance of DON FR, which demonstrates high accuracy in the interpolation regime but quickly deteriorates in the extrapolation regime due to its limited fixed-basis representation capacity tailored to the training temporal domain. In contrast, the TI-based variants consistently achieve superior performance in both the interpolation and extrapolation regimes by mitigating error accumulation through the physics-aware numerical time integration module. These qualitative trends are in good agreement with the quantitative results of relative $L_2$ error growth during inference, as shown in Table~\ref{tab:problem_summary} and Figure~\ref{fig:error_accumulation}. Notably, compared to DON AR and DON FR, which incur relative $L_2$ errors as high as 0.98 and 0.22 at the final timestep respectively, the reduction in error achieved by the TI-based variants is remarkable; approximately 96.3\% and 83.6\% improvement with respect to their autoregressive and full rollout counterparts. This example serves as a benchmark demonstrating that, especially in high-dimensional spatiotemporal dynamics (3D in this case), where spatial errors along three dimensions can compound uncontrollably, TI-based variants provide a pathway for mitigating error growth and stabilizing extrapolation owing to the strong physics-aware inductive bias imposed by the numerical time integration module. Unlike autoregressive DON, the diffusive characteristics of the heat conduction equation are extremely well captured by the TI-based variants, with TI(L)-DON slightly outperforming TI-DON, as evidenced in Figure~\ref{fig:test_err_dist_3d_heat_conduction}. \revv{This is followed by the performance of DON FR and then DON AR.}
}

\section{\rev{Computational Costs}}
\label{sec:compute_cost}
\rev{
The improved extrapolation accuracy offered by the proposed TI-DeepONet and TI(L)-DeepONet approaches comes at the cost of increased training and inference times. This additional computational overhead stems from the repeated forward and backward passes through the time integrator. However, the convergence of TI-based networks is significantly faster, indicating more efficient optimization of network parameters compared to baseline methods. Table~\ref{tab:training-inference-times} presents a detailed comparison of the training and inference costs for each method across all benchmark problems.  Moreover, incorporating adaptive, learnable slope coefficients facilitates convergence to the true solution even faster due to the solution-specific local adaptivity within the time integrator. Training and inference for the 1D Burgers', 1D KdV, and 1D KS cases were conducted on a single NVIDIA A100 GPU equipped with 40GB memory (Intel Xeon Gold Cascade Lake). The 2D Burgers', 2D Rotation Advection-Diffusion, and 3D Heat Conduction cases were performed on a NVIDIA A100 GPU with 80GB memory (Intel Xeon Gold Icy Lake).
}
\begin{table}[htb!]
     \renewcommand{\arraystretch}{1.15}
    \centering
    \caption{Computational costs of training and inference for all examples across different frameworks. \rev{Note that training speed is reported as iterations per second, where a lower value indicates slower training and higher computational cost.}}
    \begin{tabular}{|
    >{\RaggedRight\arraybackslash}m{1.6cm}|  
    >{\RaggedRight\arraybackslash}m{3cm}|  
    >{\centering\arraybackslash}m{0.8cm}|    
    >{\centering\arraybackslash}m{2cm}|    
    >{\centering\arraybackslash}m{1cm}|    
    >{\centering\arraybackslash}m{2cm}|    
    }
        \hline
        Problem & Method & Batch size & Training time (iter/sec) & \textbf{$N_{test}$} & Inference time (sec) \\
        \hline
        \multirow{6}{*}{\shortstack[l]{Burgers' \\ (1D)}} 
        & TI(L)-DON (Ours)   & \multirow{6}{*}{256} & 163.62 & \multirow{4}{*}{2500} & 5.48 \\
        & TI-DON AB (Ours)     &  & 195.43 & & 5.86 \\
        & DON Full rollout     & & 283.94 & & 0.38 \\
        & DON Autoregressive   & & 296.09 & & 6.61 \\
        & FNO Full rollout & & 24.11 & & 3.36 \\
        & FNO Autoregressive & & 285.32 & & 2.12 \\
        \hline
        \multirow{6}{*}{\shortstack[l]{KdV \\ (1D)}} 
        & TI(L)-DON (Ours)   & \multirow{6}{*}{256} & 179.82 & \multirow{6}{*}{1000} & 4.61 \\
        & TI-DON AB (Ours)     & & 208.59 & & 4.47 \\
        & DON Full rollout     & & 234.98 & & 1.22 \\
        & DON Autoregressive   & & 298.44 & & 5.98 \\
        & FNO Full rollout & & 10.96 & & 6.31 \\
        & FNO Autoregressive & & 306.53 & & 1.49 \\
        \hline
        \multirow{6}{*}{\shortstack[l]{KS \\ (1D)}} 
        & TI(L)-DON (Ours)   & \multirow{6}{*}{128} & 117.55 & \multirow{6}{*}{3000} & 7.67 \\
        & TI-DON AB (Ours)     & & 122.11 & & 8.12 \\
        & DON Full rollout     & & 22.39 & & 5.26 \\
        & DON Autoregressive   & & 232.62 & & 7.15 \\
        & FNO Full rollout & & 4.52 & & 7.16 \\
        & FNO Autoregressive & & 341.34 & & 5.13 \\
        \hline

        \multirow{6}{*}{\shortstack[l]{\rev{Burgers'} \\ \rev{(2D)}}} 
        & TI(L)-DON (Ours)   & \multirow{6}{*}{32} & 8.91 & \multirow{6}{*}{1000}  & 8.57 \\
        & TI-DON AB (Ours)     &  & 9.85 & & 4.56 \\
        & DON Full rollout     & & 64.23 & & 2.29 \\
        & DON Autoregressive   & & 37.71 & & 8.26 \\
        & FNO Full rollout &  & 2.67 & & 19.85 \\
        & FNO Autoregressive &  & 97.96 & & 13.66 \\
        \hline

        \multirow{6}{*}{\shortstack[l]{\rev{Rotation} \\ \rev{Advection} \\ \rev{Diffusion} \\ \rev{(2D)}}} 
        & \rev{TI(L)-DON (Ours)}   & \multirow{6}{*}{\rev{32}} & \rev{6.27} & \multirow{6}{*}{\rev{800}}  & \rev{7.54} \\
        & \rev{TI-DON AB (Ours)}     &  & \rev{6.89} & & \rev{4.89} \\
        & \rev{DON Full rollout}     & & \rev{55.43} & & \rev{2.48} \\
        & \rev{DON Autoregressive}   & & \rev{32.55} & & \rev{5.37} \\
        & \rev{FNO Full rollout} &  & \rev{1.67} & & \rev{21.76} \\
        & \rev{FNO Autoregressive} &  & \rev{65.19} & & \rev{15.84} \\
        \hline

        \multirow{4}{*}{\shortstack[l]{\rev{Heat} \\ \rev{Conduction} \\ \rev{(3D)}}} 
        & \rev{TI(L)-DON (Ours)}   & \multirow{4}{*}{\rev{64}} & \rev{10.64} & \multirow{4}{*}{\rev{1000}}  & \rev{14.39} \\
        & \rev{TI-DON AB (Ours)}     &  & \rev{12.83} & & \rev{11.35} \\
        & \rev{DON Full rollout}     & & \rev{2.99} & & \rev{2.62} \\
        & \rev{DON Autoregressive}   & & \rev{55.44} & & \rev{11.47} \\
        \hline
        

    \end{tabular}
    \label{tab:training-inference-times}
\end{table}

\paragraph{\rev{Training cost analysis:}}
\rev{
For the sake of comparison of the theoretical computational complexity for the TI-based variants, since the backbone NO is DeepONet, we present the theoretical comparison with respect to DON FR and DON AR. We note that a similar analysis can be performed agnostic of the specific NO architecture in a more general setting. During training, the computational complexity varies across methods. The branch network requires $\mathcal{O}(bs)$ forward passes per iteration, where $bs$ represents the batch size. The trunk network evaluation complexity depends on the strategy employed:}
\begin{itemize}[leftmargin=*]
\item \rev{Full rollout: $\mathcal{O}(n_s \times n_t)$ forward passes, where $n_s$ is the number of spatial points and $n_t$ is the number of temporal locations.} \vspace{-4pt}
\item \rev{Autoregressive methods (autoregressive DeepONet, TI-DeepONet, and TI(L)-DeepONet): $\mathcal{O}(n_s)$ forward passes due to the absence of the temporal dimension.}
\end{itemize}
\rev{
For TI(L)-DeepONet, an additional $\mathcal{O}(bs)$ forward passes are required per iteration to evaluate the auxiliary deep neural network that learns the intermediate RK4 weighting coefficients. As a result, TI(L)-DeepONet incurs the highest per-iteration cost, leading to the lowest iteration speed observed in Table~\ref{tab:training-inference-times}. Although TI-DeepONet and autoregressive DeepONet evaluate a similar number of forward passes, TI-DeepONet additionally invokes the time integrator, introducing extra forward computations. Consequently, it is more expensive to train than autoregressive DeepONet but still cheaper than TI(L)-DeepONet. The cost of the full rollout method scales directly with the spatiotemporal resolution. While it avoids recursion, the simultaneous prediction over all space-time points makes it more expensive than autoregressive methods, which predict for only spatial points at each step. The trends observed in Table~\ref{tab:training-inference-times} align with this analysis. Note that in the 2D Burgers' and 2D Rotation Advection-Diffusion cases, the relatively higher training cost incurred by the autoregressive DeepONet is due to the use of additional hidden layers in the trunk network.
}
\rev{
\paragraph{Inference cost analysis:}During inference, full rollout is the fastest method, requiring $\mathcal{O}(N_{\text{test}}) + \mathcal{O}(n_s \times n_t)$ forward passes for total evaluation, where $N_{\text{test}}$ is the number of test samples. In contrast, the autoregressive, TI-DeepONet, and TI(L)-DeepONet methods perform recursive time predictions, resulting in longer inference times. For evaluating $N_{\text{test}}$ samples, the autoregressive networks simulate $n_t \times \mathcal{O}(N_{\text{test}}) + n_t \times \mathcal{O}(n_s)$ forward passes.
During inference, the additional network in the TI(L)-DeepONet is not evaluated as the optimized parameters are already obtained, resulting in no additional cost.
}

\section{Choice of Inference Timestep}
\label{sec:temporal_step_variation}
In our primary evaluations, TI-DeepONet was trained and tested using the same temporal resolution, i.e., the training and inference timesteps ($\Delta t$) were matched. However, the architecture's compatibility with higher-order numerical integrators suggests that, in principle, inference may be conducted using timesteps different from those used during training. As long as the chosen $\Delta t$ satisfies the Courant-Friedrichs-Lewy (CFL) stability condition, the resulting time integration should remain stable. This flexibility stems from the fact that TI-DeepONet mirrors traditional numerical solvers: both employ explicit time-stepping schemes, with TI-DeepONet approximating the right-hand side of the PDE via a neural operator rather than finite-difference or finite-element discretizations of spatial derivatives.

To investigate this capability, we performed preliminary experiments on the 1D Burgers' equation, evaluating TI-DeepONet at three inference timesteps, $\Delta t \in \lbrace 0.1, 0.01, 0.001 \rbrace$, while training was performed at $\Delta t = 0.01$. Notably, TI-DeepONet preserved comparable $L_2$ error even when evaluated with a significantly coarser timestep ($\Delta t = 0.1$) than that used during training. \rev{This observation is further corroborated by the 2D Burgers' case, as shown in Figure~\ref{fig:timestep_refinement}, where both TI-DON and TI(L)-DON maintain consistent performance across varying inference timesteps.} This finding suggests the potential for developing more efficient surrogates that accelerate inference without additional training costs.

\section{Conclusions}
\label{sec:conclusion}
Accurate long-term extrapolation of dynamical systems remains a fundamental challenge for neural operators, particularly due to error accumulation in autoregressive predictions. To address this, we introduce TI-DeepONet and its adaptive variant TI(L)-DeepONet, which integrates classical numerical time integration with an operator learning framework (DeepONet). By reformulating the learning objective to approximate instantaneous time derivatives—subsequently integrated via high-order numerical schemes, our frameworks preserve the causal structure of dynamical systems while enhancing stability and accuracy in long-term forecasting. Furthermore, the operator learning paradigm enables continuous spatiotemporal solutions for parametrized initial conditions, thereby enabling the construction of an efficient and reliable emulator for complex systems. \rev{Extensive evaluations across six canonical PDEs (1D/2D Burgers', 1D KdV, 1D KS, 2D Rotation Advection-Diffusion, and 3D Heat Conduction equations) demonstrate the effectiveness of our approach.} \revv{For the majority of benchmark problems, particularly diffusion-dominated and chaotic systems, the TI-based variants demonstrate superior performance over baseline methods.} TI-DeepONet reduces relative $L_2$ extrapolation errors by \rev{96.3\%} compared to autoregressive DeepONet and \rev{83.6\%} compared to full rollout methods, while maintaining stable predictions for temporal domains extending up to twice the training interval. The learnable coefficients in TI(L)-DeepONet further refine accuracy by adaptively weighting intermediate integration slopes, enabling solution-specific adjustments that outperform static numerical schemes in \revv{most of} our examples. \revv{However, for the rotation advection-diffusion problem involving rigid body rotational dynamics, FNO AR outperforms the TI-based variants, suggesting that the choice of neural operator architecture may be as important as the time integration strategy for certain dynamical systems. This motivates a broader perspective: while we proposed TI-DeepONet with DeepONet as the neural operator in this work, the framework can be viewed more generally as TI-NO (Time-Integrator embedded Neural Operator), where both the time integration scheme and the neural operator architecture can be independently selected based on the problem characteristics.} By embedding temporal causality into the learning process and decoupling training-time integrators (e.g., Runge-Kutta) from inference-time integrators (e.g., Runge-Kutta or Adams-Bashforth), our methodology bridges neural operator flexibility with the robustness of numerical analysis. Among the proposed variants, TI(L)-DeepONet achieves the highest accuracy, albeit with a higher per-iteration computational cost due to its learnable integration coefficients (see SI Section~\ref{sec:compute_cost}). However, this increase is offset by its significantly faster convergence to lower training loss, resulting in no net increase in total compute time. This hybrid paradigm offers a reliable pathway for modeling high-dimensional, time-dependent systems, where stability and generalizability are critical\revv{, though problem characteristics should guide the selection of the most appropriate method and neural operator architecture}.

\section{Limitations and Future Work}
\label{sec:future_work}
While our proposed frameworks demonstrate superior extrapolation accuracy compared to existing neural operator training strategies, they incur higher computational costs during both training and inference, as detailed in Section~\ref{sec:compute_cost}. Although these offline training costs can be justified by the consistent reliability of predictions across a wide range of parametric initial conditions, improving computational efficiency remains a key priority. To this end, we plan to leverage multi-GPU training and investigate methods for more efficient forward and backward propagation through the ODE solver. Additionally, we aim to extend the framework's scalability to \rev{more complex} 3D and multiphysics systems and to incorporate adaptive time-stepping strategies for handling stiff dynamical systems. \revv{Furthermore, as evidenced by the rotation advection-diffusion example, the TI-based variants exhibit limitations in capturing rigid body rotational dynamics; improving performance for such problems through alternative neural operator architectures (e.g., TI-FNO) or finer temporal resolution strategies remains an important direction for future investigation.}

\section*{Acknowledgments}
The authors' research efforts were partly supported by the National Science Foundation (NSF) under Grant No. 2438193 and 2436738. The authors would like to acknowledge computing support provided by the Advanced Research Computing at Hopkins (ARCH) core facility at Johns Hopkins University and the Rockfish cluster. ARCH core facility (\url{rockfish.jhu.edu}) is supported by the NSF grant number OAC1920103. The authors would also like to acknowledge Rajyasri Roy, who investigated the flexibility of the TI-DeepONet architecture and analyzed the inference accuracy when using coarser timesteps at inference compared to the finer timesteps used during training for the 1D Burgers equation. Any opinions, findings, conclusions, or recommendations expressed in this material are those of the author(s) and do not necessarily reflect the views of the funding organizations.

\bibliographystyle{unsrt}
\bibliography{references}

\newpage
\renewcommand{\thetable}{A\arabic{table}}  
\renewcommand{\thefigure}{A\arabic{figure}} 
\makeatother
\setcounter{figure}{0}
\setcounter{table}{0}
\setcounter{section}{1}
\setcounter{page}{1}
\appendix

\section{Statistical Variability}
\label{sec:statistical_variability}
\begin{table}[htb!]
\caption{Relative $L_2$ errors reported based on five independent training trials for all frameworks in the extrapolation regime.}
    \begin{center}
\renewcommand{\arraystretch}{1.3}
\begin{tabular}{|
    >{\RaggedRight\arraybackslash}m{1.5cm}|  
    >{\RaggedRight\arraybackslash}m{1.8cm}|  
    >{\centering\arraybackslash}m{2cm}|     
    >{\centering\arraybackslash}m{2cm}|     
    >{\centering\arraybackslash}m{2cm}|     
    >{\centering\arraybackslash}m{2cm}|     
}
    \hline
    \multirow{2}{*}{Problem} & \multirow{2}{*}{Method} & \multicolumn{4}{c|}{Relative $L_2$ error} \\
    \cline{3-6}
    & & $t$+10$\Delta t_e$ & $t$+20$\Delta t_e$ & $t$+40$\Delta t_e$ & $T^{\ast}$ \\
    \hline

    \multirow{6}{*}{\shortstack[l]{Burgers'\\(1D)}} 
    & TI(L)-DON        & \textbf{0.019$\pm$0.003} & \textbf{0.023$\pm$0.003} & \textbf{0.036$\pm$0.004} & \textbf{0.044$\pm$0.005} \\
    & TI-DON AB        & \underline{0.031$\pm$0.004} & \underline{0.037$\pm$0.005} & \underline{0.057$\pm$0.008} & \underline{0.070$\pm$0.011} \\
    & DON FR                 & 0.043$\pm$0.002 & 0.095$\pm$0.004 & 0.247$\pm$0.028 & 0.336$\pm$0.053 \\
    & DON AR               & 0.710$\pm$0.089 & 1.004$\pm$0.144 & 1.556$\pm$0.206 & 1.768$\pm$0.227 \\
    & FNO FR         & 0.219$\pm$0.026 & 0.243$\pm$0.025 & 0.291$\pm$0.019 & 0.456$\pm$0.147 \\
    & FNO AR         & 0.202$\pm$0.041 & 0.212$\pm$0.048 & 0.215$\pm$0.059 & 0.216$\pm$0.062 \\
    \hline

    \multirow{6}{*}{\shortstack[l]{KdV\\(1D)}}
    & TI(L)-DON        & \textbf{0.054$\pm$0.019} & \textbf{0.065$\pm$0.027} & \textbf{0.075$\pm$0.031} & \textbf{0.111$\pm$0.051} \\
    & TI-DON AB        & \underline{0.086$\pm$0.026} & \underline{0.108$\pm$0.034} & \underline{0.129$\pm$0.043} & \underline{0.183$\pm$0.063} \\
    & DON FR                 & 0.776$\pm$0.0004 & 0.716$\pm$0.0005 & 0.719$\pm$0.0005 & 0.795$\pm$0.0007 \\
    & DON AR               & 0.823$\pm$0.073 & 0.886$\pm$0.064 & 0.922$\pm$0.069 & 0.968$\pm$0.083 \\
    & FNO FR         & 0.900$\pm$0.038 & 0.828$\pm$0.027 & 0.756$\pm$0.025 & 0.681$\pm$0.033 \\
    & FNO AR         & 0.343$\pm$0.086 & 0.443$\pm$0.096 & 0.496$\pm$0.091 & 0.543$\pm$0.095 \\
    \hline

    \multirow{6}{*}{\shortstack[l]{KS\\(1D)}}
    & TI(L)-DON        & \textbf{0.048$\pm$0.006} & \textbf{0.088$\pm$0.009} & \textbf{0.230$\pm$0.016} & \textbf{0.331$\pm$0.018} \\
    & TI-DON AB        & \underline{0.058$\pm$0.010} & \underline{0.103$\pm$0.016} & \underline{0.248$\pm$0.018} & \underline{0.349$\pm$0.017} \\
    & DON FR                 & 0.598$\pm$0.130 & 0.924$\pm$0.019 & 0.844$\pm$0.003 & 0.885$\pm$0.028 \\
    & DON AR               & 1.199$\pm$0.056 & 1.218$\pm$0.048 & 1.264$\pm$0.046 & 1.294$\pm$0.045 \\
    & FNO FR         & 0.921$\pm$0.002 & 0.946$\pm$0.004 & 0.889$\pm$0.004 & 0.922$\pm$0.008 \\
    & FNO AR         & 1.596$\pm$0.260 & 1.668$\pm$0.271 & 1.744$\pm$0.278 & 1.765$\pm$0.303 \\
    \hline

    \multirow{6}{*}{\shortstack[l]{\rev{Burgers'}\\\rev{(2D)}}}
    & TI(L)-DON        & \underline{0.186$\pm$0.003} & \underline{0.223$\pm$0.004} & \textbf{0.292$\pm$0.005} & \textbf{0.324$\pm$0.005} \\
    & TI-DON AB        & 0.189$\pm$0.002 & 0.226$\pm$0.003 & \underline{0.297$\pm$0.003} & \underline{0.330$\pm$0.003} \\
    & DON FR                 & 0.373$\pm$0.042 & 0.412$\pm$0.032 & 0.494$\pm$0.028 & 0.534$\pm$0.039 \\
    & DON AR               & 0.883$\pm$0.275 & 1.089$\pm$0.364 & 1.532$\pm$0.553 & 1.765$\pm$0.661 \\
    & FNO FR         & 0.315$\pm$0.023 & 0.345$\pm$0.027 & 0.742$\pm$0.597 & 1.689$\pm$0.764 \\
    & FNO AR         & \bf{0.096$\pm$0.013} & \bf{0.208$\pm$0.051} & 0.767$\pm$0.432 & 1.37$\pm$1.004 \\
    \hline

    \multirow{6}{*}{\shortstack[l]{\rev{Rotation} \\ \rev{Advection}\\ \rev{Diffusion} \\ \rev{(2D)}}}
    & \rev{TI(L)-DON}        & \rev{\underline{0.108$\pm$0.0005}} & \rev{\underline{0.133$\pm$0.0005}} & \rev{0.198$\pm$0.0007} & \rev{0.241$\pm$0.003} \\
    & \rev{TI-DON AB}        & \rev{0.109$\pm$0.0004} & \rev{0.133$\pm$0.0006} & \rev{\underline{0.197$\pm$0.001}} & \rev{\underline{0.238$\pm$0.003}} \\
    & \rev{DON FR}           & \rev{0.136$\pm$0.039} & \rev{0.428$\pm$0.045} & \rev{1.048$\pm$0.036} & \rev{1.121$\pm$0.036} \\
    & \rev{DON AR}           & \rev{0.756$\pm$0.212} & \rev{0.836$\pm$0.202} & \rev{1.307$\pm$0.618} & \rev{1.424$\pm$0.647} \\
    & \rev{FNO FR}           & \rev{0.498$\pm$0.011} & \rev{0.543$\pm$0.016} & \rev{0.656$\pm$0.022} & \rev{0.819$\pm$0.082} \\
    & \rev{FNO AR}           & \rev{\bf{0.012$\pm$0.0007}} & \rev{\bf{0.016$\pm$0.001}} & \rev{\bf{0.027$\pm$0.003}} & \rev{\bf{0.033$\pm$0.004}} \\
    \hline

    \multirow{4}{*}{\shortstack[l]{\rev{Heat}\\ \rev{Conduction} \\ \rev{(3D)}}}
    & \rev{TI(L)-DON}        & \rev{\bf{0.014$\pm$0.0001}} & \rev{\bf{0.019$\pm$0.0002}} & \rev{\bf{0.032$\pm$0.0006}} & \rev{\bf{0.036$\pm$0.0007}} \\
    & \rev{TI-DON AB}        & \rev{\underline{0.014$\pm$0.0002}} & \rev{\underline{0.019$\pm$0.0002}} & \rev{\underline{0.033$\pm$0.0006}} & \rev{\underline{0.037$\pm$0.0007}} \\
    & \rev{DON FR}           & \rev{0.262$\pm$0.068} & \rev{0.267$\pm$0.063} & \rev{0.291$\pm$0.049} & \rev{0.300$\pm$0.048} \\
    & \rev{DON AR}           & \rev{0.691$\pm$0.163} & \rev{0.827$\pm$0.250} & \rev{1.049$\pm$0.416} & \rev{1.093$\pm$0.445} \\
    \hline


\end{tabular}
\end{center}
    \label{tab:problem_summary_independent_runs}
\end{table}

To evaluate the performance of the model, we compute the relative $L_2$ error of the extrapolation predictions, and report the mean and standard deviation of this metric based on five independent training trials in Table~\ref{tab:problem_summary_independent_runs}. Across five independent trial runs, TI(L)-DeepONet consistently achieves the lowest extrapolation error across all benchmark problems. In addition to superior accuracy, it also exhibits reduced variance in error estimates, indicating enhanced robustness and reliability. 
As expected, both the full rollout and autoregressive DeepONet models suffer from substantially higher errors, particularly outside the training temporal domain. Similarly, the FNO variants, both full rollout and autoregressive incur higher errors. FNO FR generally exhibits higher errors throughout the entire temporal domain and yields particularly unstable solutions due to its inability to capture all global frequencies of the system. Notably, this limitation also manifests as higher variance in the predictions. For FNO AR, spectral instability causes error spikes in the extrapolation domain, leading to higher solution variance - consistent with our observations highlighted in Sec.~\ref{sec:results}. Overall, the TI-based frameworks not only enhance extrapolation accuracy but also offer greater robustness across random initializations, underscoring their effectiveness for modeling time-dependent PDEs.

\section{Additional details}
\label{sec:add_details}

\subsection{One-dimensional Burgers’ Equation}
\label{subsec:example1_data}

\begin{table}[htb!]
    \renewcommand{\arraystretch}{1.25}
    \centering
    \caption{1D Burgers' Equation: Neural network architecture and training settings across different DeepONet-based methods. \rev{Note that each epoch corresponds to a single gradient descent update over a randomly sampled batch, equivalent to one stochastic gradient descent iteration.}}
    \begin{tabular}{|
        >{\RaggedRight\arraybackslash}m{2.25cm}|  
        >{\RaggedRight\arraybackslash}m{3cm}|  
        >{\RaggedRight\arraybackslash}m{2.25cm}|  
        >{\centering\arraybackslash}m{1.35cm}|    
        >{\centering\arraybackslash}m{1.4cm}|    
        >{\centering\arraybackslash}m{0.9cm}|    
    }
        \hline
        Method & Branch Net & Trunk Net & Activation & Epochs & Batch Size\\
        \hline
        \shortstack[l]{TI(L)-DON} 
        & [101], [100]$\times$6, [60] & [100]$\times$6, [60] & Tanh & $1\times10^5$ & 256\\
        \hline
        \shortstack[l]{TI-DON} 
        & [101], [100]$\times$6, [60] & [100]$\times$6, [60] & Tanh & $1\times10^5$ & 256\\
        \hline
        DON FR 
        & [128]$\times$6, [60]  & [128]$\times$6, [60]  & Tanh & $2\times10^5$ & 256\\
        \hline
        DON AR 
        & [101], [100]$\times$6, [60] & [100]$\times$6, [60] & Tanh & $1\times10^5$ & 256\\
        \hline
    \end{tabular}
    \label{tab:1d_burgers_don-architecture}
\end{table}

\begin{table}[htb!]
    \renewcommand{\arraystretch}{1.25}
    \centering
    \caption{1D Burgers' Equation: Neural network architecture and training settings across different FNO-based methods.}
    \begin{tabular}{|>{\centering\arraybackslash}p{1.35cm}|
    >{\centering\arraybackslash}p{1.3cm}|
    >{\centering\arraybackslash}p{1.1cm}|>{\centering\arraybackslash}p{1.3cm}|>{\centering\arraybackslash}p{1.4cm}|>{\centering\arraybackslash}p{1.2cm}|>{\centering\arraybackslash}p{1.25cm}|}
        \hline
        Method & Fourier Blocks & Modes & Hidden Dim & Activation & Epochs & Batch Size\\
        \hline
        FNO AR & 6 & 24 & 64 & GELU & 30000 & 128 \\
        \cline{1-7}
        FNO FR & 4 & 16 & 32 & GELU & 5000 & 64 \\
        \hline
    \end{tabular}
    \label{tab:1d_burgers_fno-architecture}
\end{table}
\begin{figure}[htb!]
    \centering
    \includegraphics[width=\linewidth]{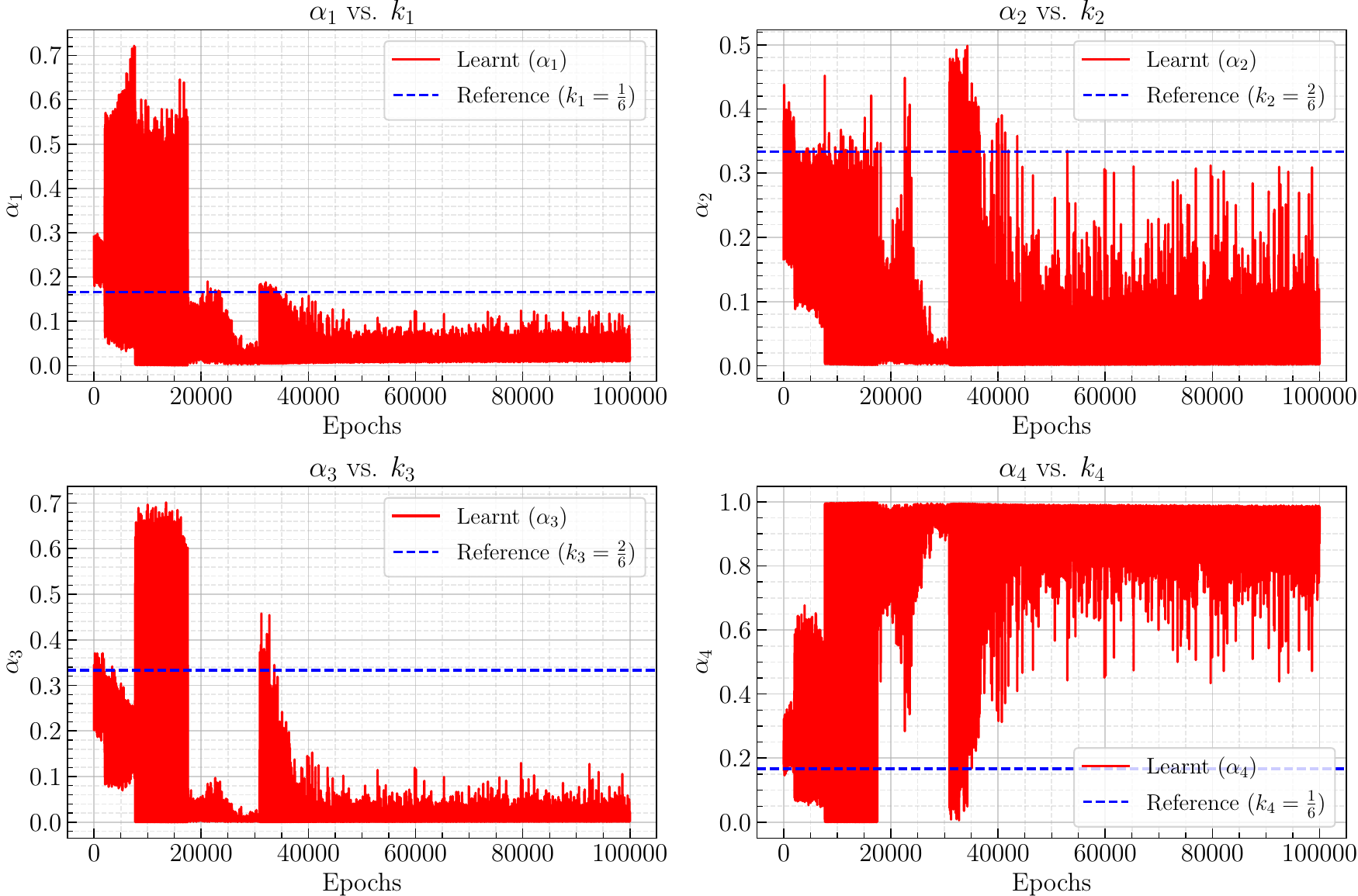}
    \caption{1D Burgers' Equation: Epoch-wise change of the learnable RK4 slope coefficients $\boldsymbol{\alpha} = [\alpha_1, \alpha_2, \alpha_3, \alpha_4]$.}
    \label{fig:learnableweights}
\end{figure}

\paragraph{\textbf{Learnable Weights:}}As previously discussed, \rev{in the TI(L)-DeepONet architecture}, the learnable slope coefficients $\boldsymbol{\alpha} = [\alpha_1, \alpha_2, \alpha_3, \alpha_4]$ are predicted by the additional NN, which takes the current solution state $u^i$ as input, i.e., $\boldsymbol{\alpha} = \text{NN}_{\theta}(u^i)$. Analyzing the evolution of these solution-specific coefficients over training epochs provides insight into how the network adapts the time integrator to local dynamics, in contrast to using a fixed set of RK4 coefficients. This comparison is illustrated in Fig.~\ref{fig:learnableweights}. The dashed blue line in each plot indicates the reference slope coefficients typically used in the RK4 method, derived from a Taylor series expansion. These coefficients result in a fourth-order accurate scheme and are generally expected to provide high accuracy for traditional time integrators. However, in the context of the operator learning framework, these coefficients also adapt to the approximation error of the neural operator. Making these slope coefficients learnable allows the model to better compensate for such errors, thereby improving the overall accuracy of the framework.

\clearpage
\subsection{One-dimensional Korteweg-de Vries (KdV) Equation}
\label{subsec:example2_data}
\begin{table}[htb!]
    \renewcommand{\arraystretch}{1.25}
    \centering
    \caption{1D KdV Equation: Neural network architecture and training settings across different DeepONet methods. \rev{Note that each epoch corresponds to a single gradient descent update over a randomly sampled batch, equivalent to one stochastic gradient descent iteration.}}
    \begin{tabular}{|
        >{\RaggedRight\arraybackslash}m{2.25cm}|  
        >{\RaggedRight\arraybackslash}m{2.65cm}|  
        >{\RaggedRight\arraybackslash}m{2.65cm}|  
        >{\centering\arraybackslash}m{1.35cm}|    
        >{\centering\arraybackslash}m{1.4cm}|    
        >{\centering\arraybackslash}m{1cm}|    
    }
        \hline
        Method & Branch Net & Trunk Net & Activation & Epochs & Batch Size\\
        \hline
        \shortstack[l]{TI(L)-DON} 
        & [128]$\times$6, [80] & [128]$\times$6, [80] & Tanh & $1.5 \times 10^5$ & 256 \\
        \hline
        \shortstack[l]{TI-DON} 
        & [128]$\times$6, [80] & [128]$\times$6, [80] & Tanh & $1\times10^5$ & 256 \\
        \hline
        DON FR 
        & [150, 250, 450, 380, 320, 300, 80] 
        & [200, 220, 240, 250, 260, 280, 300, 80] & Swish & $1\times10^5$ & 256\\
        \hline
        DON AR 
        & [128]$\times$6, [80] & [128]$\times$6, [80] & Tanh & $1\times10^5$ & 256\\
        \hline
    \end{tabular}
    \label{tab:1d_kdv_don-architecture}
\end{table}

\begin{table}[htb!]
    \renewcommand{\arraystretch}{1.25}
    \centering
    \caption{1D KdV Equation: Neural network architecture and training settings across different FNO-based methods.}
    \begin{tabular}{|>{\centering\arraybackslash}p{1.35cm}|
    >{\centering\arraybackslash}p{1.3cm}|
    >{\centering\arraybackslash}p{1.1cm}|>{\centering\arraybackslash}p{1.3cm}|>{\centering\arraybackslash}p{1.4cm}|>{\centering\arraybackslash}p{1.2cm}|>{\centering\arraybackslash}p{1.25cm}|}
        \hline
        Method & Fourier Blocks & Modes & Hidden Dim & Activation & Epochs & Batch Size\\
        \hline
        FNO AR & 6 & 24 & 64 & GELU & 30000 & 128 \\
        \cline{1-7}
        FNO FR & 6 & 32 & 64 & GELU & 10000 & 64 \\
        \hline
    \end{tabular}
    \label{tab:1d_kdv_fno-architecture}
\end{table}

\subsection{One-dimensional Kuramuto-Sivashinsky (KS) Equation}
\label{subsec:example3_data}
\begin{table}[htb!]
    \renewcommand{\arraystretch}{1.25}
    \centering
    \caption{1D KS Equation: Neural network architecture and training settings across different DeepONet methods. \rev{Note that each epoch corresponds to a single gradient descent update over a randomly sampled batch, equivalent to one stochastic gradient descent iteration.}}
    \begin{tabular}{|
        >{\RaggedRight\arraybackslash}m{1.75cm}|  
        >{\RaggedRight\arraybackslash}m{2.25cm}|  
        >{\RaggedRight\arraybackslash}m{2.25cm}|  
        >{\centering\arraybackslash}m{2cm}|    
        >{\centering\arraybackslash}m{1.5cm}|    
        >{\centering\arraybackslash}m{1cm}|    
    }
        \hline
        Method & Branch Net & Trunk Net & Activation & Epochs & Batch Size\\
        \hline
        \shortstack[l]{TI(L)-DON} 
        & [128]$\times$7, [100] & [128]$\times$7, [100] & Branch: Tanh Trunk: Sine & $1.75 \times 10^5$ & 128 \\
        \hline
        \shortstack[l]{TI-DON} 
        & [128]$\times$7, [100] & [128]$\times$7, [100] & Branch: Tanh Trunk: Sine & $1.5 \times 10^5$ & 128 \\
        \hline
        DON FR 
        & [128]$\times$7, [100] & [128]$\times$7, [100] & Branch: SiLU Trunk: Sine & $1.5 \times 10^5$ & 128\\
        \hline
        DON AR 
        & [128]$\times$7, [100] & [128]$\times$7, [100] & Branch: Tanh Trunk: Sine & $1.5 \times 10^5$ & 128\\
        \hline
    \end{tabular}
    \label{tab:1d_ks_don-architecture}
\end{table}

\begin{table}[htb!]
    \renewcommand{\arraystretch}{1.25}
    \centering
    \caption{1D KS Equation: Neural network architecture and training settings across different FNO-based methods.}
    \begin{tabular}{|>{\centering\arraybackslash}p{1.35cm}|
    >{\centering\arraybackslash}p{1.3cm}|
    >{\centering\arraybackslash}p{1.1cm}|>{\centering\arraybackslash}p{1.3cm}|>{\centering\arraybackslash}p{1.4cm}|>{\centering\arraybackslash}p{1.2cm}|>{\centering\arraybackslash}p{1.25cm}|}
        \hline
        Method & Fourier Blocks & Modes & Hidden Dim & Activation & Epochs & Batch Size\\
        \hline
        FNO AR & 6 & 64 & 64 & GELU & 10000 & 64 \\
        \cline{1-7}
        FNO FR & 6 & 64 & 64 & GELU & 7000 & 64 \\
        \hline
    \end{tabular}
    \label{tab:1d_ks_fno-architecture}
\end{table}

\clearpage
\subsection{\rev{Two-dimensional Burgers’ Equation}}
\label{subsec:example4_data}
\begin{longtable}{|
    >{\RaggedRight\arraybackslash}m{1.75cm}|  
    >{\RaggedRight\arraybackslash}m{3.25cm}|  
    >{\RaggedRight\arraybackslash}m{2.15cm}|  
    >{\centering\arraybackslash}m{1.8cm}|    
    >{\centering\arraybackslash}m{1.3cm}|    
    >{\centering\arraybackslash}m{0.75cm}|    
}
    \caption{\rev{2D Burgers' Equation}: Neural network architecture and training settings across different methods. \rev{Note that each epoch corresponds to a single gradient descent update over a randomly sampled batch, equivalent to one stochastic gradient descent iteration.}}
    \label{tab:2d_burgers_don-architecture}\\
    \hline
    Method & Branch Net & Trunk Net & Activation & Epochs & Batch Size \\
    \hline
    \endfirsthead
    
    \multicolumn{6}{c}{\tablename\ \thetable{} -- Continued from previous page} \\
    \hline
    Method & Branch Net & Trunk Net & Activation & Epochs & Batch Size \\
    \hline
    \endhead
    
    \hline
    \multicolumn{6}{r}{Continued on next page} \\
    \endfoot
    
    \hline
    \endlastfoot
    
    \shortstack[l]{TI(L)-DON} 
    & Conv2D(features = 64, kernel size = (3,3), strides = 1), MaxPool(window shape=(2, 2), strides = (2, 2)),
    Conv2D(features = 64, kernel size = (3,3), strides = 1), MaxPool(window shape=(2, 2), strides = (2, 2)),
    Conv2D(features = 64, kernel size = (2,2), strides = 1), AvgPool(window shape = (2,2), strides = (2,2)), MLP([256, 128, 100])   
    & [128]$\times$6, [100] & (1) Branch: Conv2D - ReLU, MLP - GELU, (2) Trunk: WaveletAct & $2.25\times10^5$ & 64 \\
    \hline
    \shortstack[l]{TI-DON} 
    & Conv2D(features = 64, kernel size = (3,3), strides = 1), MaxPool(window shape=(2, 2), strides = (2, 2)),
    Conv2D(features = 64, kernel size = (3,3), strides = 1), MaxPool(window shape=(2, 2), strides = (2, 2)),
    Conv2D(features = 64, kernel size = (2,2), strides = 1), AvgPool(window shape = (2,2), strides = (2,2)), MLP([256, 128, 100])   
    & [128]$\times$6, [100] & (1) Branch: Conv2D - ReLU, MLP - GELU, (2) Trunk: WaveletAct & $2\times10^5$ & 64 \\
    \hline
    DON FR 
    & Conv2D(features = 64, kernel size = (3,3), strides = 1), MaxPool(window shape=(2, 2), strides = (2, 2)),
    Conv2D(features = 64, kernel size = (3,3), strides = 1), MaxPool(window shape=(2, 2), strides = (2, 2)),
    Conv2D(features = 64, kernel size = (2,2), strides = 1), AvgPool(window shape = (2,2), strides = (2,2)), MLP([256, 128, 100])   
    & [128]$\times$5, [100] & (1) Branch: Conv2D - GELU, MLP - GELU, (2) Trunk: WaveletAct & $1\times10^5$ & 128 \\
    \hline
    DON AR 
    & Conv2D(features = 64, kernel size = (3,3), strides = 1), MaxPool(window shape=(2, 2), strides = (2, 2)),
    Conv2D(features = 64, kernel size = (3,3), strides = 1), MaxPool(window shape=(2, 2), strides = (2, 2)),
    Conv2D(features = 64, kernel size = (2,2), strides = 1), AvgPool(window shape = (2,2), strides = (2,2)), MLP([256, 128, 128, 100])   
    & [128]$\times$6, [100] & (1) Branch: Conv2D - ReLU, MLP - Tanh, (2) Trunk: Sine & $2\times10^5$ & 64 \\
\end{longtable}

\begin{table}[htb!]
    \renewcommand{\arraystretch}{1.25}
    \centering
    \caption{2D Burgers' Equation: Neural network architecture and training settings across different FNO-based methods.}
    \begin{tabular}{|>{\centering\arraybackslash}p{1.35cm}|
    >{\centering\arraybackslash}p{1.3cm}|
    >{\centering\arraybackslash}p{1.1cm}|>{\centering\arraybackslash}p{1.3cm}|>{\centering\arraybackslash}p{1.4cm}|>{\centering\arraybackslash}p{1.2cm}|>{\centering\arraybackslash}p{1.25cm}|}
        \hline
        Method & Fourier Blocks & Modes & Hidden Dim & Activation & Epochs & Batch Size\\
        \hline
        FNO AR & 6 & 12 & 64 & GELU & 1000 & 128 \\
        \cline{1-7}
        FNO FR & 6 & 12 & 64 & GELU & 5000 & 32 \\
        \hline
    \end{tabular}
    \label{tab:2d_burgers_fno-architecture}
\end{table}

\begin{figure}[htb!]
    \centering
    \includegraphics[width=0.9\linewidth]{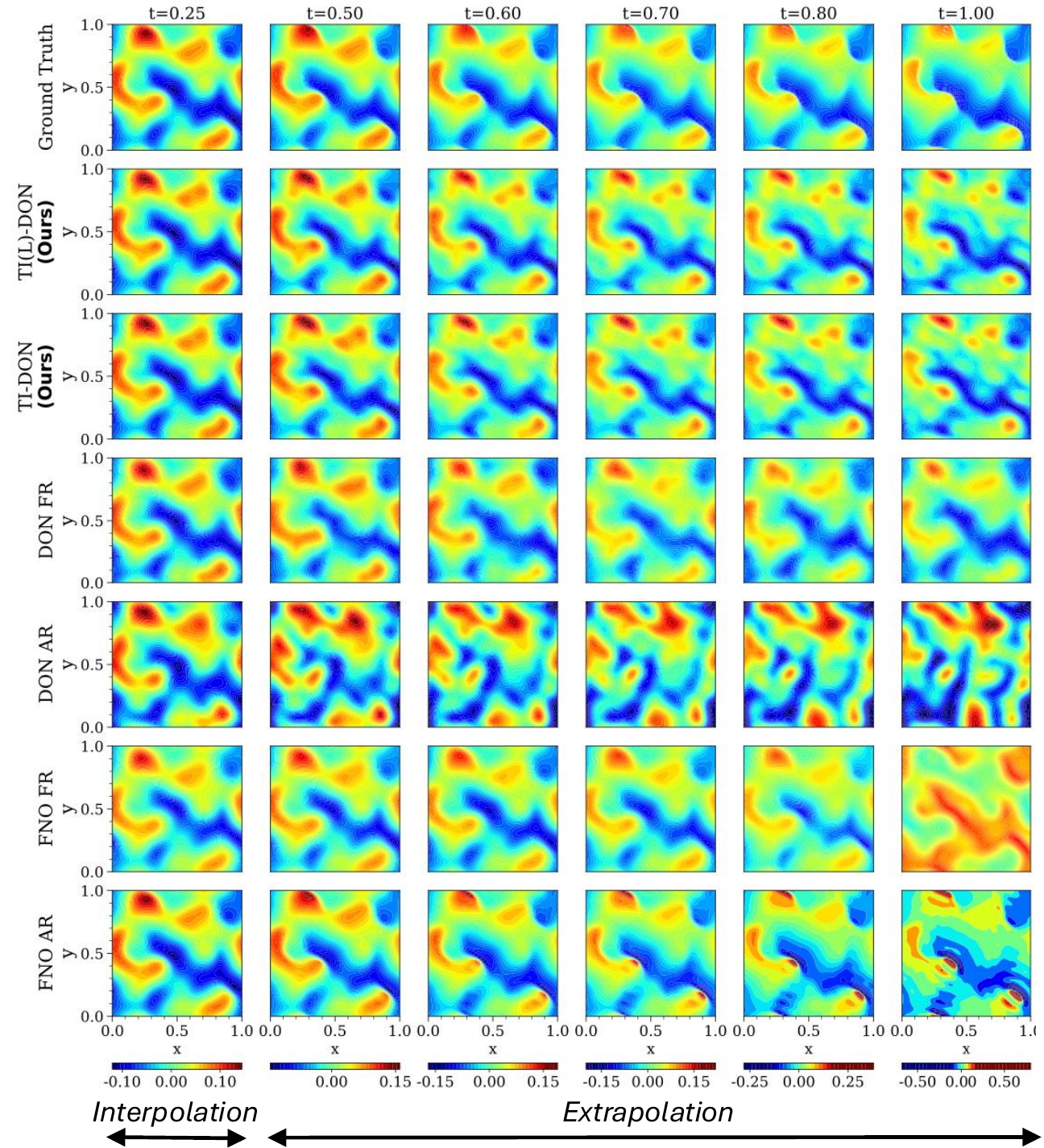}
    \caption{\rev{2D Burgers' Equation}: Solution field for a representative sample, shown across both the training regime ($t \in [0, 0.33]$) and the extrapolation regime ($t \in [0.33, 1]$), for all frameworks. Corresponding error plots are presented in Fig.~\ref{fig:sample_contourplots_2d_burgers_error}.}
    \label{fig:sample_contourplots_2d_burgers}
\end{figure}

\clearpage
\subsection{\rev{Two-dimensional Scalar Rotation Advection-Diffusion}}
\label{subsec:example5_data}
\begin{longtable}{|
    >{\RaggedRight\arraybackslash}m{1.75cm}|  
    >{\RaggedRight\arraybackslash}m{3.25cm}|  
    >{\RaggedRight\arraybackslash}m{2.15cm}|  
    >{\centering\arraybackslash}m{1.8cm}|    
    >{\centering\arraybackslash}m{1.3cm}|    
    >{\centering\arraybackslash}m{0.75cm}|    
}
    \caption{\rev{2D Scalar Rotation Advection-Diffusion}: \rev{Neural network architecture and training settings across different methods.} \rev{Note that each epoch corresponds to a single gradient descent update over a randomly sampled batch, equivalent to one stochastic gradient descent iteration.}}
    \label{tab:2d_rotating_advection_diffusion_don-architecture}\\
    \hline
    Method & Branch Net & Trunk Net & Activation & Epochs & Batch Size \\
    \hline
    \endfirsthead
    
    \multicolumn{6}{c}{\tablename\ \thetable{} -- Continued from previous page} \\
    \hline
    Method & Branch Net & Trunk Net & Activation & Epochs & Batch Size \\
    \hline
    \endhead
    
    \hline
    \multicolumn{6}{r}{Continued on next page} \\
    \endfoot
    
    \hline
    \endlastfoot
    
    \shortstack[l]{TI(L)-DON} 
    & Conv2D(features = 64, kernel size = (3,3), strides = 1), MaxPool(window shape=(2, 2), strides = (2, 2)),
    Conv2D(features = 64, kernel size = (3,3), strides = 1), MaxPool(window shape=(2, 2), strides = (2, 2)),
    Conv2D(features = 64, kernel size = (2,2), strides = 1), AvgPool(window shape = (2,2), strides = (2,2)), MLP([256, 128, 100])   
    & [128]$\times$6, [100] & (1) Branch: Conv2D - GELU, MLP - GELU, (2) Trunk: WaveletAct & $2\times10^5$ & 64 \\
    \hline
    \shortstack[l]{TI-DON} 
    & Conv2D(features = 64, kernel size = (3,3), strides = 1), MaxPool(window shape=(2, 2), strides = (2, 2)),
    Conv2D(features = 64, kernel size = (3,3), strides = 1), MaxPool(window shape=(2, 2), strides = (2, 2)),
    Conv2D(features = 64, kernel size = (2,2), strides = 1), AvgPool(window shape = (2,2), strides = (2,2)), MLP([256, 128, 100])   
    & [128]$\times$6, [100] & (1) Branch: Conv2D - GELU, MLP - GELU, (2) Trunk: WaveletAct & $1.75\times10^5$ & 64 \\
    \hline
    DON FR 
    & Conv2D(features = 64, kernel size = (3,3), strides = 1), MaxPool(window shape=(2, 2), strides = (2, 2)),
    Conv2D(features = 64, kernel size = (3,3), strides = 1), MaxPool(window shape=(2, 2), strides = (2, 2)),
    Conv2D(features = 64, kernel size = (2,2), strides = 1), AvgPool(window shape = (2,2), strides = (2,2)), MLP([256, 128, 100])   
    & [128]$\times$5, [100] & (1) Branch: Conv2D - GELU, MLP - GELU, (2) Trunk: WaveletAct & $1.5\times10^5$ & 64 \\
    \hline
    DON AR 
    & Conv2D(features = 64, kernel size = (3,3), strides = 1), MaxPool(window shape=(2, 2), strides = (2, 2)),
    Conv2D(features = 64, kernel size = (3,3), strides = 1), MaxPool(window shape=(2, 2), strides = (2, 2)),
    Conv2D(features = 64, kernel size = (2,2), strides = 1), AvgPool(window shape = (2,2), strides = (2,2)), MLP([256, 128, 128, 100])   
    & [128]$\times$6, [100] & (1) Branch: Conv2D - ReLU, MLP - Tanh, (2) Trunk: Tanh & $1.5\times10^5$ & 128 \\
\end{longtable}

\begin{table}[htb!]
    \renewcommand{\arraystretch}{1.25}
    \centering
    \caption{\rev{2D Scalar Rotation Advection-Diffusion: Neural network architecture and training settings across different FNO-based methods.}}
    \begin{tabular}{|>{\centering\arraybackslash}p{1.35cm}|
    >{\centering\arraybackslash}p{1.3cm}|
    >{\centering\arraybackslash}p{1.1cm}|>{\centering\arraybackslash}p{1.3cm}|>{\centering\arraybackslash}p{1.4cm}|>{\centering\arraybackslash}p{1.2cm}|>{\centering\arraybackslash}p{1.25cm}|}
        \hline
        Method & Fourier Blocks & Modes & Hidden Dim & Activation & Epochs & Batch Size\\
        \hline
        FNO AR & 6 & 12 & 64 & GELU & 1000 & 64 \\
        \cline{1-7}
        FNO FR & 6 & 12 & 64 & GELU & 5000 & 32 \\
        \hline
    \end{tabular}
    \label{tab:2d_rotating_advection_diffusion_fno-architecture}
\end{table}

\begin{figure}[htb!]
    \centering
    \includegraphics[width=0.9\linewidth]{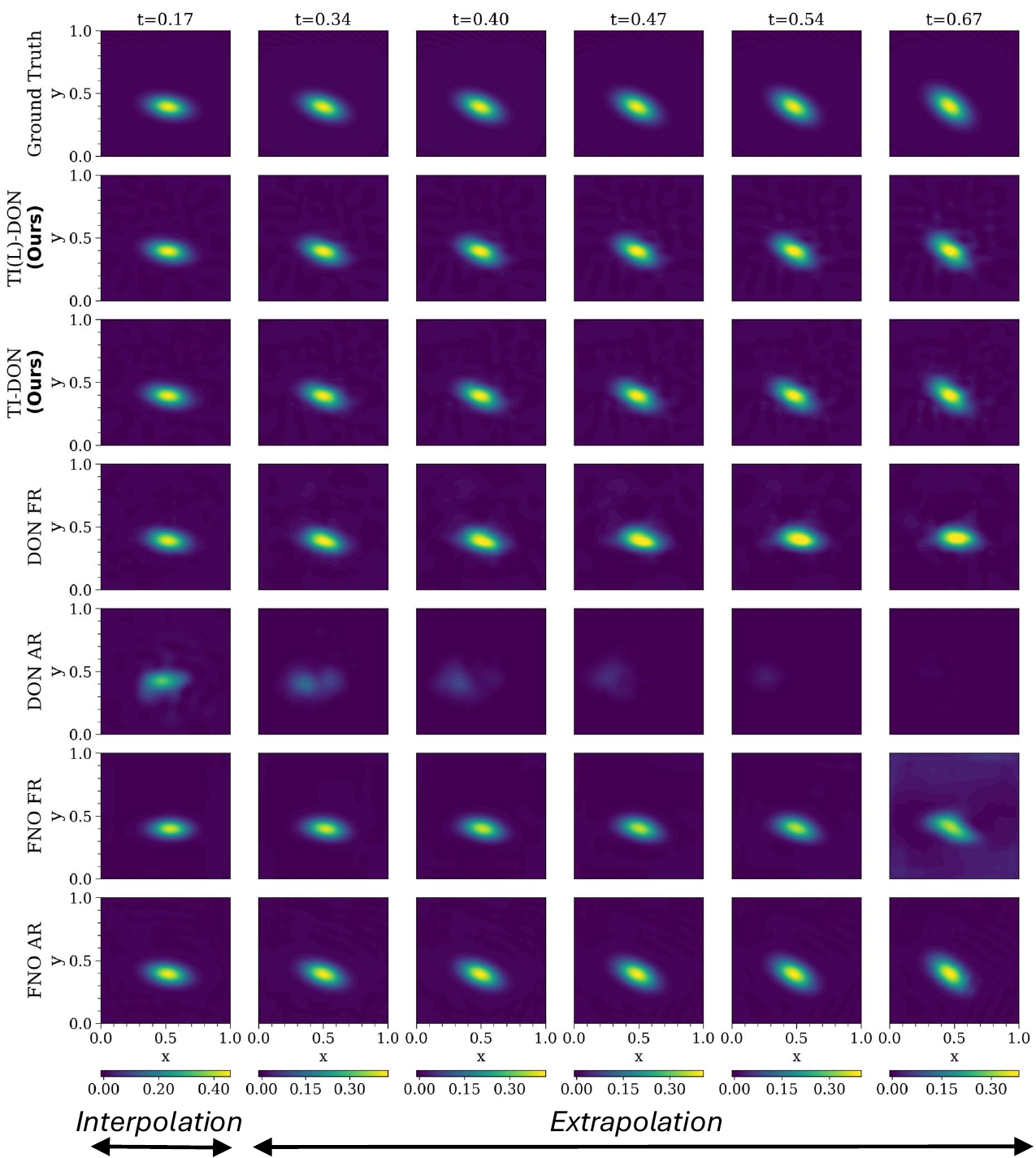}
    \caption{\rev{2D Rotation Advection-Diffusion equation}: \rev{Solution field for a representative sample, shown across both the training regime ($t \in [0, 0.27]$) and the extrapolation regime ($t \in [0.27, 0.67]$), for all frameworks. Corresponding error plots are presented in Fig.~\ref{fig:sample_contourplots_2d_rotating_advection_diffusion_error}.}}
    \label{fig:sample_contourplots_2d_rotating_advection_diffusion}
\end{figure}

\clearpage
\subsection{\rev{Three-dimensional Heat Conduction}}
\label{subsec:example6_data}
\begin{longtable}{|
    >{\RaggedRight\arraybackslash}m{1.75cm}|  
    >{\RaggedRight\arraybackslash}m{3.25cm}|  
    >{\RaggedRight\arraybackslash}m{2.15cm}|  
    >{\centering\arraybackslash}m{1.8cm}|    
    >{\centering\arraybackslash}m{1.3cm}|    
    >{\centering\arraybackslash}m{0.75cm}|    
}
    \caption{\rev{3D Heat Conduction}: \rev{Neural network architecture and training settings across different methods.} \rev{Note that each epoch corresponds to a single gradient descent update over a randomly sampled batch, equivalent to one stochastic gradient descent iteration.}}
    \label{tab:3d_heat_don-architecture}\\
    \hline
    Method & Branch Net & Trunk Net & Activation & Epochs & Batch Size \\
    \hline
    \endfirsthead
    
    \multicolumn{6}{c}{\tablename\ \thetable{} -- Continued from previous page} \\
    \hline
    Method & Branch Net & Trunk Net & Activation & Epochs & Batch Size \\
    \hline
    \endhead
    
    \hline
    \multicolumn{6}{r}{Continued on next page} \\
    \endfoot
    
    \hline
    \endlastfoot
    
    \shortstack[l]{TI(L)-DON} 
    & Conv3D(features = 32, kernel size = (3,3,3), strides = (2,2,1)), MaxPool(window shape=(2,2,2), strides = (2,2,2)),
    Conv3D(features = 32, kernel size = (2,2,2), strides = (2,2,1)), AvgPool(window shape=(2,2,2), strides = (2,2,2)), MLP([256, 128, 100])   
    & [128]$\times$4, [100] & (1) Branch: Conv3D - GELU, MLP - GELU, (2) Trunk: WaveletAct & $2\times10^5$ & 64 \\
    \hline
    \shortstack[l]{TI-DON} 
    & Conv3D(features = 32, kernel size = (3,3,3), strides = (2,2,1)), MaxPool(window shape=(2,2,2), strides = (2,2,2)),
    Conv3D(features = 32, kernel size = (2,2,2), strides = (2,2,1)), AvgPool(window shape=(2,2,2), strides = (2,2,2)), MLP([256, 128, 100])   
    & [128]$\times$4, [100] & (1) Branch: Conv3D - GELU, MLP - GELU, (2) Trunk: WaveletAct & $1.5\times10^5$ & 64 \\
    \hline
    DON FR 
    & Conv3D(features = 32, kernel size = (3,3,3), strides = (2,2,1)), MaxPool(window shape=(2,2,2), strides = (2,2,2)),
    Conv3D(features = 32, kernel size = (2,2,2), strides = (2,2,1)), AvgPool(window shape=(2,2,2), strides = (2,2,2)), MLP([256, 128, 100])   
    & [128]$\times$4, [100] & (1) Branch: Conv3D - GELU, MLP - GELU, (2) Trunk: WaveletAct & $1.5\times10^5$ & 64 \\
    \hline
    DON AR 
    & Conv3D(features = 32, kernel size = (3,3,3), strides = (2,2,1)), MaxPool(window shape=(2,2,2), strides = (2,2,2)),
    Conv3D(features = 32, kernel size = (2,2,2), strides = (2,2,1)), AvgPool(window shape=(2,2,2), strides = (2,2,2)), MLP([256, 128, 100])   
    & [128]$\times$4, [100] & (1) Branch: Conv3D - ReLU, MLP - Tanh, (2) Trunk: Tanh & $1.25\times10^5$ & 32 \\
\end{longtable}

\begin{figure}[htb!]
    \centering
    \includegraphics[width=\linewidth]{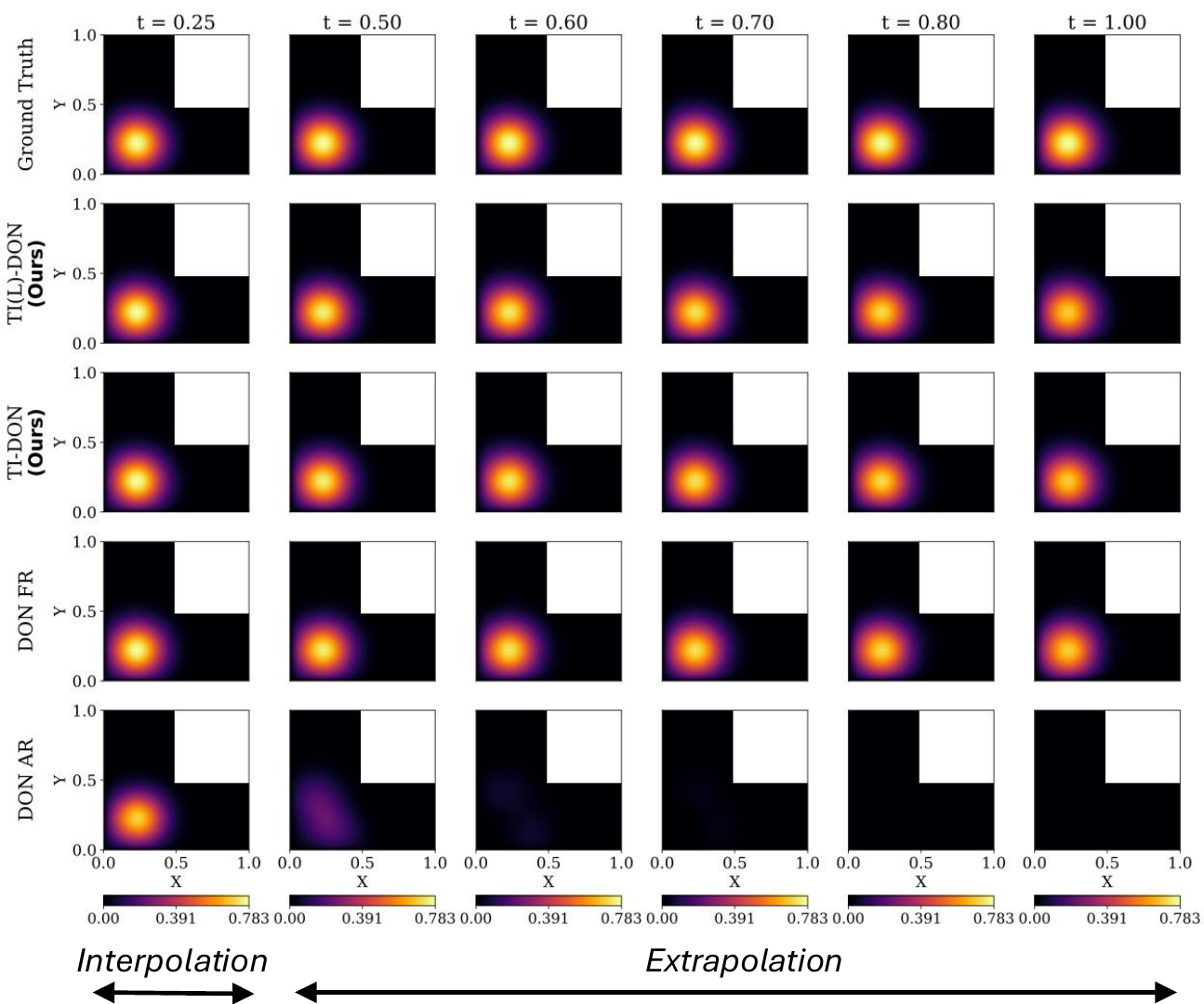}
    \caption{\rev{3D Heat Conduction: Solution field for a representative sample, shown across both the training regime ($t \in [0, 0.33]$) and the extrapolation regime ($t \in [0.33, 1]$), for all frameworks. Corresponding error plots are presented in Fig.~\ref{fig:3d_heat_error_contours_XY}. The contours are presented along the XY slicing plane (see Fig.~\ref{fig:3d_heat_slicing_planes} for details regarding the location of the XY slice).}}
    \label{fig:sample_contourplots_3d_heat_XY}
\end{figure}

\begin{figure}[htb!]
    \centering
    \includegraphics[width=\linewidth]{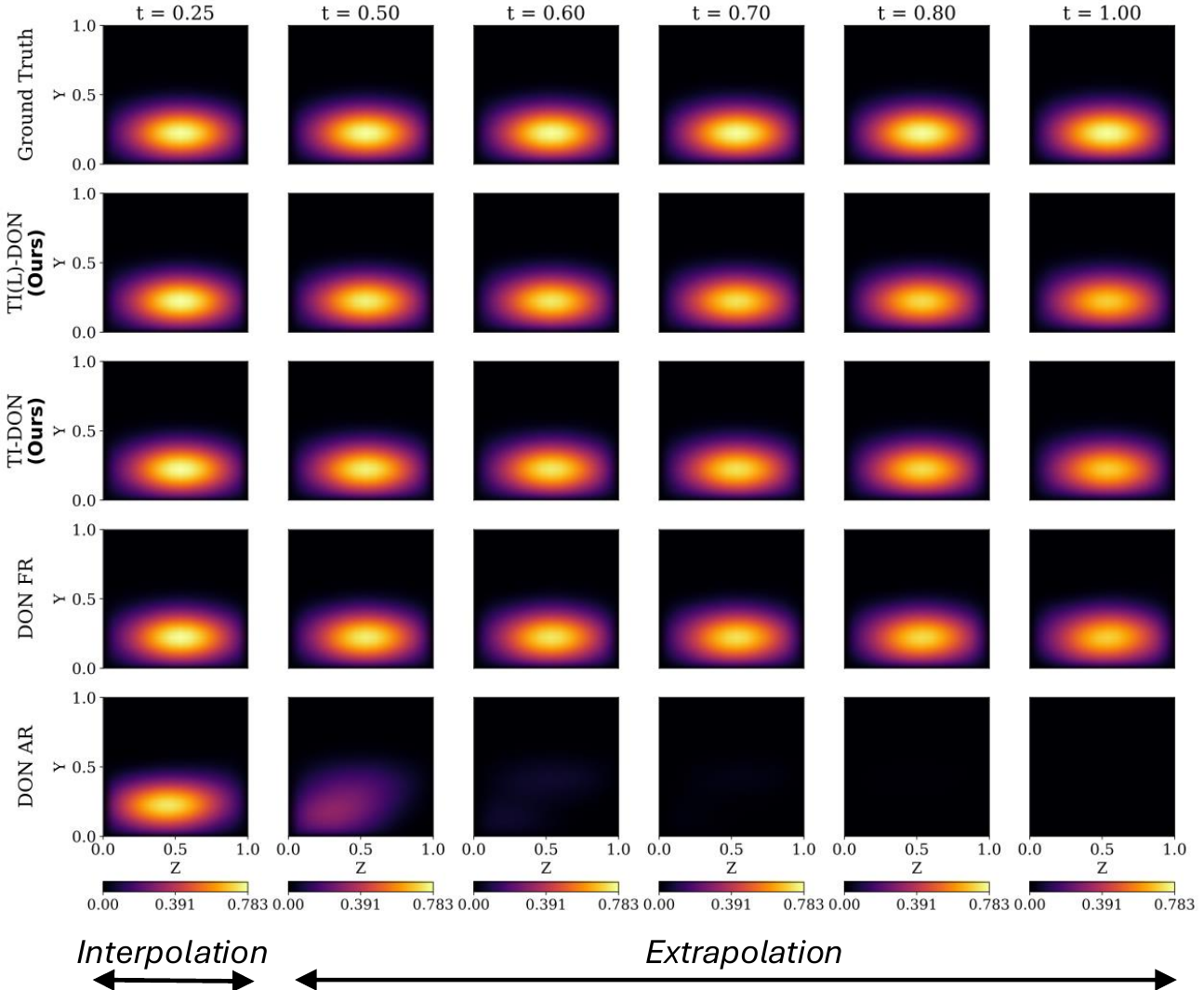}
    \caption{\rev{3D Heat Conduction: Solution field for a representative sample, shown across both the training regime ($t \in [0, 0.33]$) and the extrapolation regime ($t \in [0.33, 1]$), for all frameworks. Corresponding error plots are presented in Fig.~\ref{fig:3d_heat_error_contours_YZ}. The contours are presented along the YZ slicing plane (see Fig.~\ref{fig:3d_heat_slicing_planes} for details regarding the location of the YZ slice).}}
    \label{fig:sample_contourplots_3d_heat_YZ}
\end{figure}

\begin{figure}[htb!]
    \centering
    \includegraphics[width=\linewidth]{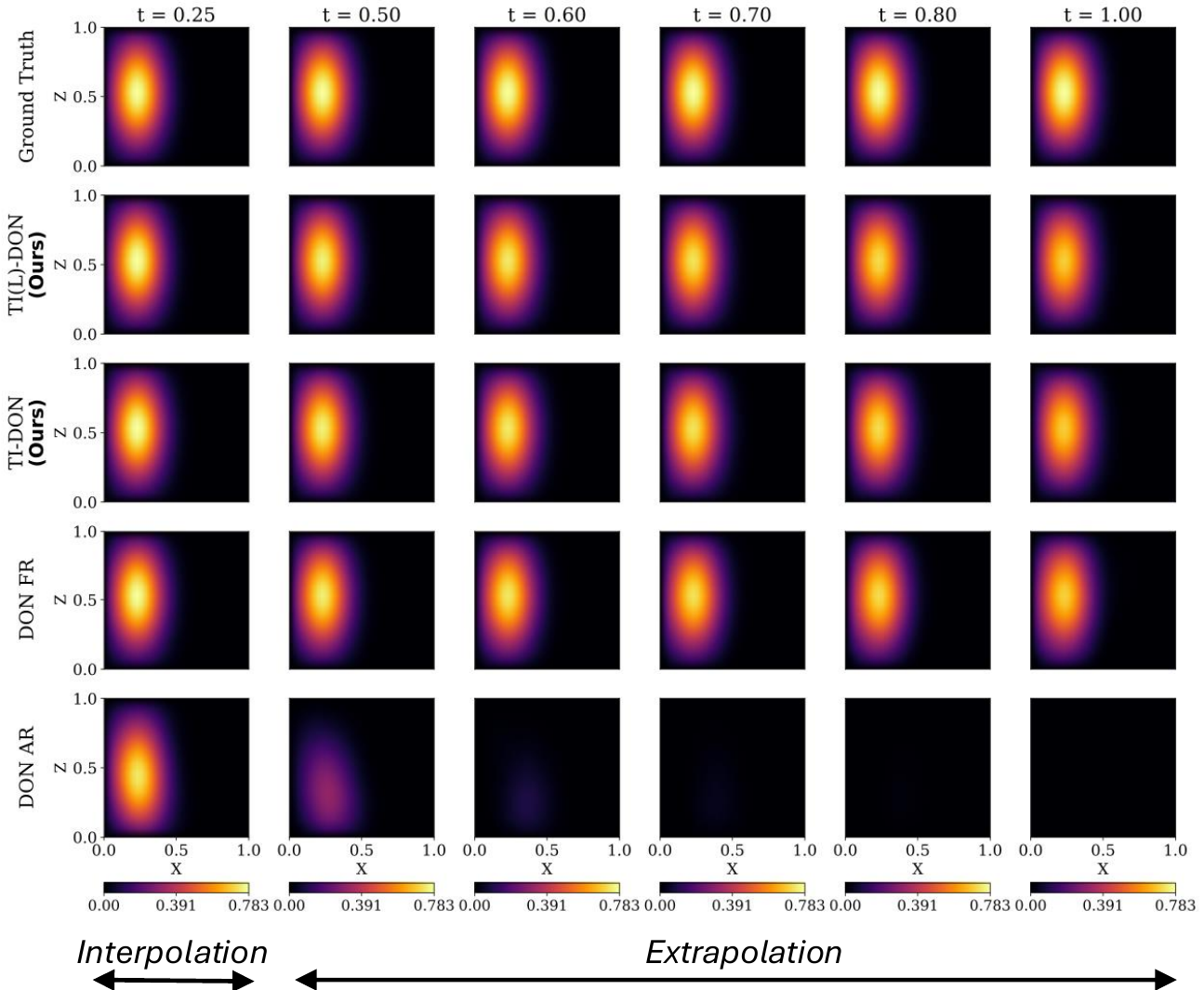}
    \caption{\rev{3D Heat Conduction: Solution field for a representative sample, shown across both the training regime ($t \in [0, 0.33]$) and the extrapolation regime ($t \in [0.33, 1]$), for all frameworks. Corresponding error plots are presented in Fig.~\ref{fig:3d_heat_error_contours_XZ}. The contours are presented along the XZ slicing plane (see Fig.~\ref{fig:3d_heat_slicing_planes} for details regarding the location of the XZ slice).}}
    \label{fig:sample_contourplots_3d_heat_XZ}
\end{figure}

\end{document}